\documentclass[sigconf]{acmart}

\AtBeginDocument{%
  \providecommand\BibTeX{{%
    \normalfont B\kern-0.5em{\scshape i\kern-0.25em b}\kern-0.8em\TeX}}}

\usepackage{graphicx}
\usepackage{booktabs}
\usepackage{multirow}
\usepackage{colortbl}

\begin{document}

\title{GVT2RPM: An Empirical Study for General Video Transformer Adaptation to Remote Physiological Measurement}

\author{Hao Wang}
\email{hao.wang2@sydney.edu.au}
\orcid{0000-0002-6235-8425}
\affiliation{%
  \institution{The University of Sydney}
  \city{Sydney}
  \state{NSW}
  \country{Australia}
  \postcode{2006}
}

\author{Euijoon Ahn}
\email{euijoon.ahn@jcu.edu.au}
\orcid{0000-0001-7027-067X}
\affiliation{%
  \institution{James Cook University}
  \city{Cairns}
  \state{QLD}
  \country{Australia}
  \postcode{4870}
}

\author{Jinman Kim}
\email{jinman.kim@sydney.edu.au}
\affiliation{%
  \institution{The University of Sydney}
  \city{Sydney}
  \state{NSW}
  \country{Australia}
  \postcode{2006}
}



\begin{abstract}
Remote physiological measurement (RPM) is an essential tool for healthcare monitoring as it enables the measurement of physiological signs, e.g., heart rate, in a remote setting via physical wearables. Recently, with facial videos, we have seen rapid advancements in video-based RPMs. However, adopting facial videos for RPM in the clinical setting largely depends on the accuracy and robustness (work across patient populations). Fortunately, the capability of the state-of-the-art transformer architecture in general (natural) video understanding has resulted in marked improvements and has been translated to facial understanding, including RPM. However, existing RPM methods usually need RPM-specific modules, e.g., temporal difference convolution and handcrafted feature maps. Although these customized modules can increase accuracy, they are not demonstrated for their robustness across datasets. Further, due to their customization of the transformer architecture, they cannot use the advancements made in general video transformers (GVT). In this study, we interrogate the GVT architecture and empirically analyze how the training designs, i.e., data pre-processing and network configurations, affect the model performance applied to RPM. Based on the structure of video transformers, we propose to configure its spatiotemporal hierarchy to align with the dense temporal information needed in RPM for signal feature extraction. We define several practical guidelines and gradually adapt GVTs for RPM without introducing RPM-specific modules. Our experiments demonstrate favorable results to existing RPM-specific module counterparts. We conducted extensive experiments with five datasets using intra-dataset (train and test on the same dataset) and cross-dataset settings (train and test on different datasets). We highlight that the proposed guidelines GVT2RPM can be generalized to any video transformers and is robust to various datasets. Code is available at \url{https://github.com/Dylan-H-Wang/facial-ai}.
\end{abstract}

\ccsdesc[300]{Human-centered computing~Empirical studies in HCI}
\ccsdesc[500]{Applied computing~Health care information systems}
\ccsdesc[300]{Computing methodologies~Biometrics}

\keywords{Remote Physiological Measurement, Video Transformer, Empirical Study, Video Analysis}
\begin{teaserfigure}
  \includegraphics[width=\linewidth]{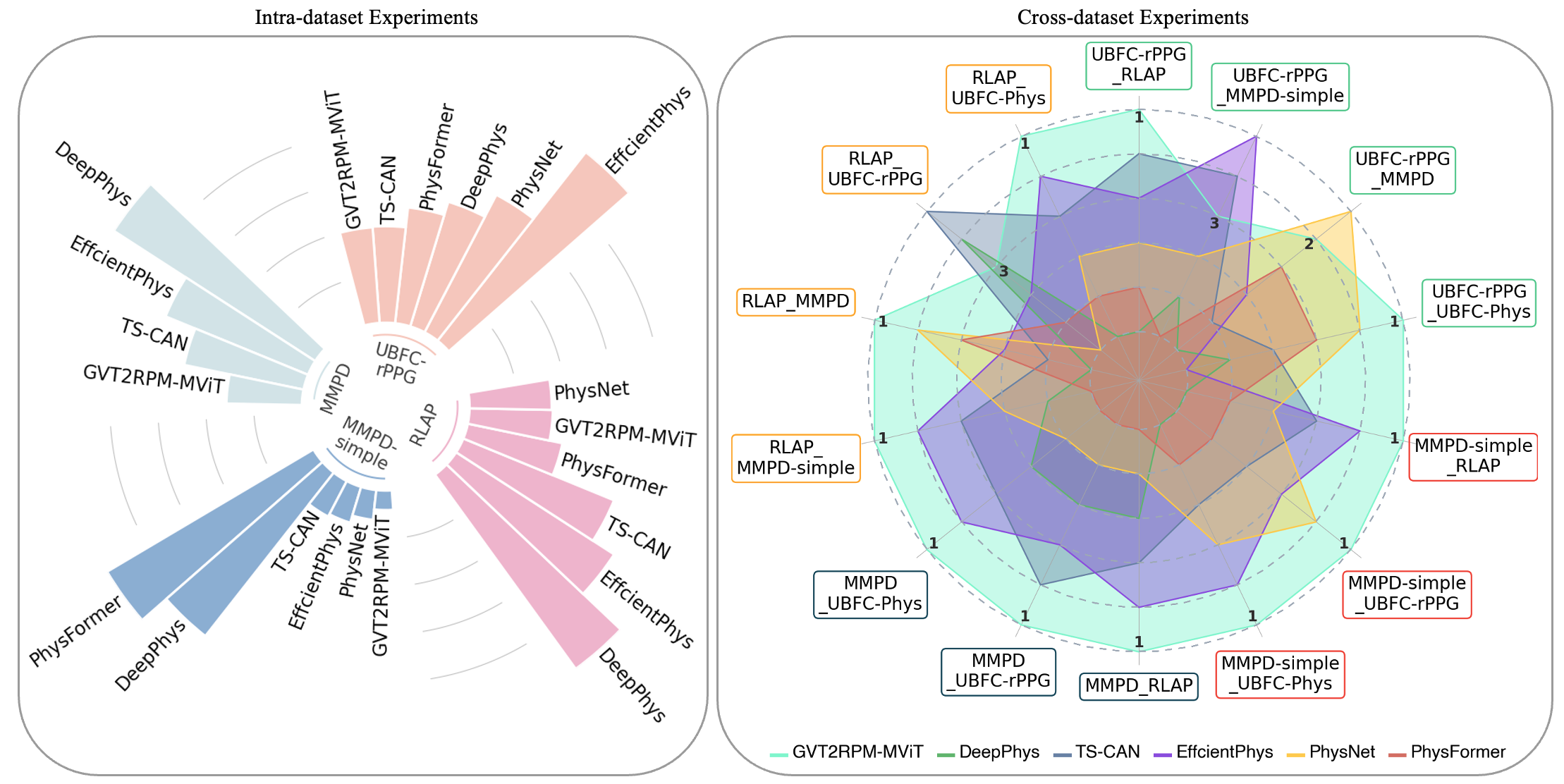}
  \caption{Overall performance evaluations of our GVT2RPM. A general video transformer (exemplified with MViTv2) was adapted for remote patient monitoring (GVT2RPM-MViT), compared to five other state-of-the-art methods. The left graph shows the Mean Absolute Errors (MAEs) from the comparison methods obtained in intra-dataset experiments where the training and testing sets were from the same dataset. A lower bar means a better result. We used different colors for each dataset. The right graph shows the ranking (based on their MAEs) of the comparison methods in cross-dataset experiments where the training and testing sets were from different datasets. A smaller number means a better result. Labels of the x-axis represent the name of training and testing datasets concatenated by an underscore. We used colored frames to indicate experiments with the same training set.}
  \label{fig:1}
\end{teaserfigure}

\maketitle

\section{Introduction}
\label{sec:intro}

Telemedicine has seen rapid growth in recent years due to the necessity and convenience offered by remote patient care \cite{ref1}. One of the key telemedicine services is remote patient monitoring (RPM), which can measure physiological signs that are necessary to cater for chronic and long-term patients at the convenience of their home environment \cite{ref2}. Traditional approaches for physiological measurement relied on physical contact wearables, e.g., heart rate or blood pressure monitors, which, although practical, are limited by their dependency on continuous physical attachment and can be cumbersome for long-term monitoring. To overcome these limitations, contactless RPM methods via remote photoplethysmography (rPPG) have garnered increased interest \cite{ref3}. A key advantage of rPPG is that it only requires a simple video camera, e.g., a smartphone camera \cite{ref4}. From the acquired video, rPPG signals are derived from light changes reflected from the skin caused by the Blood Volume Pulse (BVP) \cite{ref5} that are evident within the video frames, which can then be used to extract physiological parameters, such as Blood Pressure (BP) \cite{ref6}, Atrial Fibrillation (AF) \cite{ref7}, and Heart Rate (HR) \cite{ref8}. 

In early studies, video-based RPM relied on conventional machine learning techniques to detect and process rPPG signals. For example, researchers applied blind-source signal separation techniques, e.g., independent component analysis (ICA) \cite{ref9} and principal component analysis (PCA) \cite{ref10}, to reduce noises and recover the underlying rPPG waveforms from video frames. Furthermore, Wang et al. \cite{ref11} proposed to define a plane orthogonal to the color space of the skin, which eliminated specular reflections and improved the robustness of signal processing for rPPG recovery. More recently, deep learning approaches such as Convolutional Neural Networks (CNNs) have shown promising performance in image representation learning \cite{ref12,ref13,ref14} and video understanding \cite{ref15,ref16,ref17}. Researchers have leveraged advanced CNN architectures to improve the efficacy of video-based RPM algorithms. For instance, Chen et al. \cite{ref18} proposed a CNN with an attention mechanism to estimate and reduce signal noises by head motions. Špetlík et al. \cite{ref19} designed a two-stage CNN, where an “extractor” module was applied to learn video features. Then, a “predictor” module was used to analyze learned features for prediction. Other researchers learned video frame relationships by incorporating both spatial and temporal information, i.e., Yu et al. \cite{ref20} proposed to model the spatiotemporal relationships in videos using 3D CNN or 2D CNN combined with Recurrent Neural Network (RNN).

Transformer \cite{ref34}, a famous architecture in Natural Language Processing (NLP), has become a trending model architecture in computer vision, especially after the appearance of the Vision Transformer (ViT) \cite{ref21}. Compared to CNNs, the transformer has a larger receptive field than CNNs and thus provides a better long-range dependency. When applied to videos, the transformer can process long time-series data and demonstrates promising capability in modeling temporal relationships between the video frames, which is essential for understanding actions and scenes in videos \cite{ref22,ref23,ref24}. The advantages of the transformer architecture have also been applied to video-based RPM. For instance, Yu et al. \cite{ref25} proposed a video-transformer-based architecture for rPPG signal representation learning that Temporal Difference Convolution (TDC) \cite{ref26} was used for self-attention calculation capturing temporal difference features. Similarly, Liu et al. \cite{ref27} adopted the Swin Transformer, a transformer variant incorporating CNN multi-scale hierarchy, and converted 3D inputs to 2D feature maps for signal extraction.

However, our experiments found that transformer-based RPM methods required modifications, such as replacing the conventional transformer blocks with RPM-specific modules, e.g., the TDC \cite{ref26}, to enhance temporal signal extraction. The primary reason for such modification is the semantic differences between general video recognition tasks and RPM. For instance, significant variations in the human anatomy are modeled in general video recognition tasks. In contrast, detecting rPPG in RPM focuses on capturing the subtle color fluctuations from the human skin \cite{ref28} with a fixed camera viewpoint of a face (consistent anatomy). However, RPM-specific modules cannot be generalized to different transformer architectures and are not robust to different datasets. Instead of relying on specific modules, recent studies demonstrated the general capabilities of the transformer architecture that could be adapted for different tasks. For example, Khan et al. \cite{ref29} have shown that a transformer architecture can effectively extract audio signal features due to its long-term dependency modeling ability from the time-series data. Thus, it is intuitive to expect a similar performance on rPPG signal learning. 

In this work, we conducted an empirical study to adapt General Video Transformers to RPM (GVT2RPM) and proposed practical guidelines for their adaptation to solve RPM challenges. This adaptation maintains the structure of the original transformer architecture, thus making it adaptable to various video transformer architectures and robust to different datasets and settings. We propose several guidelines contributing to the adaptation, such as appropriate data pre-processing and additional temporal downsampling between the transformer blocks. Following our GVT2RPM guidelines in Section~\ref{sec:guideline}, we show steps to obtain optimal configurations for GVTs specific to datasets. As shown in Figure~\ref{fig:1}, we evaluated various methods on five commonly used public datasets, including MMPD-simple \cite{ref30}, MMPD \cite{ref30}, RLAP \cite{ref31}, UBFC-rPPG \cite{ref32}, and UBFC-Phys \cite{ref33} under intra- and cross-dataset settings using HR estimation to measure the quality of learned rPPG signals. Our results show that after applying the proposed GVT2RPM, the GVTs can yield favorable performance compared with the state-of-the-art (SOTA) RPM-specific transformer and CNN models. Moreover, we conducted a majority voting based on empirical results in Section~\ref{sec:empirical} and proposed general configurations for GVTs to achieve reasonable results on RPM, as shown in Section~\ref{sec:results}. Our GVT2RPM maintains the capacity of the original transformer architecture, making it easy to switch to the newly advanced transformer architecture and datasets and thus potentially reducing the need for extensive customization of RPM-specific modules.

\begin{figure*}[t!]
  \centering
  \includegraphics[width=.9\linewidth]{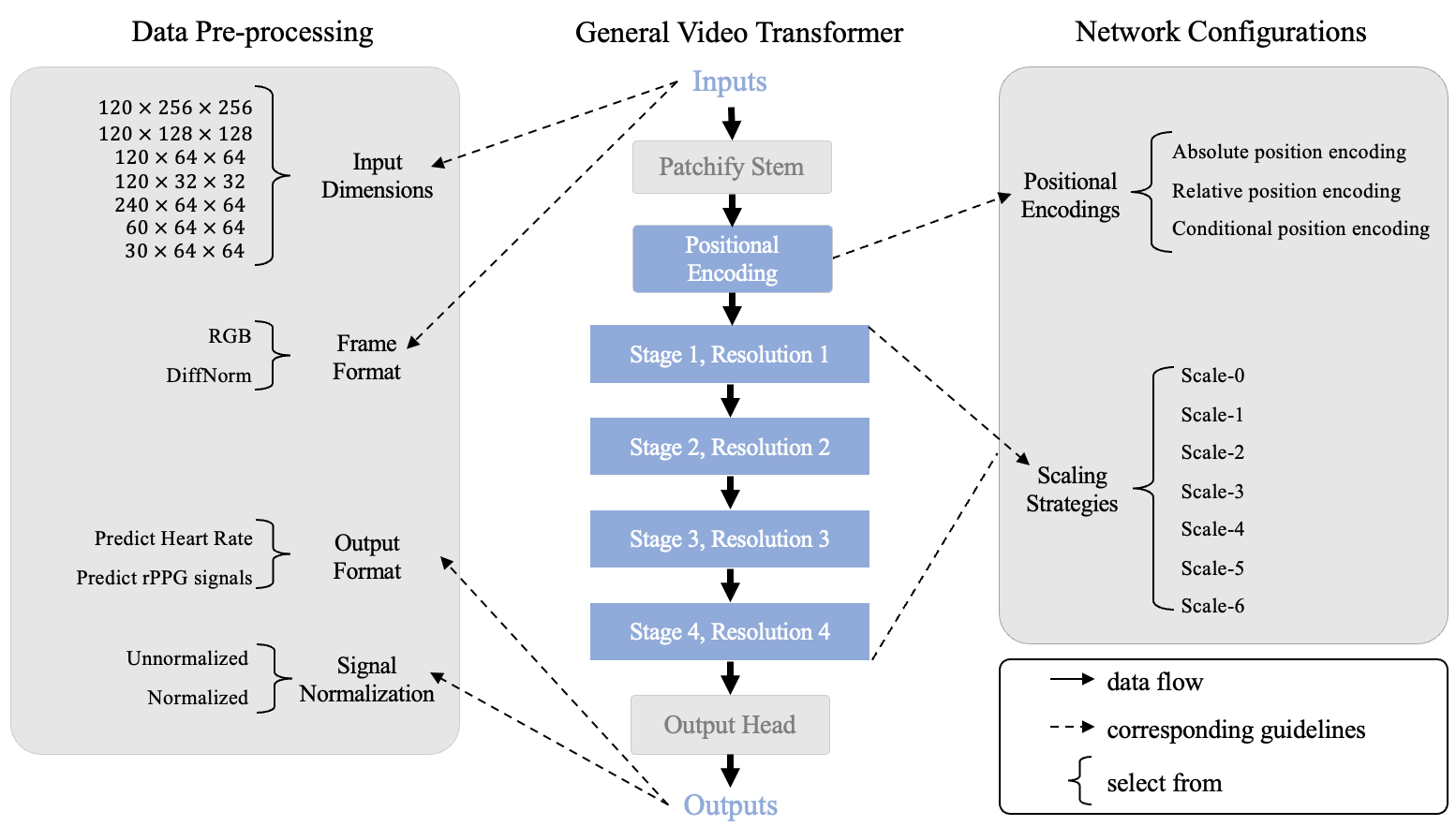}
  \caption{Overview of our proposed guidelines for adapting GVTs to remote physiological measurement. We used blue color to highlight the designs that could affect final performance. The parameters of each guideline are listed within the bracket and are selected based on empirical results.
  }
  \label{fig:2}
\end{figure*}

\section{Related Works}

\subsection{Transformer for General Video Analysis}
Applying 3D CNNs \cite{ref15,ref17} to capture the spatiotemporal relationships in the videos is intuitive. However, these methods have been constrained to the usage of short videos due to the limited receptive fields of the CNNs. In contrast, transformer \cite{ref34} designed for sequential data learning can handle long-range relationships and is therefore suitable for processing time-series data. For example, Bertasius et al. \cite{ref35} extended the ViT \cite{ref21} architecture to process 3D volume inputs where the self-attention mechanism was applied to spatial and temporal dimensions separately. Similarly, Arnab et al. \cite{ref36} proposed a transformer-based model in which video inputs were tokenized along spatial and temporal axes to produce 3D cubes, followed by a stack of transformer layers to learn spatiotemporal relationships. In contrast, rather than feeding raw cubes into the transformer, Neimark et al. \cite{ref37} integrated inductive biases and applied CNNs to extract features from each frame before sending them into the transformer to model the temporal relationships. Moreover, Fan et al. \cite{ref22} implemented multiscale feature hierarchies for the transformer to achieve efficient and effective video recognition, which they learned from the success of CNNs.

\subsection{Transformer for Remote Physiological Measurement}
The transformer can be helpful when applied to RPM. For example, Liu et al. \cite{ref27} proposed using tensor-shifted 2D convolutions \cite{ref39} to generate 2D feature maps from 3D videos, which were then fed into the 2D Transformer to learn the spatiotemporal relationships. Similarly, Liu et al. \cite{ref41} converted video inputs into handcrafted 2D spatiotemporal Map (STMap) representations \cite{ref42}, and then ViT was applied to extract underlying signal features. Instead of converting videos into 2D representations, Yu et al. \cite{ref25,ref43} used raw video inputs and applied TDC \cite{ref26} to calculate self-attention similarities for enhanced temporal feature learning. However, two limitations are identified for the above methods: 1) converting 3D facial videos into handcrafted 2D STMap requires prior knowledge about physiology, resulting in biases, and 2) customized transformer modules for the RPM deteriorate the model generalizability and hinder sharing the advancements from general video recognition.

\section{General Video Transformers (GVTs)}

Due to the additional time dimension in video inputs, learning the temporal dependencies among the video frames is essential for video understanding. Initially, researchers \cite{ref25,ref36} applied self-attention along the temporal axis to learn spatiotemporal relationships. This style of transformer designs evolved into a hybrid structure consisting of CNN and transformer \cite{ref23,ref24,ref38}. Specifically, a standard scheme of these methods is to integrate multiscale hierarchy in the CNN into the transformer to achieve a better speed-accuracy trade-off. In practice, the transformer architecture consists of five sequential stages. Firstly, the patchify stem is applied to project inputs into space-time cubes for later self-attention operations. Then, positional encodings are added for each cube. Afterward, cubes are fed into four stages sequentially to extract spatiotemporal features. Each stage contains multiple blocks consisting of transformer encoder or CNN block structures. The number of blocks in each stage depends on different model designs, but the feature map dimensions inside each stage remain the same. Therefore, the multiscale hierarchy only happens during the transition of stages and is agnostic to different hybrid video transformers.

\section{Guidelines for adapting GVT to RPM: Exemplified with MViTv2}
\label{sec:guideline}

\begin{figure}[tb]
  \centering
  \includegraphics[width=\linewidth]{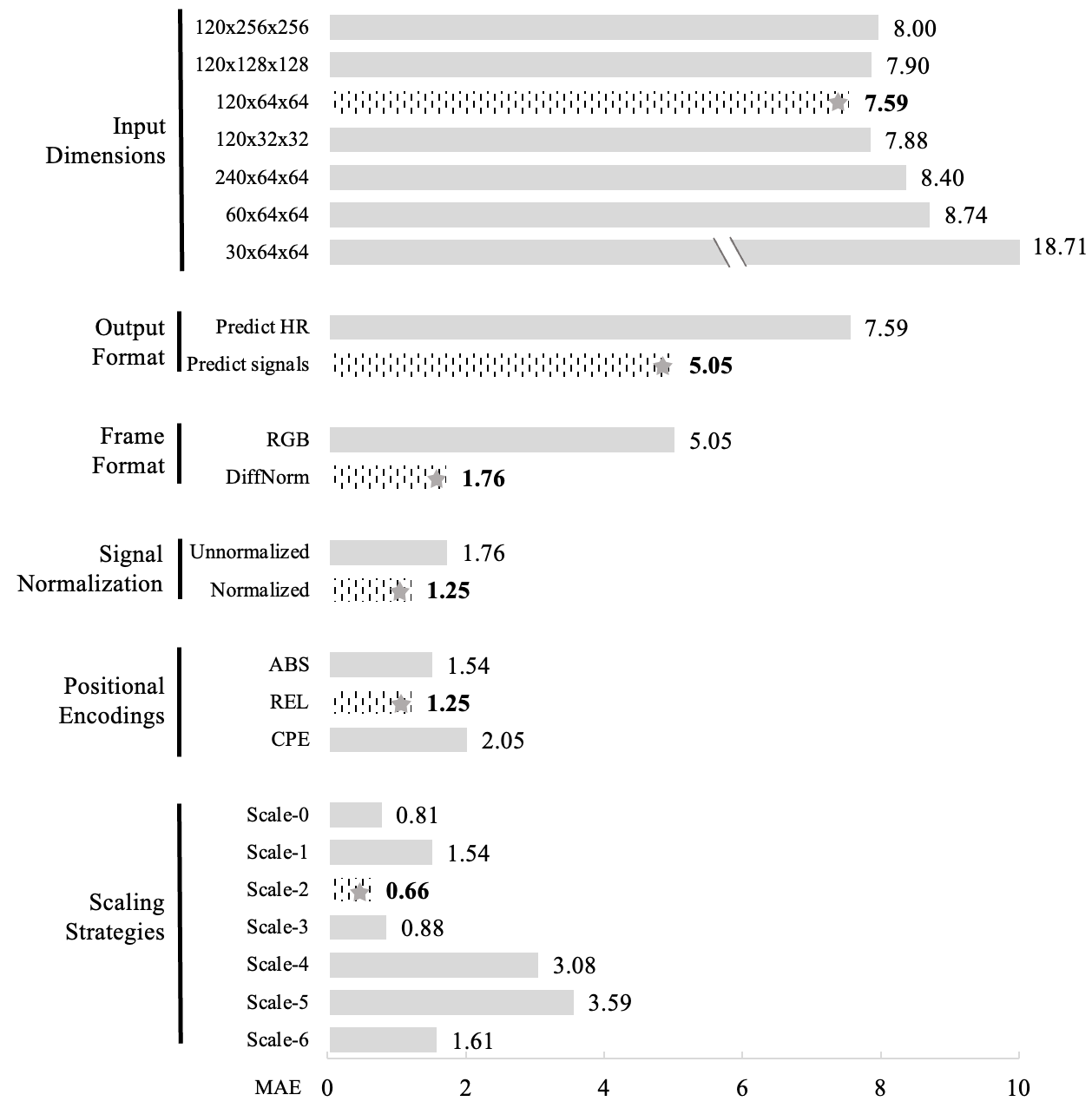}
  \caption{Experiment results under MMPD-simple intra-dataset setting by exploring the adaption from MViTv2 to GVT2RPM-MViT.
  }
  \label{fig:4}
\end{figure}

This section defines our guidelines for adapting the GVT to RPM for a specific dataset, as shown in Figure~\ref{fig:2}. We used the recent video transformer of Multiscale Vision Transformer v2 (MViTv2) \cite{ref38} as our baseline. Starting from the MViTv2’s standard settings, we show the changes in its performance from our adaptation in Figure~\ref{fig:4}.

Due to the differences between general video recognition and RPM, the training strategies can vary. To make experiments consistent and reproducible, we integrated the official MViTv2 implementation\footnote{\url{ https://github.com/facebookresearch/SlowFast/tree/main/projects/mvitv2}} with rPPG-Toolbox \cite{ref44} to benchmark algorithms and run experiments. We used the PyTorch \cite{ref45} library and kept most default training strategies in the rPPG-Toolbox: the batch size was set to 4, and the optimizer was AdamW \cite{ref46}. We extended the number of epochs to 50, modified the learning rate to 1e-3, and removed the learning rate scheduler to reduce hyperparameters. For simplicity, we used the MViTv2-S\footnote{Later use of MViTv2 refers to this model size unless specified.} as the backbone  and kept the hyperparameters unchanged unless specified. The model performance was evaluated under the intra-dataset setting (train/validation/test ratio of 7:1:2) on MMPD-simple \cite{ref30}, and the Mean Absolute Error (MAE) was used as the metric.

The guidelines consisted of two parts: 1) data pre-processing with four sub-parts: input dimensions, output format, frame format, and signal normalization, and 2) network configuration with two sub-parts: positional encodings and scaling strategies. Following each part sequentially, the model was adapted to RPM. The choice of each part is based on the greedy algorithm. For example, after grid-searching input dimensions, the $120\times64\times64$ achieves the best result, and then the output format exploration is conducted with this selected dimension.

\subsection{Data Pre-processing}
\subsubsection{Input dimensions}
The input dimensions of video recognition are usually $16\times224\times224$ (Frame length $T \times$ Height $H \times$ Width $W$), which is different from the RPM, e.g., $180\times72\times72$ is the default setting of input dimensions in rPPG-Toolbox. Therefore, these two tasks have opposite biases to the spatial and temporal information. In general video recognition, models need more spatial details to detect objects within the video. Still, they need sparse time-related information (e.g., key frames) to define an action (such as a tennis hit's beginning, middle, and ending moments). In contrast, RPM requires dense temporal information to capture continuous rPPG signal features. The spatial information, however, is less critical, providing facial semantics and containing potential noises disturbing the training process \cite{ref18,ref47}. Therefore, we tested a set of spatial dimensions \{256, 128, 64, 32\} where numbers are powers of 2 and a set of temporal dimensions \{240, 120, 60, 30\} where numbers are multiples of 30, which is the standard setting of Frames Per Second (FPS) for RPM. As shown in Figure~\ref{fig:4}, we first fixed the temporal dimension to 120 since it is close to the common setting of RPM and found that spatial dimension 64 achieved the best result with an MAE of 7.59. We then fixed the spatial dimension to 64 and found that the temporal dimension 120 performed the best. This suggests that compared with general video recognition, RPM requires a smaller frame size to reduce environmental noises and a longer clip length for enriched signal features.

\subsubsection{Output format}

\begin{figure}[tb]
  \centering
  \includegraphics[width=.9\linewidth]{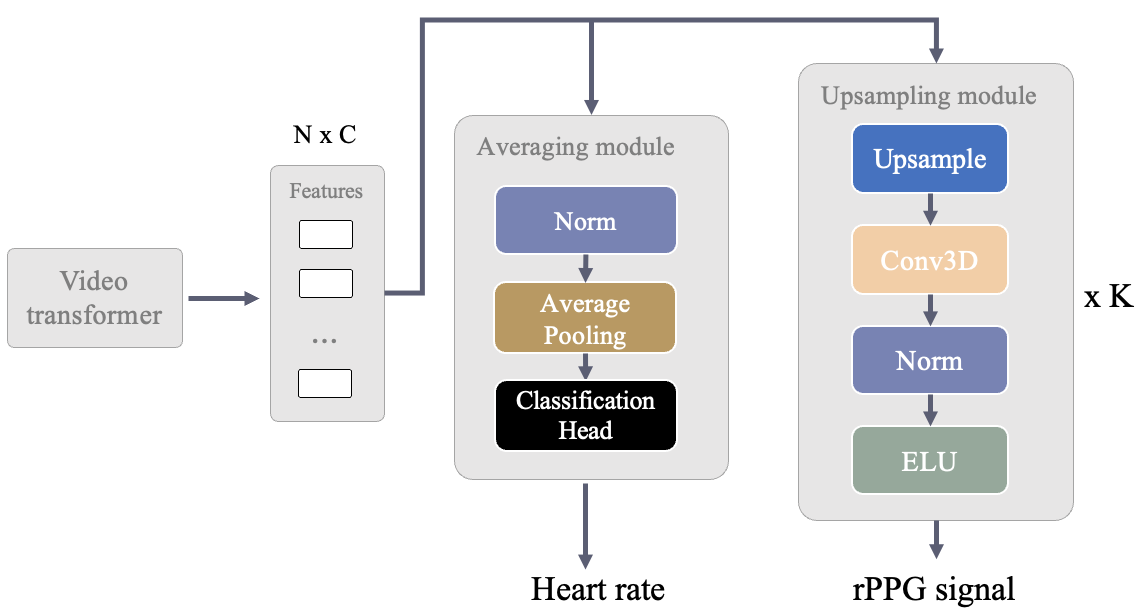}
  \caption{Design of upsampling module.
  }
  \label{fig:5}
\end{figure}

For video classification, the final predictions are made by either the class (CLS) token \cite{ref21} or by averaging output tokens from the last transformer block and applying a classification head \cite{ref28}. In RPM, outputs can be either rPPG signals or HR values derived from the rPPG signals. When the output format is HR, we can use the averaging module to make predictions without modifications. We also appended an upsampling module to map learned features to signals when predicting continuous signals. In detail, after the last transformer block, we added K of upsampling modules depending on the feature map dimensions and target signals. As shown in Figure~\ref{fig:5}, each upsampling module consists of a nearest upsampling layer, a 3D convolutional layer, a 3D batch normalization layer, and an Exponential Linear Unit (ELU) activation layer. The scaling factor of the upsampling layer was set to (2, 1, 1). The 3D convolutional layer kept the input and output dimensions the same with a kernel size of (3, 1, 1), stride of 1, and padding of (1, 0, 0). In the experiment, using rPPG signals as ground truth reduces the MAE to 5.05, showing that signals contain more information than HR values.

\subsubsection{Video frame format}
Using raw RGB video frames as inputs is common in video recognition. However, based on the skin reflection model \cite{ref11}, RGB inputs can be sub-optimal, affecting algorithm performances due to the reflection noises resulting from the light source and skin tone of subjects. Therefore, calculating Differences of Normlized frames (DiffNorm) \cite{ref18}, which minimizes the RGB input limitations, has become a popular data pre-processing to capture underlying rPPG signals under various illumination conditions. After applying DiffNorm to raw inputs, the MAE dropped from 5.05 to 1.76 (see Figure~\ref{fig:4}), indicating the effectiveness of DiffNorm in reducing inherent noises in RGB-format videos. 

\subsubsection{Signal normalization}
Normalizing signals into the same scale can help stabilize the gradient descent steps and improve the model convergence rate in most cases \cite{ref48}. Standardization, which transforms values to have a mean of 0 and a standard deviation of 1, is a prevalent normalization technique, assuming the data follows a Gaussian distribution \cite{ref49}. However, this assumption can be invalid for some datasets and thus hinder the model training. In our experiment, signal normalization reduced the MAE by 0.51 to 1.25 (see Figure~\ref{fig:4}).

\subsection{Network Configurations}
\subsubsection{Position encodings}
In contrast to CNNs, which inherently contain spatial information by the sliding window operation, the transformer processes all input tokens in parallel without referring to the order or positions. Understanding positional information is essential in vision recognition, and it helps to learn high-level semantic meanings like relationships between objects \cite{ref50}. In our studies, we evaluated three different position encodings, including absolute position encodings (ABS) \cite{ref22}, relative position encodings (REL) \cite{ref38}, and conditional position encodings (CPE) \cite{ref51}. Although ViT \cite{ref21} speculated that ABS and REL have no differences in image classification, MViTv2 \cite{ref38} found that REL can achieve better performances in video recognition. Recently, CPE was proposed to integrate translation equivalence into the vision transformer to improve performance. Since CPE was initially proposed for 2D images, we extended it for 3D video inputs by replacing the 2D CNN layers with 3D CNN. In this experiment, REL performed better than the other two position encodings with an MAE of 1.25 (see Figure~\ref{fig:4}).

\subsubsection{Scaling strategies}

The key idea of modern hybrid video transformers is to integrate multiscale feature hierarchies in CNNs with the transformer model. This is implemented by reducing the spatial resolution of feature maps and increasing the channel capacity at certain stages. With prior knowledge about rPPG signals, algorithms are required to extract dense temporal signals from facial videos with more frames than general video analysis (e.g., 120 v.s. 16 frames). Therefore, the scaling strategies in GVT, which only downsampling over spatial dimensions after the first stage, can be suboptimal for RPM. We experimented with different scaling strategies to investigate how space-time hierarchies affect model performances. In addition to reducing spatial resolutions only (Scale-0), we implemented reduction over temporal resolution to emphasize time hierarchy efficacy in RPM. As shown in Figure~\ref{fig:2}, Scale-1, Scale-2, and Scale-3 involve temporal downsampling at Stage 1, Stage 2, and Stage 3, respectively. Meanwhile, Scale-4, Scale-5, and Scale-6 employ more aggressive temporal scaling, reducing the temporal resolution at two stages through different combinations. Details are shown in the supplementary material. We find that introducing appropriate temporal downsampling is beneficial; in our case, Scale-2 achieved the lowest MAE of 0.66 (see Figure~\ref{fig:4}).

In summary, following the above guidelines, we obtained optimal configurations for MViTv2 to RPM (GVT2RPM-MViT) on MMPD-simple dataset: using an input dimension of $120\times64\times64$, using rPPG signals as labels, applying DiffNorm to represent frames, using normalized signals, using REL position encoding, and using Scale-2 strategy. Compared with the standard MViTv2 using input dimension $120\times64\times64$, the MAE dropped from 7.59 to 0.66 by 91.3\%.

\begin{table*}[t!]
  \caption{Results of adapting MViTv2 to GVT2RPM-MViT on MMPD-simple, MMPD, RLAP, and UBFC-rPPG under intra-dataset settings. MAE is used as the metric. The best combination of video frame format and signal normalization has a gray background, the best positional encoding has a blue background, and the best scaling strategy has a yellow background. 
  }
  \label{tab:1}
  \resizebox{\linewidth}{!}{%
  \centering
\begin{tabular}{llrrrrrrrlrrrrrrrrr} 
\toprule
\multirow{2}{*}{Datasets} &  & \multicolumn{2}{c}{\begin{tabular}[c]{@{}c@{}}Raw\\ Input\end{tabular}} & \multicolumn{1}{l}{} & \multicolumn{2}{c}{\begin{tabular}[c]{@{}c@{}}DiffNorm\\ Input\end{tabular}} & \multicolumn{1}{l}{} & \multicolumn{3}{c}{\begin{tabular}[c]{@{}c@{}}Positional\\ Encoding\end{tabular}} & \multicolumn{1}{l}{} & \multicolumn{7}{c}{\begin{tabular}[c]{@{}c@{}}Scaling\\ Strategies\end{tabular}} \\ 
\cmidrule(l){3-4}\cmidrule(l){6-7}\cmidrule(l){9-11}\cmidrule(l){13-19}
 &  & \multicolumn{1}{c}{\begin{tabular}[c]{@{}c@{}}w/o\\ signal\\ norm\end{tabular}} & \multicolumn{1}{c}{\begin{tabular}[c]{@{}c@{}}with\\ signal\\ norm\end{tabular}} & \multicolumn{1}{l}{} & \multicolumn{1}{c}{\begin{tabular}[c]{@{}c@{}}w/o\\ signal\\ norm\end{tabular}} & \multicolumn{1}{c}{\begin{tabular}[c]{@{}c@{}}with\\ signal\\ norm\end{tabular}} & \multicolumn{1}{l}{} & \multicolumn{1}{c}{ABS} & REL & \multicolumn{1}{c}{CPE} & \multicolumn{1}{l}{} & \multicolumn{1}{c}{Scale-0} & \multicolumn{1}{c}{Scale-1} & \multicolumn{1}{c}{Scale-2} & \multicolumn{1}{c}{Scale-3} & \multicolumn{1}{c}{Scale-4} & \multicolumn{1}{c}{Scale-5} & \multicolumn{1}{c}{Scale-6} \\ 
\midrule
MMPD-simple &  & 5.05 & 4.03 &  & 1.76 & {\cellcolor[rgb]{0.643,0.69,0.659}}\textbf{1.25} &  & 1.54 & {\cellcolor[rgb]{0.443,0.549,0.69}}\textbf{1.25} & 2.05 &  & 0.81 & 1.54 & {\cellcolor[rgb]{0.949,0.808,0.639}}\textbf{0.66} & 0.88 & 3.08 & 3.59 & 1.61 \\
MMPD &  & 13.06 & 7.23 &  & {\cellcolor[rgb]{0.643,0.69,0.659}}\textbf{7.22} & 7.72 &  & 7.3 & {\cellcolor[rgb]{0.443,0.549,0.69}}\textbf{7.22} & 8.83 &  & 8.65 & 7.21 & 8.67 & 8.37 & 7.69 & 7.04 & {\cellcolor[rgb]{0.949,0.808,0.639}}\textbf{6.83} \\
RLAP &  & 17.36 & 1.69 &  & 1.67 & {\cellcolor[rgb]{0.643,0.69,0.659}}\textbf{1.48} &  & 1.87 & {\cellcolor[rgb]{0.443,0.549,0.69}}\textbf{1.48} & 2.11 &  & 1.7 & 1.69 & {\cellcolor[rgb]{0.949,0.808,0.639}}\textbf{1.38} & 1.45 & 1.54 & 1.82 & 1.74 \\
UBFC-rPPG &  & 3.03 & 3.91 &  & 2.93 & {\cellcolor[rgb]{0.643,0.69,0.659}}\textbf{2.83} &  & {\cellcolor[rgb]{0.443,0.549,0.69}}\textbf{2.14} & 2.83 & 2.15 &  & 2.15 & 1.76 & 2.25 & 1.66 & 1.95 & {\cellcolor[rgb]{0.949,0.808,0.639}}\textbf{1.56} & 2.93 \\
\bottomrule
\end{tabular}
    }
\end{table*}

\section{Empirical Evaluations on Different Datasets and Settings}
\label{sec:empirical}

Following the guidelines in Section~\ref{sec:guideline}, we exemplified how to adapt the GVT (e.g., MViTv2) to RPM and find optimal configurations for a particular dataset without using RPM-specific modules. In this section, we evaluate the robustness of the GVT2RPM by conducting intra-dataset experiments on three additional datasets and cross-dataset experiments on five datasets.

\subsection{Datasets}
\begin{itemize}
    \item MMPD \cite{ref30}: This dataset contains videos recorded by a Samsung Galaxy S22 Ultra mobile phone at 30 FPS with a resolution of $1280\times720$ and compressed to $320\times240$ stored in H.264 format. An HKG-07C+ oximeter records the ground truth PGG signals. Videos are recorded under four lighting conditions, motions, and skin tones.
    \item MMPD-Simple \cite{ref30}: Due to the difficulty of the original MMPD, authors created a subset to contain videos with stationary, skin tone type 3, and artificial light conditions.
    \item RLAP \cite{ref31}: This dataset contains videos recorded by a Logitech C920c webcam at 30 FPS with a resolution of $1920\times1080$ stored in MJPG format. A CMS50E transmissive pulse oximeter records the ground truth PPG signals. During video recording, subjects completed tasks or watched videos under different lighting conditions.
    \item UBFC-rPPG \cite{ref32}: This dataset contains videos recorded by a Logitech C920 HD Pro webcam at 30 FPS with a resolution of $640\times480$ in uncompressed 8-bit RGB format. A CMS50E transmissive pulse oximeter records corresponding PPG signals. The recording is conducted indoors with sufficient sunlight and artificial illumination. 
    \item UBFC-Phys \cite{ref33}: This dataset contains videos recorded by an EO-23121C RGB digital camera at 35 FPS with a resolution of $1024\times1024$ stored in MJPG format. The underlying BVP signals were recorded by the Empatica E4 wristband. The collection is conducted with three tasks with significant amounts of unconstrained motion under static lighting conditions.
\end{itemize}

\subsection{Experiment Settings}
We conducted two types of experiments, including intra-dataset and cross-dataset experiments. For the intra-dataset experiments, models were trained, validated, and tested on the same dataset with a split ratio of 7:1:2. For the cross-dataset experiments, models were trained and validated on the same dataset with a split ratio of 8:2, and then tested on another dataset.

The training process was the same as in the Section~\ref{sec:guideline}. It was consistent for intra-dataset and cross-dataset experiments, except that we fixed the input dimensions to $120\times64\times64$ and the output format to rPPG signals, as this combination consistently had good performances. We evaluated the model performance based on the metric MAE.

The exploration of model designs consisted of 3 parts: video frame format and signal normalization, positional encodings, and scaling strategies. Each part choice was based on the greedy algorithm.

\begin{table*}[th]
\caption{Results of adapting MViTv2 to GVT2RPM-MViT on MMPD-simple, MMPD, RLAP, UBFC-rPPG, and UBFC-Phys under cross-dataset settings. MAE is used as the evaluation metric. The best combination of video frame format and signal normalization has a gray background, the best positional encoding has a blue background, and the best scaling strategy has a yellow background.}
\label{tab:2}
\centering
\setlength{\extrarowheight}{0pt}
\addtolength{\extrarowheight}{\aboverulesep}
\addtolength{\extrarowheight}{\belowrulesep}
\setlength{\aboverulesep}{0pt}
\setlength{\belowrulesep}{0pt}
\resizebox{\linewidth}{!}{%
\begin{tabular}{llrrrrrrrrlrlrrrrrrr} 
\toprule
\multicolumn{1}{c}{\multirow{2}{*}{\begin{tabular}[c]{@{}c@{}}Train\\ Dataset\end{tabular}}} & \multicolumn{1}{c}{\multirow{2}{*}{\begin{tabular}[c]{@{}c@{}}Test\\ Dataset\end{tabular}}} & \multicolumn{1}{l}{} & \multicolumn{2}{c}{\begin{tabular}[c]{@{}c@{}}Raw\\ Input\end{tabular}} & \multicolumn{1}{l}{} & \multicolumn{2}{c}{\begin{tabular}[c]{@{}c@{}}DiffNorm\\ Input\end{tabular}} & \multicolumn{1}{l}{} & \multicolumn{3}{c}{\begin{tabular}[c]{@{}c@{}}Positional\\ Encoding\end{tabular}} & \multicolumn{1}{r}{} & \multicolumn{7}{c}{\begin{tabular}[c]{@{}c@{}}Scaling\\ Strategies\end{tabular}} \\ 
\cmidrule{4-5}\cmidrule{7-8}\cmidrule{10-12}\cmidrule{14-20}
\multicolumn{1}{c}{} & \multicolumn{1}{c}{} & \multicolumn{1}{l}{} & \multicolumn{1}{c}{\begin{tabular}[c]{@{}c@{}}w/o\\ signal\\ norm\end{tabular}} & \multicolumn{1}{c}{\begin{tabular}[c]{@{}c@{}}with\\ signal\\ norm\end{tabular}} & \multicolumn{1}{l}{} & \multicolumn{1}{c}{\begin{tabular}[c]{@{}c@{}}w/o\\ signal\\ norm\end{tabular}} & \multicolumn{1}{c}{\begin{tabular}[c]{@{}c@{}}with\\ signal\\ norm\end{tabular}} & \multicolumn{1}{l}{} & \multicolumn{1}{c}{ABS} & REL & \multicolumn{1}{c}{CPE} &  & \multicolumn{1}{c}{Scale-0} & \multicolumn{1}{c}{Scale-1} & \multicolumn{1}{c}{Scale-2} & \multicolumn{1}{c}{Scale-3} & \multicolumn{1}{c}{Scale-4} & \multicolumn{1}{c}{Scale-5} & \multicolumn{1}{c}{Scale-6} \\ 
\midrule
{\cellcolor[rgb]{0.945,0.945,0.945}} & UBFC-rPPG & \multicolumn{1}{l}{} & 28.17 & 27.71 &  & {\cellcolor[rgb]{0.643,0.69,0.659}}\textbf{8.14} & 9.12 &  & 7.3 & 8.14 & {\cellcolor[rgb]{0.443,0.549,0.69}}\textbf{1.46} &  & {\cellcolor[rgb]{0.949,0.808,0.635}}\textbf{6.11} & 13.81 & 12.72 & 7.76 & 25.3 & 23.14 & 21.66 \\
{\cellcolor[rgb]{0.945,0.945,0.945}} & UBFC-Phys & \multicolumn{1}{l}{} & 12.42 & 11.13 &  & {\cellcolor[rgb]{0.643,0.69,0.659}}\textbf{6.04} & 7.14 &  & {\cellcolor[rgb]{0.443,0.549,0.69}}\textbf{5.09} & 6.04 & 7.04 &  & {\cellcolor[rgb]{0.949,0.808,0.635}}\textbf{5.47} & 5.92 & 5.92 & 5.89 & 7.5 & 6.28 & 7.28 \\
\multirow{-3}{*}{{\cellcolor[rgb]{0.945,0.945,0.945}}\begin{tabular}[c]{@{}>{\cellcolor[rgb]{0.945,0.945,0.945}}l@{}}MMPD-\\simple\end{tabular}} & RLAP & \multicolumn{1}{l}{} & 7.68 & 8.82 &  & {\cellcolor[rgb]{0.643,0.69,0.659}}\textbf{5.23} & 5.31 &  & 5.81 & 5.23 & {\cellcolor[rgb]{0.443,0.549,0.69}}\textbf{3.7} &  & 5.78 & 3.29 & {\cellcolor[rgb]{0.949,0.808,0.635}}\textbf{3.29} & 4.58 & 6.44 & 4.99 & 4.96 \\ 
\midrule
{\cellcolor[rgb]{0.945,0.945,0.945}} & UBFC-rPPG &  & 20.57 & 15.95 &  & {\cellcolor[rgb]{0.643,0.69,0.659}}\textbf{5.75} & 8.06 &  & 2.78 & 5.75 & {\cellcolor[rgb]{0.443,0.549,0.69}}\textbf{2.49} &  & {\cellcolor[rgb]{0.949,0.808,0.635}}\textbf{2.2} & 3.87 & 4.46 & 3.98 & 8.56 & 7.47 & 9.14 \\
{\cellcolor[rgb]{0.945,0.945,0.945}} & UBFC-Phys &  & 10.8 & 24.68 &  & {\cellcolor[rgb]{0.643,0.69,0.659}}\textbf{5.49} & 10.18 &  & {\cellcolor[rgb]{0.443,0.549,0.69}}\textbf{5.16} & 5.49 & 6.24 &  & {\cellcolor[rgb]{0.949,0.808,0.635}}\textbf{4.5} & 5.69 & 6.07 & 6.05 & 5.26 & 4.95 & 7.9 \\
\multirow{-3}{*}{{\cellcolor[rgb]{0.945,0.945,0.945}}MMPD} & RLAP &  & 20.19 & 15.48 &  & {\cellcolor[rgb]{0.643,0.69,0.659}}\textbf{3.42} & 7.16 &  & {\cellcolor[rgb]{0.443,0.549,0.69}}\textbf{3.4} & 3.42 & 3.76 &  & {\cellcolor[rgb]{0.949,0.808,0.635}}\textbf{3.98} & 5.23 & 4.75 & 4.57 & 4.04 & 5.14 & 4.76 \\ 
\midrule
{\cellcolor[rgb]{0.945,0.945,0.945}} & MMPD-simple &  & 18.07 & 4.71 &  & 3.23 & {\cellcolor[rgb]{0.643,0.69,0.659}}\textbf{1.92} &  & {\cellcolor[rgb]{0.443,0.549,0.69}}\textbf{0.79} & 1.92 & 1.38 &  & {\cellcolor[rgb]{0.949,0.808,0.635}}\textbf{0.97} & 2.31 & 3.44 & 1.17 & 2.59 & 2.6 & 4.14 \\
{\cellcolor[rgb]{0.945,0.945,0.945}} & MMPD &  & 13.48 & 11.75 &  & 10.03 & {\cellcolor[rgb]{0.643,0.69,0.659}}\textbf{9.02} &  & 9.95 & {\cellcolor[rgb]{0.443,0.549,0.69}}\textbf{9.02} & 9.75 &  & 9.2 & 9.44 & {\cellcolor[rgb]{0.949,0.808,0.635}}\textbf{8.58} & 9.72 & 8.86 & 9.39 & 9.7 \\
{\cellcolor[rgb]{0.945,0.945,0.945}} & UBFC-rPPG &  & 19.48 & 12.74 &  & 6.36 & {\cellcolor[rgb]{0.643,0.69,0.659}}\textbf{5.57} &  & 2.05 & 5.57 & {\cellcolor[rgb]{0.443,0.549,0.69}}\textbf{1.48} &  & 2.36 & 5.44 & 4.33 & {\cellcolor[rgb]{0.949,0.808,0.635}}\textbf{1.9} & 5.57 & 5.61 & 4.04 \\
\multirow{-4}{*}{{\cellcolor[rgb]{0.945,0.945,0.945}}RLAP} & UBFC-Phys &  & 10.97 & 4.96 &  & 4.57 & {\cellcolor[rgb]{0.643,0.69,0.659}}\textbf{4.32} &  & {\cellcolor[rgb]{0.443,0.549,0.69}}\textbf{4.19} & 4.32 & 4.31 &  & 4.31 & 4.68 & 4.9 & {\cellcolor[rgb]{0.949,0.808,0.635}}\textbf{4.17} & 4.44 & 4.76 & 4.49 \\ 
\midrule
{\cellcolor[rgb]{0.945,0.945,0.945}} & MMPD-simple &  & 16.8 & 22.24 &  & 1.96 & {\cellcolor[rgb]{0.643,0.69,0.659}}\textbf{1.87} &  & 3.14 & {\cellcolor[rgb]{0.443,0.549,0.69}}\textbf{1.87} & 2.55 &  & {\cellcolor[rgb]{0.949,0.808,0.635}}\textbf{2.73} & 4.34 & 5.94 & 4.39 & 10.6 & 7.32 & 5.54 \\
{\cellcolor[rgb]{0.945,0.945,0.945}} & MMPD &  & 14.12 & 17.31 &  & 12.2 & {\cellcolor[rgb]{0.643,0.69,0.659}}\textbf{11.5} &  & 12.47 & {\cellcolor[rgb]{0.443,0.549,0.69}}\textbf{11.5} & 12.67 &  & 11.71 & {\cellcolor[rgb]{0.949,0.808,0.635}}\textbf{11.28} & 11.71 & 13.16 & 12.02 & 12.02 & 11.45 \\
{\cellcolor[rgb]{0.945,0.945,0.945}} & UBFC-Phys &  & 6.46 & 6.2 &  & {\cellcolor[rgb]{0.643,0.69,0.659}}\textbf{4.49} & 4.75 &  & {\cellcolor[rgb]{0.443,0.549,0.69}}\textbf{4.36} & 4.49 & 5.3 &  & 4.72 & 5.31 & 4.78 & {\cellcolor[rgb]{0.949,0.808,0.635}}\textbf{4.36} & 5.01 & 4.99 & 5.11 \\
\multirow{-4}{*}{{\cellcolor[rgb]{0.945,0.945,0.945}}\begin{tabular}[c]{@{}>{\cellcolor[rgb]{0.945,0.945,0.945}}l@{}}UBFC-\\rPPG\end{tabular}} & RLAP &  & 6.63 & 7.41 &  & {\cellcolor[rgb]{0.643,0.69,0.659}}\textbf{3.01} & 3.29 &  & 3.49 & {\cellcolor[rgb]{0.443,0.549,0.69}}\textbf{3.01} & 3.49 &  & 3.32 & 3.45 & 3.26 & {\cellcolor[rgb]{0.949,0.808,0.635}}\textbf{3.07} & 3.91 & 3.88 & 3.35 \\
\bottomrule
\end{tabular}
}
\end{table*}

\subsection{Intra-dataset Experiments}
We conducted intra-dataset experiments on MMPD-simple, MMPD, RLAP, and UBFC-rPPG. Table~\ref{tab:1} summarizes the performance of different designs of GVT2RPM-MViT on intra-dataset experiments. There are some findings we have identified:

\subsubsection{DiffNorm always helps.}
Compared with raw RGB inputs, videos pre-processed by the DiffNorm always perform better. Considering the situation without signal normalization, DiffNorm reduced MAE over half for MMPD-simple and RLAP, 44.7\% for MMPD, and 3.3\% for UBFC-rPPG. This demonstrates that DiffNorm is key to adapting from a GVT to RPM. It amplifies the underlying rPPG signals by suppressing the motion and illumination noises in the raw RGB videos and enforces the transformer to focus on subtle pixel variations instead of human anatomy.

\subsubsection{Signal normalization helps in simple scenarios.}
We observe that signal normalization helps when the dataset contains relatively simple settings, e.g., non-rigid movement and constant and sufficient illumination, such that the MAE was reduced from 1.76 to 1.25 in MMPD-simple, from 1.67 to 1.48 in RLAP, and from 2.93 to 2.83 in UBFC-rPPG. MMPD, having rigid head motions, various skin tones, and changing lighting conditions, can cause many outliers and break Gaussian distribution, thus making it incompatible with signal normalization.

\subsubsection{Relative positional encoding is robust in most cases.}
The REL obtained lower MAEs on MMPD-simple, MMPD, and RLAP than the other two positional encodings. Except for UBFC-rPPG, the ABS achieved a lower MAE of 2.14.

\subsubsection{Appropriate temporal hierarchy helps signal learning.} 
We find that introducing temporal scaling between transformer blocks assists better understanding of signals such that the MAE was reduced from 0.81 to 0.66 in MMPD-simple, from 8.65 to 6.83 in MMPD, from 1.7 to 1.38 in RLAP, and from 2.14 to 1.56 in UBFC-rPPG.

\subsection{Cross-dataset Experiments}

We conducted cross-dataset experiments on MMPD-simple, MMPD, UBFC-rPPG, UBFC-Phys, and RLAP. Table~\ref{tab:2} shows the performance of different designs of GVT2RPM-MViT on cross-dataset experiments. The choices of designs have different effects compared with the intra-dataset setting, and we conclude:

\subsubsection{DiffNorm significantly improves transfer learning.}
Similar to intra-dataset experiments, DiffNorm is also beneficial for transfer learning. Considering the situation without signal normalization, we observe that applying DiffNorm can reduce MAEs by an average of 55.6\% over all cases.

\subsubsection{The efficacy of signal normalization depends on the training dataset.}
We noticed that normalizing signals can hinder the model learning when the training was conducted on MMPD-simple and MMPD. It suggests that transferring from low-quality video datasets (H.264 compressed) to higher-quality datasets (MJPG compressed or uncompressed) should not normalize signals.  In contrast, when training on the RLAP, signal normalization can help the transfer learning.

\subsubsection{Positional encoding can be selected based on the target dataset.}
Unlike intra-dataset experiments where REL was better in most cases, CPE performed better when the target dataset was UBFC-rPPG, while ABS succeeded when the target dataset was UBFC-Phys. Specifically, replacing REL with CPE decreased MAEs by an average of 70.7\% when the target dataset is UBFC-rPPG. Similarly, altering REL with ABS reduced MAEs by an average of 6.91\% when the target dataset is UBFC-Phys.

\subsubsection{Spatial hierarchy is more robust for transfer learning.}
We observe that half of the cases showed better performances for Scale-0, which is the default setting of GVT without involving downsampling over temporal dimensions, and the other half of cases performed better when the strategy was Scale-1, Scale-2, or Scale-3 where temporal scaling was conducted once between transformer blocks. Applying more than one temporal scaling, e.g., Scale-4, Scale-5, and Scale-6, prevented the model from learning robust signal features.

\section{Evaluations of Different GVTs}
\label{sec:results}

\begin{figure*}[th!]
  \centering
  \includegraphics[width=\linewidth]{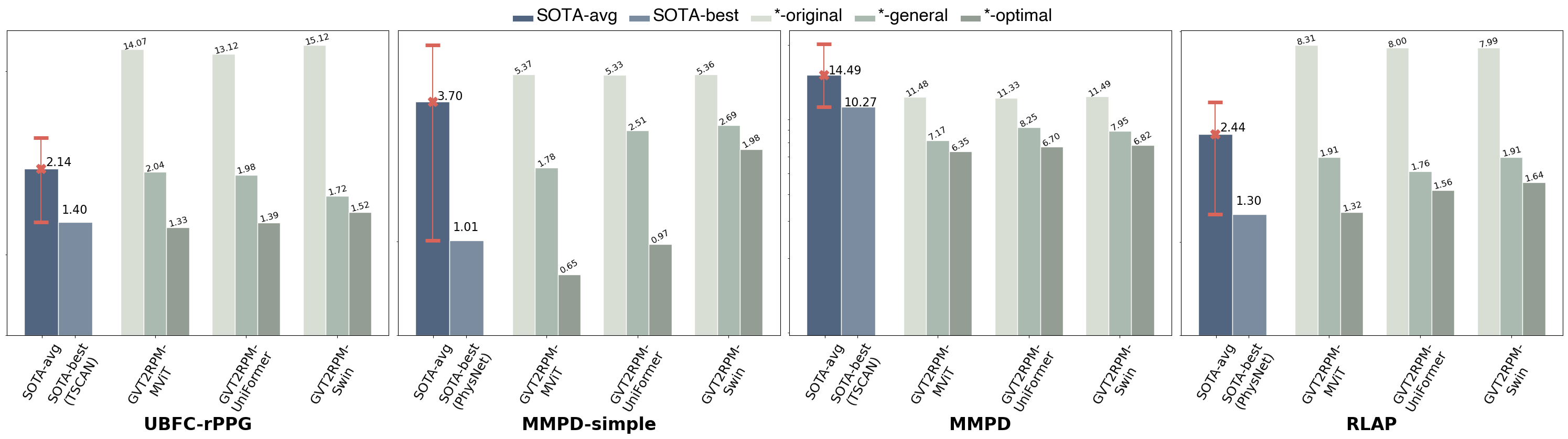}
  \caption{Intra-dataset experiment results on MMPD-simple, MMPD, RLAP, and UBFC-rPPG. We evaluated five RPM SOTA methods. Their averaged results are denoted by SOTA-avg with error bar. The best performing method is denoted by SOTA-best. Also, we tested three GVTs, including MViTv2, UniFormer, and Video Swin. Based on our empirical results in Section~\ref{sec:empirical}, we constructed GVT2RPM-MViT-general, GVT2RPM-UniFormer-general, and GVT2RPM-Swin-general. Following Section~\ref{sec:guideline}, we further optimized them into GVT2RPM-MViT-optimal, GVT2RPM-UniFormer-optimal, and GVT2RPM-Swin-optimal. We evaluated the performance by MAE.
  }
  \label{fig:7}
\end{figure*}

This section demonstrates the application of our guidelines to multiple GVTs where we employed the recent SOTA GVT of UniFormer \cite{ref24}, and Video Swin \cite{ref23}, in addition to the MViTv2 \cite{ref38} from Section~\ref{sec:guideline}. We conducted intra-dataset experiments to evaluate their performance on MMPD-simple, MMPD, RLAP, and UBFC-rPPG. For fair comparisons, we applied 3-fold cross-validation except for RLAP, where the official split \cite{ref31} was used. 

Based on the empirical results of intra-dataset experiments in Section~\ref{sec:empirical}, we conducted a majority voting and proposed general configurations for GVTs: 1) for data pre-processing, the input clip dimension was $120\times64\times64$, signals were used for outputs, and each frame was pre-processed by DiffNorm; 2) depending on dataset complexity, output signals were normalized for simple scenarios; 3) for network configurations, REL positional encoding was used, and the scaling strategy was set to Scale-2. Therefore, the above GVTs were adapted to GVT2RPM-MViT-general, GVT2RPM-UniFormer-general, and GVT2RPM-Swin-general. Additionally, each model was optimized following the proposed guidelines in Section~\ref{sec:guideline} to obtain the optimal configurations. Therefore, we have GVT2RPM-MViT-optimal, GVT2RPM-UniFormer-optimal, and GVT2RPM-Swin-optimal (configuration details in the supplementary). To compare, we trained SOTA RPM methods, including DeepPhys \cite{ref18}, PhysNet \cite{ref20}, TSCAN \cite{ref52}, EfficientPhys \cite{ref27}, and PhysFormer \cite{ref43}, using rPPG-toolbox with default settings and averaged their performance as the baseline, named SOTA-avg. The best-performing method was denoted as SOTA-best.

Experimental results are shown in Figure~\ref{fig:7}. We observe that applying our general configurations successfully adapted GVTs to RPM, where all three GVT2RPM-*-general methods achieved better MAEs than the SOTA-avg in four datasets. In contrast, naively using GVTs for RPM with their original versions (GVT2RPM-*-original), as expected, performed worse than the SOTA-avg except with the MMPD. This indicates that our GVT2RPM is generalizable to different GVTs and robust to various datasets.

Moreover, optimizing GVTs using our guidelines achieved better results and competed favorably with the RPM-specific SOTA methods. GVT2RPM-MViT-optimal obtained better results than the other two GVTs and outperformed the SOTA-best in UBFC-rPPG, MMPD-simple, and MMPD. The best performance of GVT2RPM-MViT is consistent with performance patterns on general video tasks, where the MViTv2 with more practical designs (e.g., residual connections and Key-Value pooling) yielded better video representations than UniFormer and Video Swin. Moreover, we observe that optimizing MViTv2 from its general version resulted in more benefits than the optimization with the other two GVTs. 

Over the four datasets, all methods did not perform well in the MMPD. We suggest this is due to the MMPD exhibiting videos with complex scenarios, such as rigid patient face movements, varying lighting conditions, and different skin tones, relative to the other video sets. It is interesting to note that in this complex dataset, all GVT2RPM-*-origin outperformed the SOTA-avg, which suggests that the GVTs are more capable of handling complex data. Additionally, our GVT2RPM further improved their performance such that both GVT2RPM-*-general and GVT2RPM-*-optimal achieved better MAEs than the SOTA-best.

\section{Conclusions}
In this paper, we conducted empirical research for adapting GVTs to the RPM. We demonstrated that, without RPM-specific modules such as TDC or handcrafted STMap representations, GVTs can still obtain reasonable results with simple adjustments for the data pre-processing. Furthermore, optimizing the transformer configurations, such as introducing different biases via positional encodings and integrating different spatiotemporal hierarchies via different scaling strategies, helps the model compete favorably with SOTA methods in terms of intra- and cross-dataset experiments. Additionally, we identified different behaviors of the model designs between intra- and cross-dataset experiments and summarized several points for choosing the optimal network configurations. We validated our proposed GVT2RPM guidelines by employing three SOTA video transformers, including MViTv2, UniFormer, and Video Swin, and evaluating them under intra- and cross-dataset settings using five public datasets. We highlight that our GVT2RPM is general to different transformer architectures and robust to different datasets.

We identified three limitations in this work. Firstly, we did not analyze the influence of patients’ skin tone on performance, which adds complexities to the algorithm to be robust due to varied reflections on the skin tone. We will explore color normalization and video synthesis techniques to solve this problem. Secondly, we chose to use a small version of the video transformer, which contains relatively small parameters in their family. The small version yields inferior results compared to its larger counterpart, but it is more efficient and easy to train. As part of our future work, we will evaluate if our guidelines are applicable when the model is scaled up. Lastly, the optimal configurations in our experiments were selected manually. However, this can potentially be automated by e.g., following steps as in nnUNet \cite{ref53}, which demonstrated that model configurations can be selected automatically based on interdependent rules and empirical decisions.

%
\bibliographystyle{ACM-Reference-Format}
\bibliography{main}
\end{document}


\title{Supplementary Materials for GVT2RPM}


\author{Hao Wang}
\email{hao.wang2@sydney.edu.au}
\orcid{0000-0002-6235-8425}
\affiliation{%
  \institution{The University of Sydney}
  \city{Sydney}
  \state{NSW}
  \country{Australia}
  \postcode{2006}
}

\author{Euijoon Ahn}
\email{euijoon.ahn@jcu.edu.au}
\orcid{0000-0001-7027-067X}
\affiliation{%
  \institution{James Cook University}
  \city{Cairns}
  \state{QLD}
  \country{Australia}
  \postcode{4870}
}

\author{Jinman Kim}
\email{jinman.kim@sydney.edu.au}
\affiliation{%
  \institution{The University of Sydney}
  \city{Sydney}
  \state{NSW}
  \country{Australia}
  \postcode{2006}
}

\begin{teaserfigure}
  \caption{Details of scaling strategies.}
  \includegraphics[width=\textwidth]{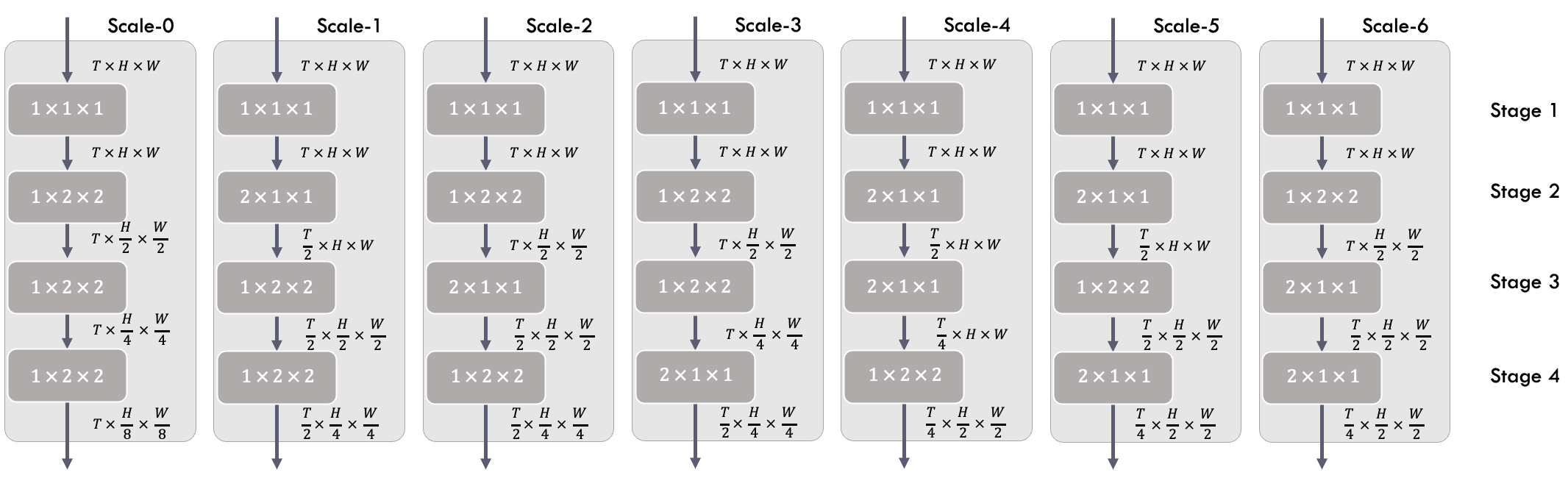}  
  \label{fig:1}
\end{teaserfigure}


\maketitle

\section{Details of Scaling Strategies}

The key idea of modern hybrid video transformers is to integrate multiscale feature hierarchies in convolutional neural networks (CNNs) with the transformer model. With prior knowledge about remote photoplethysmography (rPPG) signals, algorithms are required to extract dense temporal signals from facial videos with more frames than general video analysis. Therefore, finding optimal scaling strategies in general video transformer (GVT) to adjust multiscale hierarchies is important for rPPG signal extraction. In addition to the default scaling strategy Scale-0, we defined 6 scaling strategies to implement different spatiotemporal hierarchies for GVTs, as shown in Figure~\ref{fig:1}.

\section{Details of GVT2RPM-MViT-optimal Configurations}
We demonstrate the optimal configurations for adapting MViTv2 \cite{ref38} to remote physiological measurement (RPM), i.e., GVT2RPM-MViT-optimal, in Table\ref{tab:1}.

\begin{table}[h!]
\caption{Optimal configurations for GVT2RPM-MViT.}
\label{tab:1}
\resizebox{\linewidth}{!}{%
\begin{tabular}{@{}llcccc@{}}
\toprule
\multicolumn{1}{c}{\begin{tabular}[c]{@{}c@{}}Training\\ Set\end{tabular}} & \multicolumn{1}{c}{\begin{tabular}[c]{@{}c@{}}Testing\\ Set\end{tabular}} & \begin{tabular}[c]{@{}c@{}}Frame\\ Format\end{tabular} & \begin{tabular}[c]{@{}c@{}}Signal\\ Normalization\end{tabular} & \begin{tabular}[c]{@{}c@{}}Positional\\ Encoding\end{tabular} & \begin{tabular}[c]{@{}c@{}}Scaling\\ Strategy\end{tabular} \\ \midrule
\multicolumn{2}{c}{MMPD-simple} & DiffNorm & Yes & REL & Scale-2 \\
\multicolumn{2}{c}{MMPD} & DiffNorm & No & REL & Scale-6 \\
\multicolumn{2}{c}{RLAP} & DiffNorm & Yes & REL & Scale-2 \\
\multicolumn{2}{c}{UBFC-rPPG} & DiffNorm & Yes & ABS & Scale-5 \\ \midrule
\multicolumn{6}{l}{} \\
\multirow{3}{*}{\begin{tabular}[c]{@{}l@{}}MMPD-\\ simple\end{tabular}} & UBFC-rPPG & DiffNorm & No & CPE & Scale-0 \\
 & UBFC-Phys & DiffNorm & No & ABS & Scale-0 \\
 & RLAP & DiffNorm & No & CPE & Scale-2 \\ \midrule
\multicolumn{6}{l}{} \\
\multirow{3}{*}{MMPD} & UBFC-rPPG & DiffNorm & No & CPE & Scale-0 \\
 & UBFC-Phys & DiffNorm & No & ABS & Scale-0 \\
 & RLAP & DiffNorm & No & ABS & Scale-0 \\ \midrule
\multicolumn{6}{l}{} \\
\multirow{4}{*}{RLAP} & MMPD-simple & DiffNorm & Yes & ABS & Scale-0 \\
 & MMPD & DiffNorm & Yes & REL & Scale-2 \\
 & UBFC-rPPG & DiffNorm & Yes & CPE & Scale-3 \\
 & UBFC-Phys & DiffNorm & Yes & ABS & Scale-3 \\ \midrule
\multicolumn{6}{l}{} \\
\multirow{4}{*}{\begin{tabular}[c]{@{}l@{}}UBFC-\\ rPPG\end{tabular}} & MMPD-simple & DiffNorm & Yes & REL & Scale-0 \\
 & MMPD & DiffNorm & Yes & REL & Scale-1 \\
 & UBFC-Phys & DiffNorm & No & ABS & Scale-3 \\
 & RLAP & DiffNorm & No & REL & Scale-3 \\ \bottomrule
\end{tabular}%
}
\end{table}

\section{Details of Experiment Results}
We used rPPG-Toolbox \cite{ref44} to evaluated five state-of-the-art (SOTA) methods, including DeepPhys \cite{ref18}, PhysNet \cite{ref20}, TS-CAN \cite{ref52}, EfficientPhys \cite{ref27}, and PhysFormer \cite{ref43}. We compared them with GVT2RPM adapted version of MViTv2 \cite{ref38}, UniFormer \cite{ref24}, and Video Swin \cite{ref23}. In addition to Mean Absolute Error (MAE), we also used metrics of Root Mean Square Error (RMSE) and Pearson Correlation Coefficient ($\rho$).

\begin{figure*}[]
\caption{Intra-dataset experiment results on UBFC-rPPG.}
\label{fig:2}
\begin{tabular}{@{}p{0.05\textwidth}lll@{}}
\toprule
\multicolumn{1}{c}{Methods} & \multicolumn{1}{c}{Fold 0} & \multicolumn{1}{c}{Fold 1} & \multicolumn{1}{c}{Fold 2} \\ \midrule
\begin{tabular}[c]{@{}l@{}}GVT2RPM-\\ MViT-\\ optimal\end{tabular} & 
    \begin{minipage}{.29\textwidth} \includegraphics[width=\linewidth]{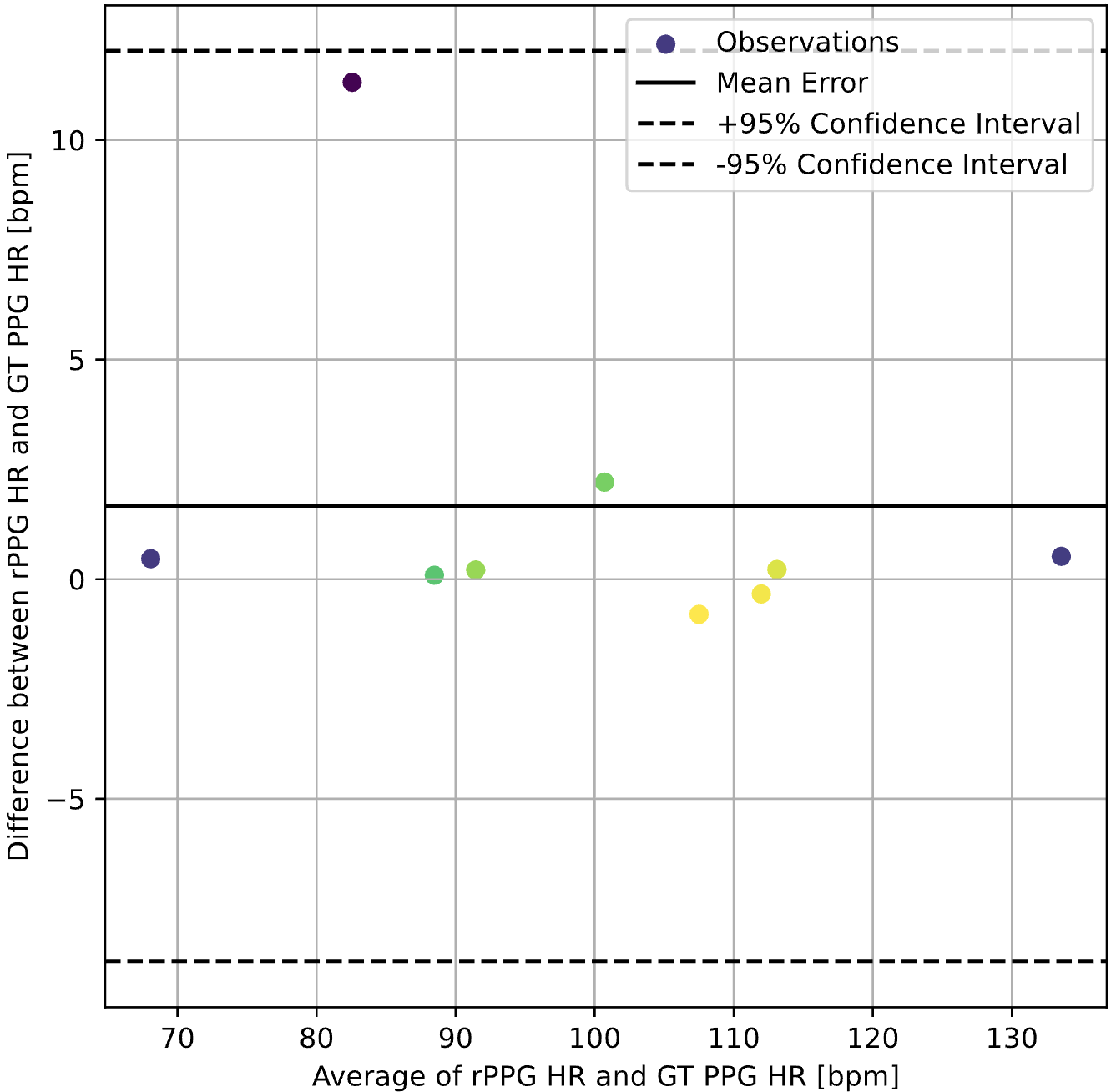} \end{minipage} &
    \begin{minipage}{.29\textwidth} \includegraphics[width=\linewidth]{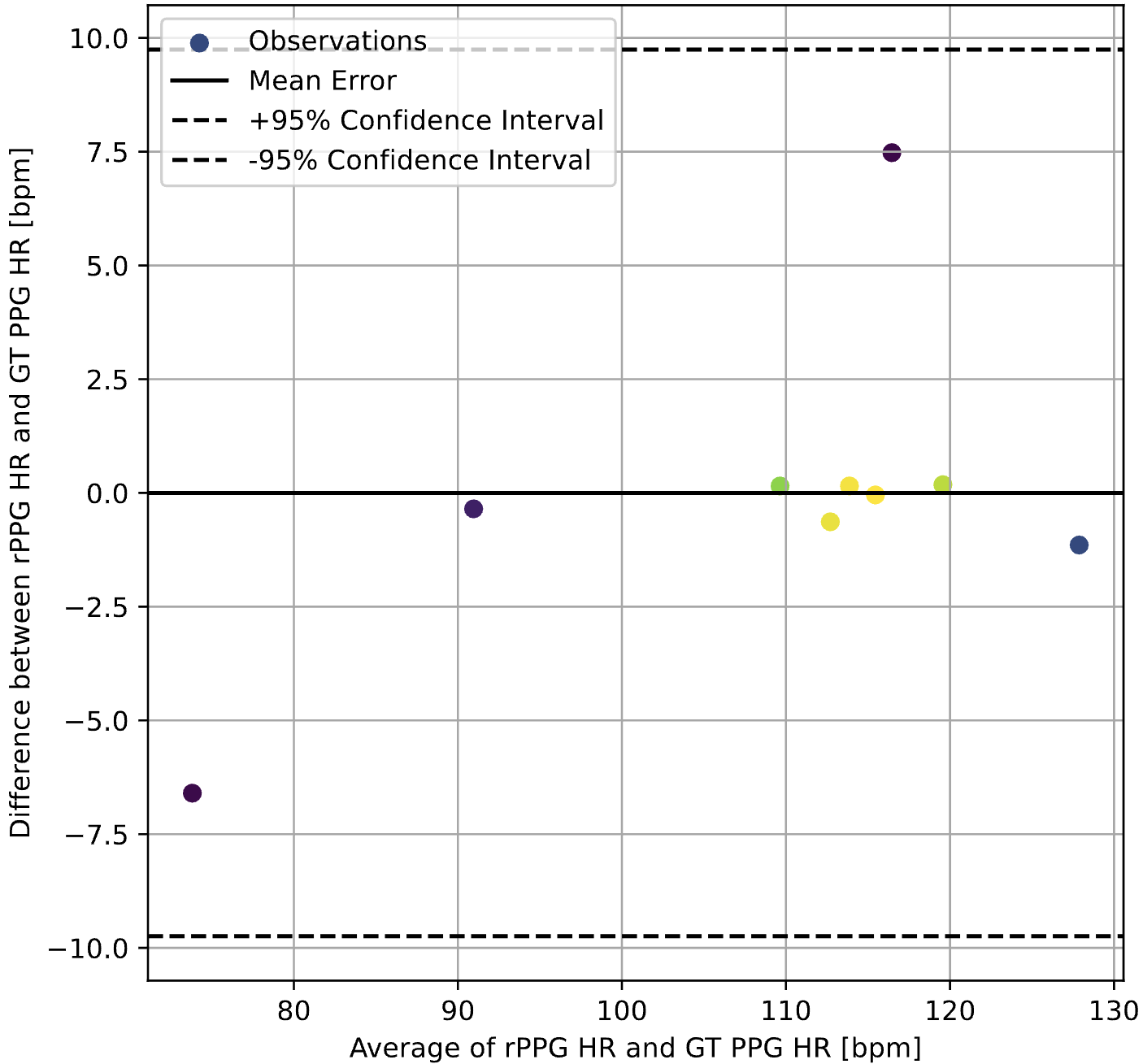} \end{minipage} &
    \begin{minipage}{.29\textwidth} \includegraphics[width=\linewidth]{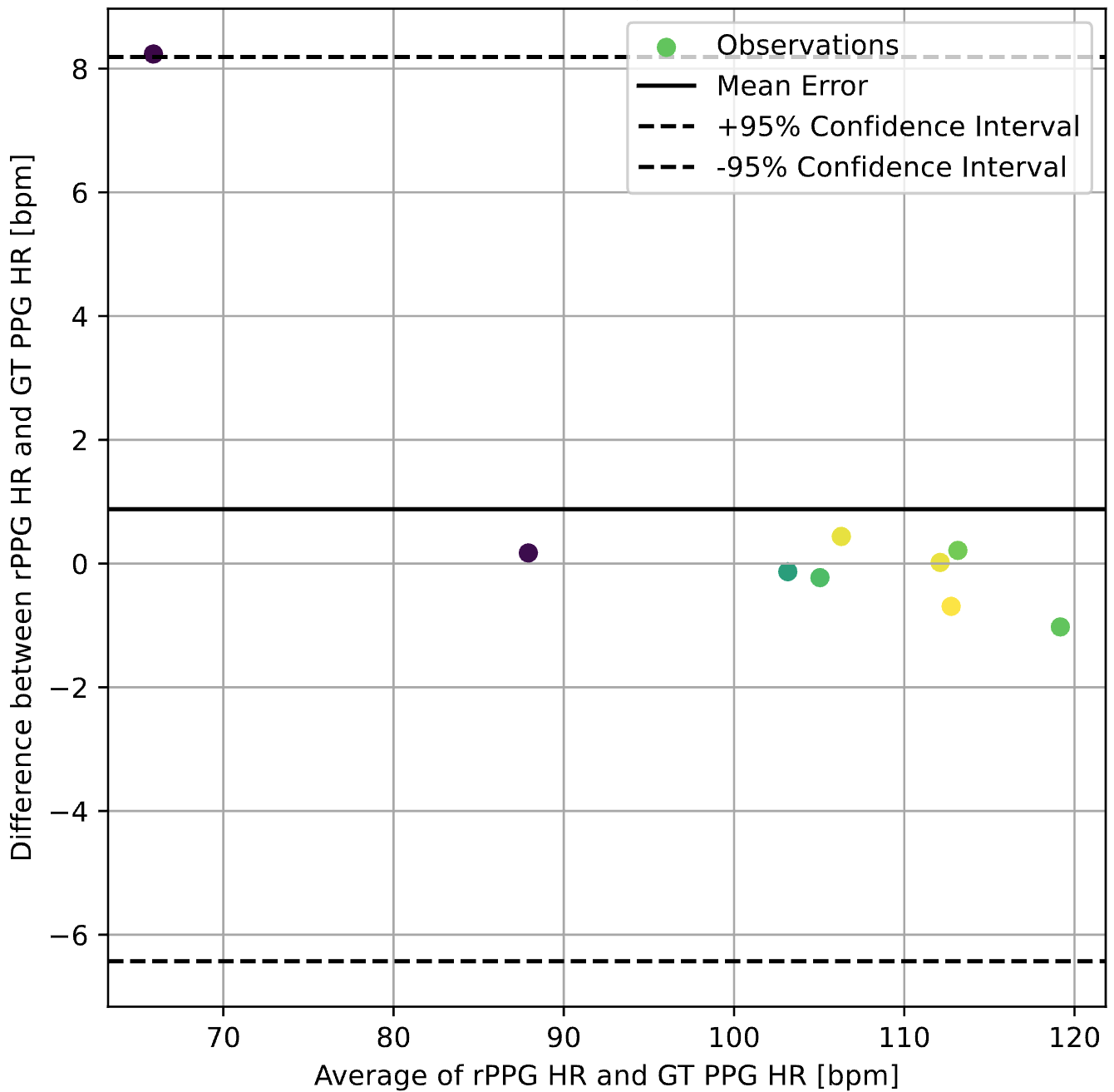} \end{minipage} \\
\begin{tabular}[c]{@{}l@{}}GVT2RPM-\\ UniFormer-\\ optimal\end{tabular} &  
    \begin{minipage}{.29\textwidth} \includegraphics[width=\linewidth]{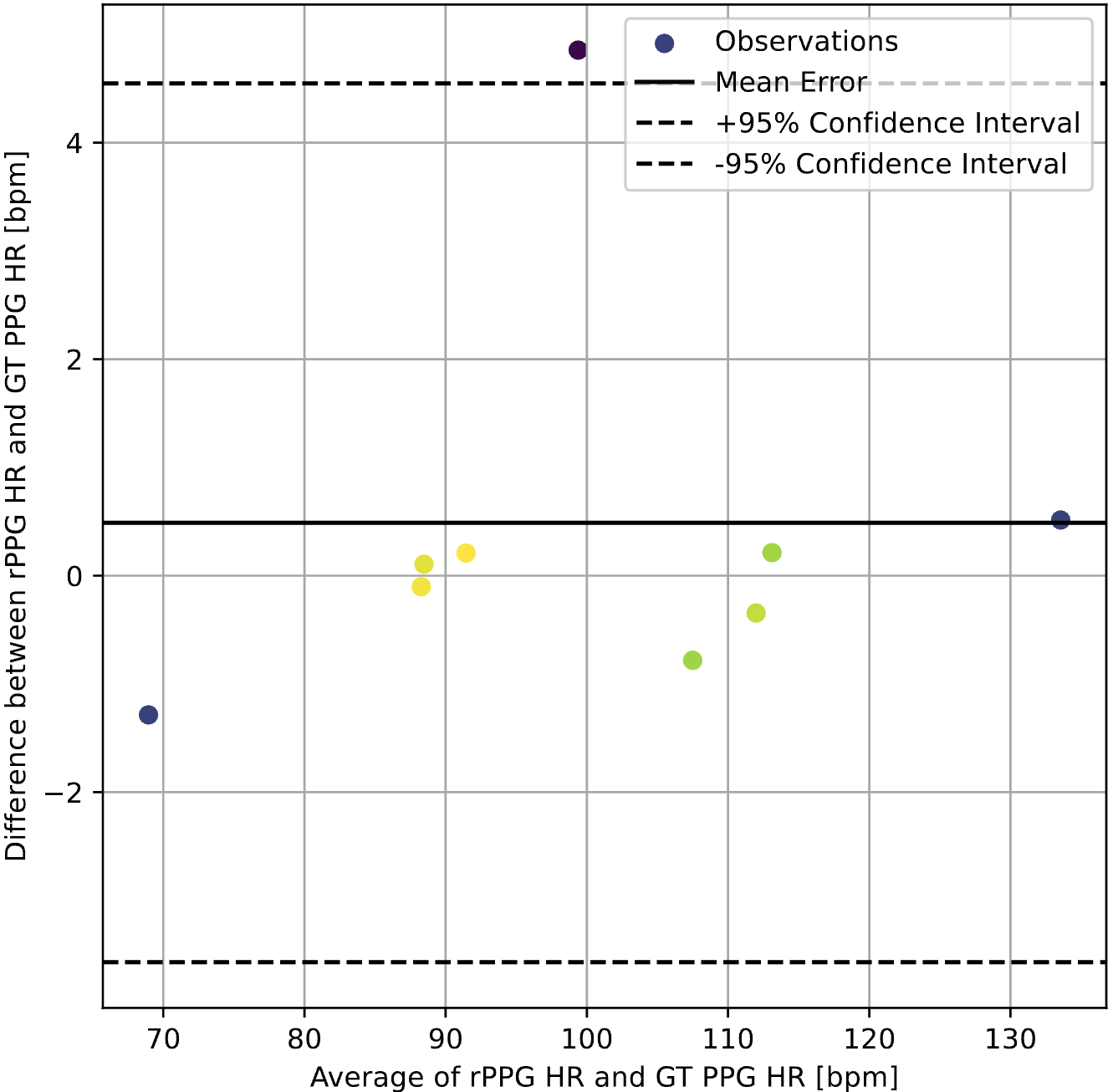} \end{minipage} &
    \begin{minipage}{.29\textwidth} \includegraphics[width=\linewidth]{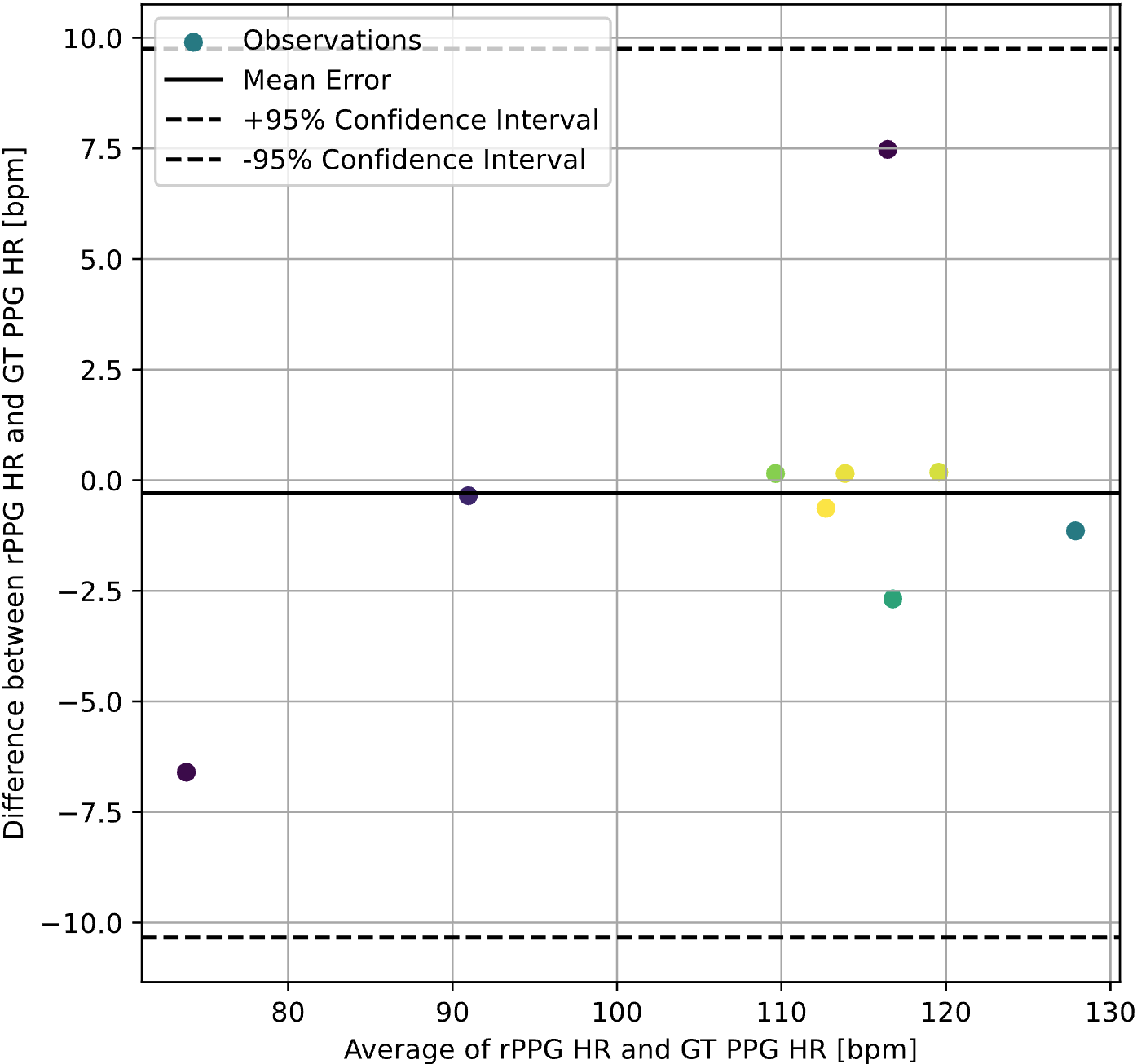} \end{minipage} &
    \begin{minipage}{.29\textwidth} \includegraphics[width=\linewidth]{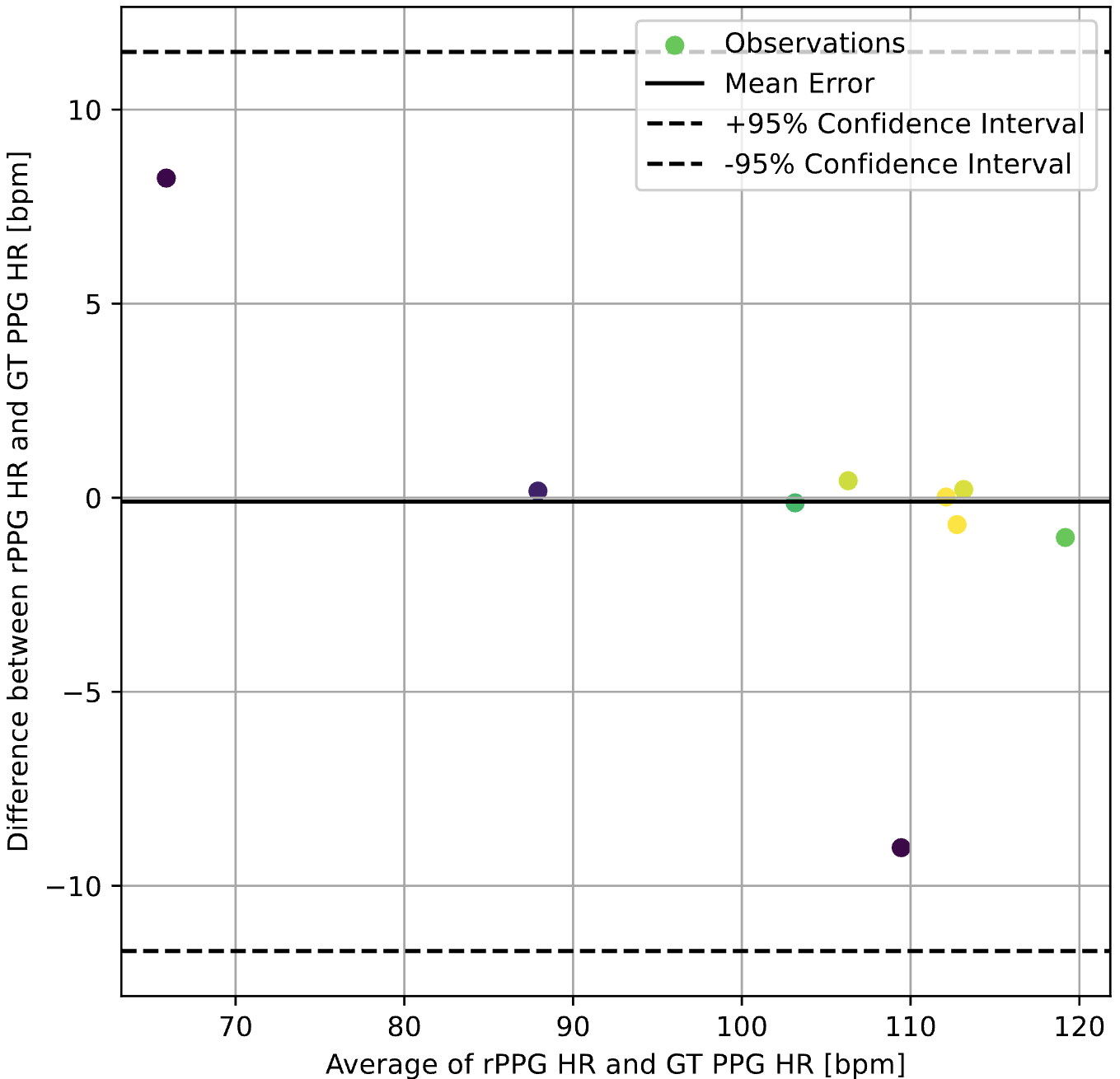} \end{minipage} \\
\begin{tabular}[c]{@{}l@{}}GVT2RPM-\\ Swin-\\ optimal\end{tabular} & 
    \begin{minipage}{.29\textwidth} \includegraphics[width=\linewidth]{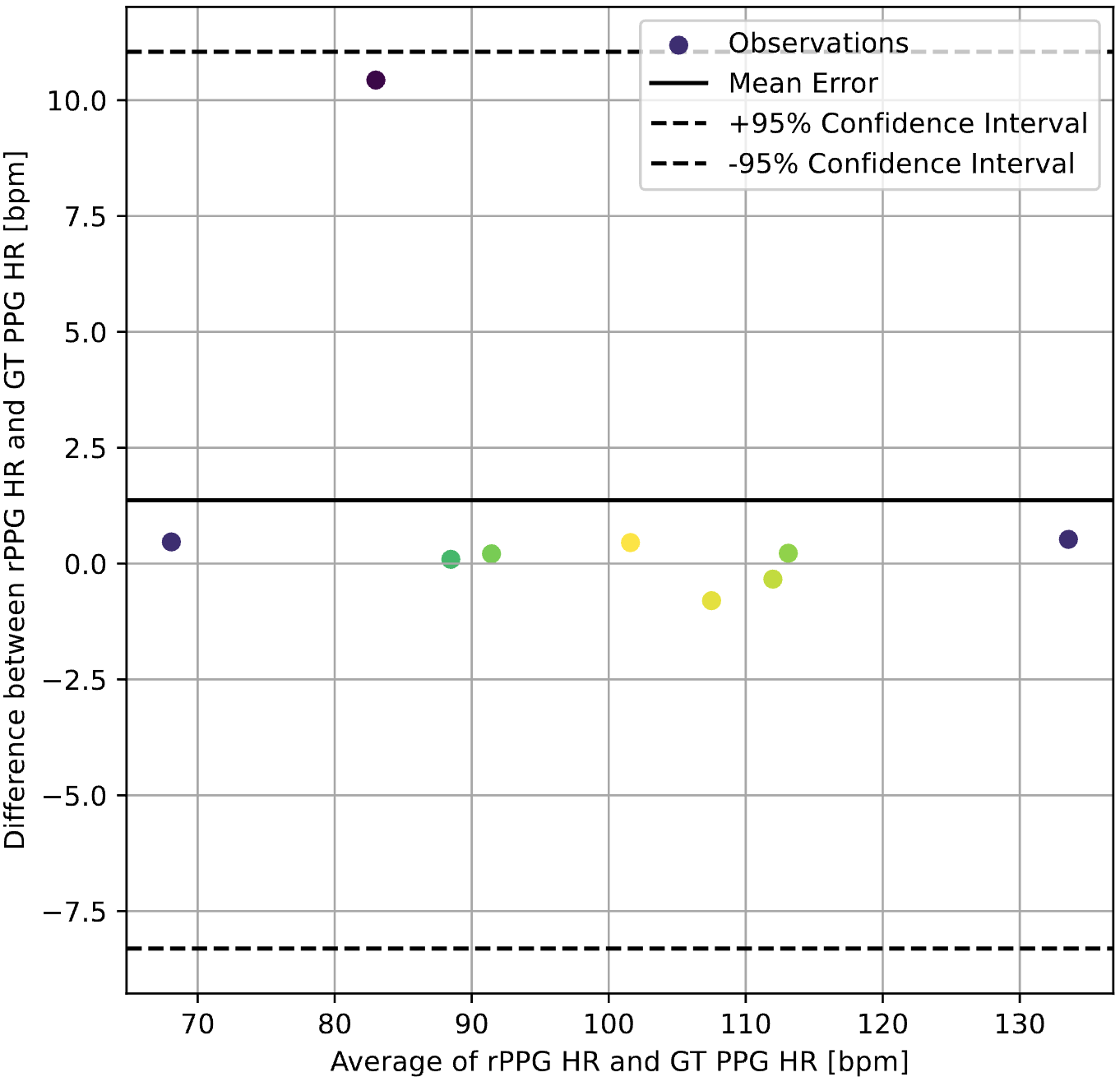} \end{minipage} &
    \begin{minipage}{.29\textwidth} \includegraphics[width=\linewidth]{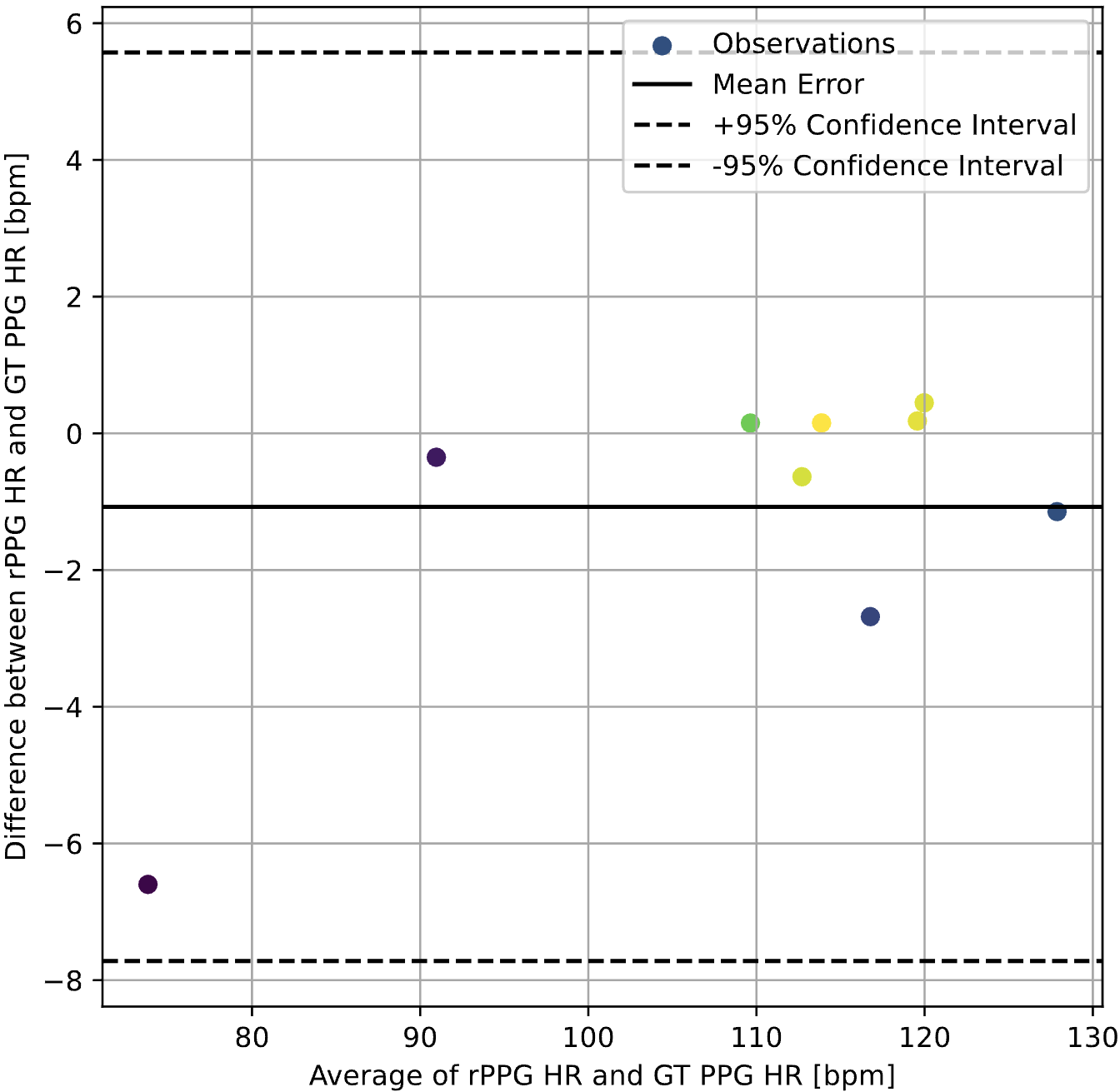} \end{minipage} &
    \begin{minipage}{.29\textwidth} \includegraphics[width=\linewidth]{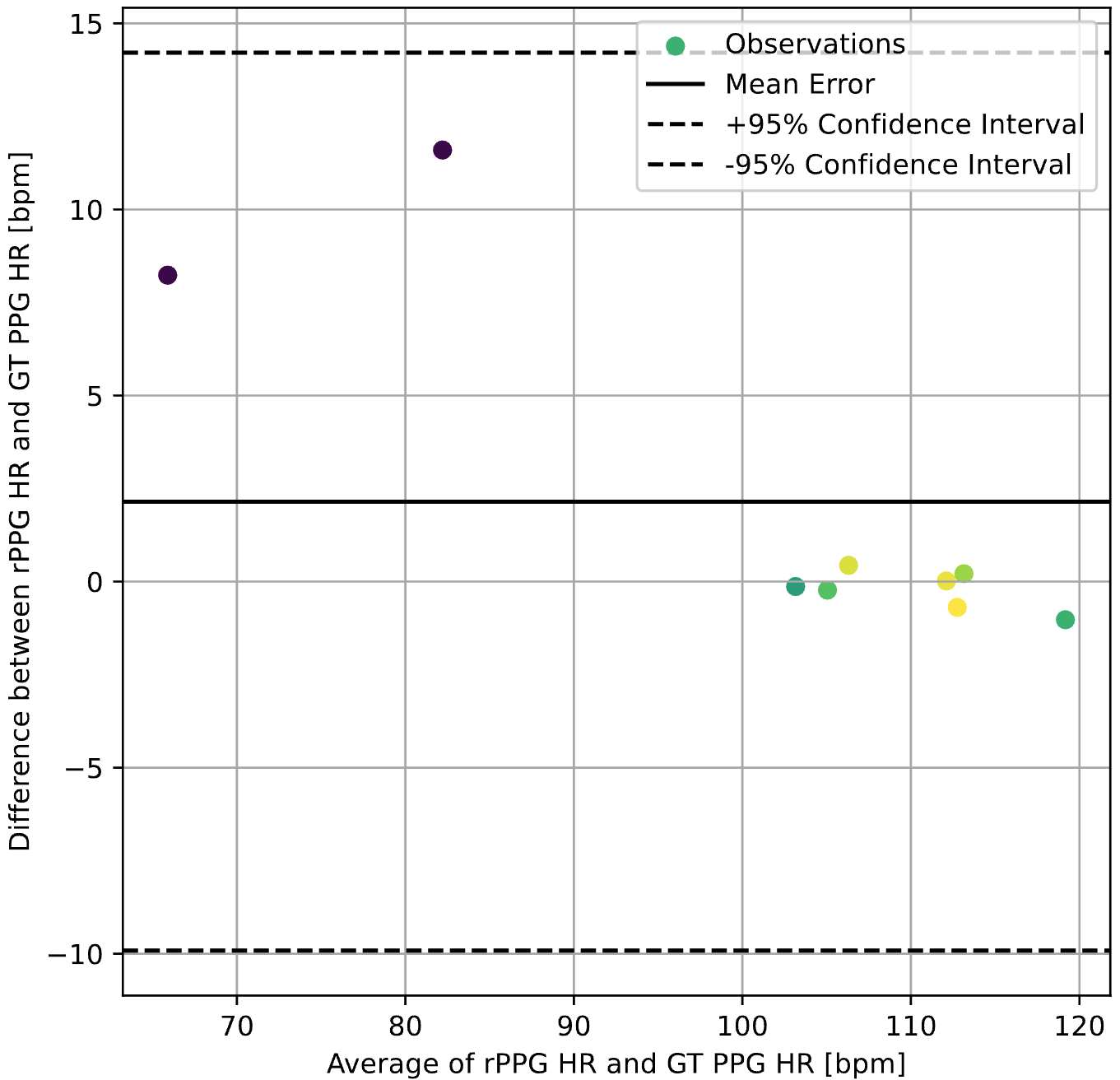} \end{minipage} \\ \bottomrule
\end{tabular}
\end{figure*}

\begin{figure*}[]
\caption{Intra-dataset experiment results on MMPD-simple.}
\label{fig:3}
\begin{tabular}{@{}p{0.05\textwidth}lll@{}}
\toprule
\multicolumn{1}{c}{Methods} & \multicolumn{1}{c}{Fold 0} & \multicolumn{1}{c}{Fold 1} & \multicolumn{1}{c}{Fold 2} \\ \midrule
\begin{tabular}[c]{@{}l@{}}GVT2RPM-\\ MViT-\\ optimal\end{tabular} & 
    \begin{minipage}{.29\textwidth} \includegraphics[width=\linewidth]{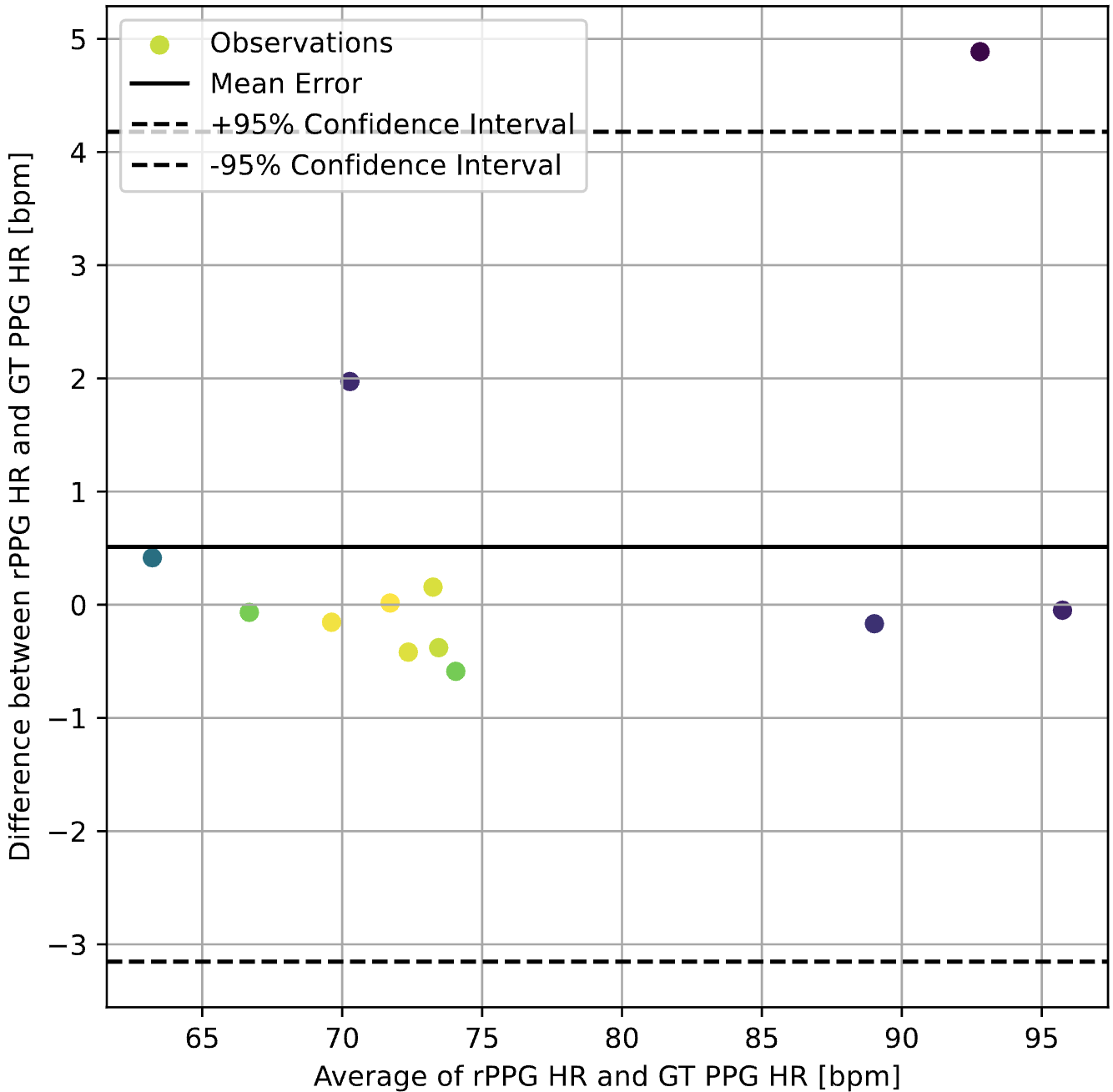} \end{minipage} &
    \begin{minipage}{.29\textwidth} \includegraphics[width=\linewidth]{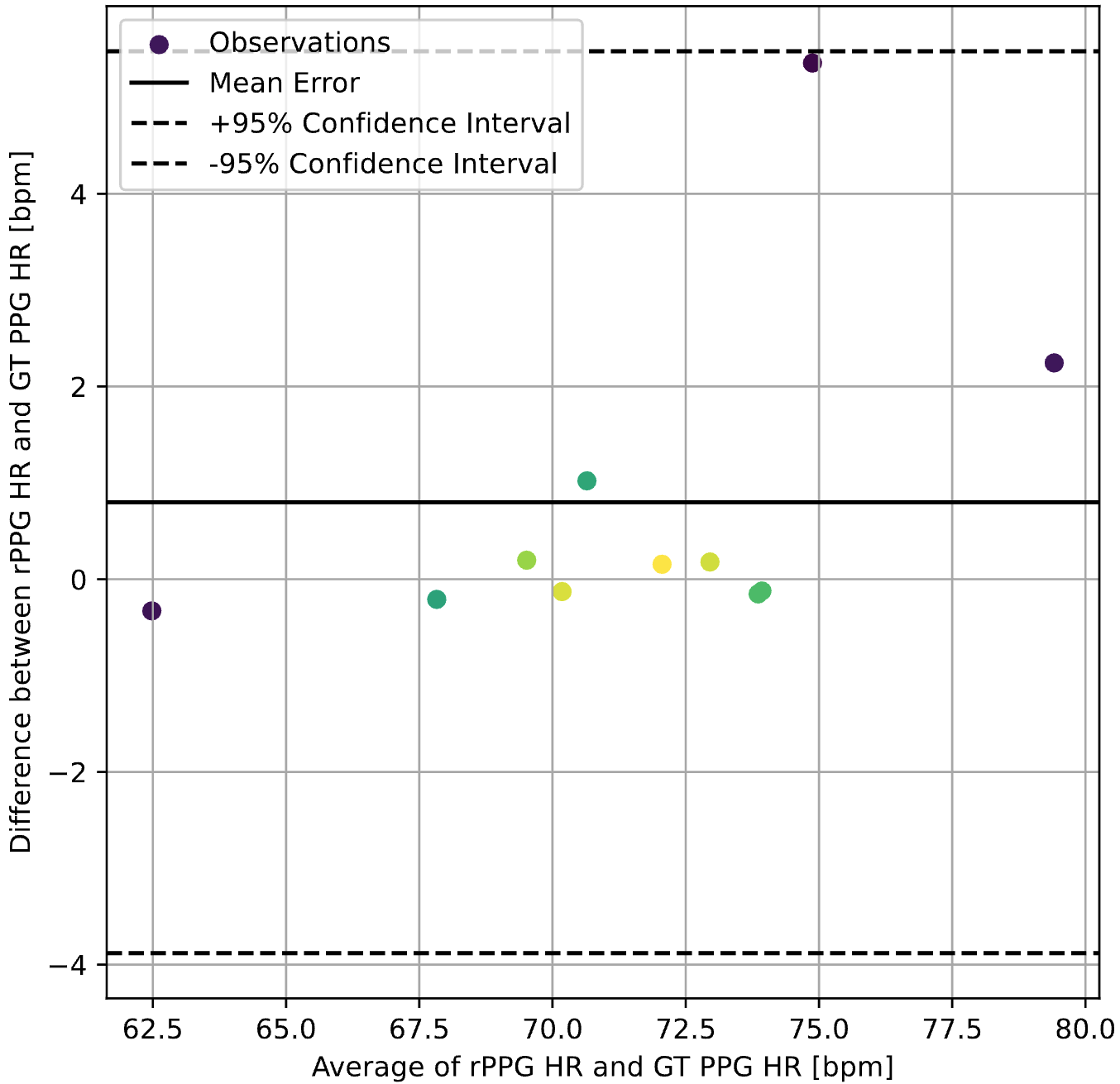} \end{minipage} &
    \begin{minipage}{.29\textwidth} \includegraphics[width=\linewidth]{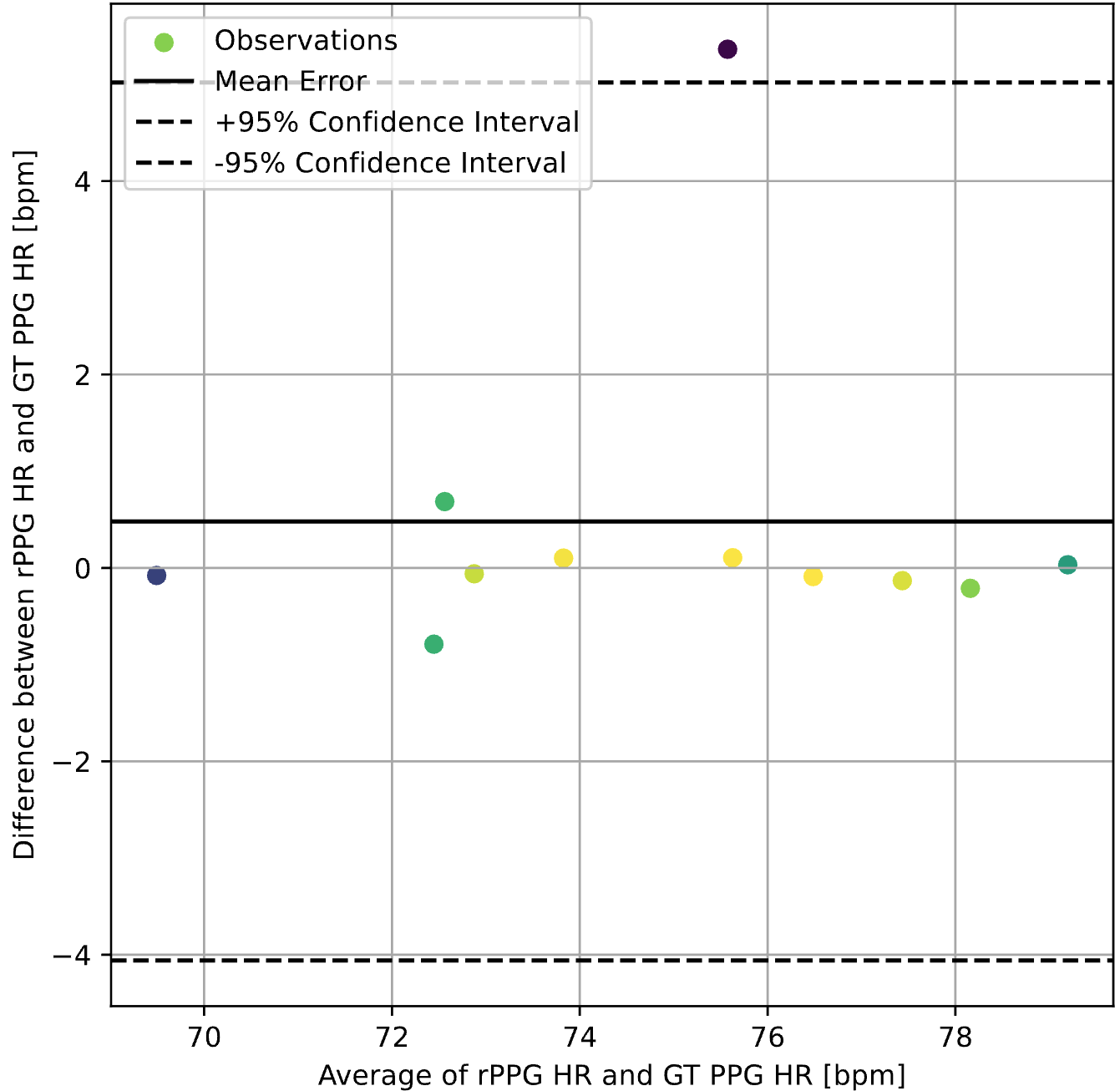} \end{minipage} \\
\begin{tabular}[c]{@{}l@{}}GVT2RPM-\\ UniFormer-\\ optimal\end{tabular} &  
    \begin{minipage}{.29\textwidth} \includegraphics[width=\linewidth]{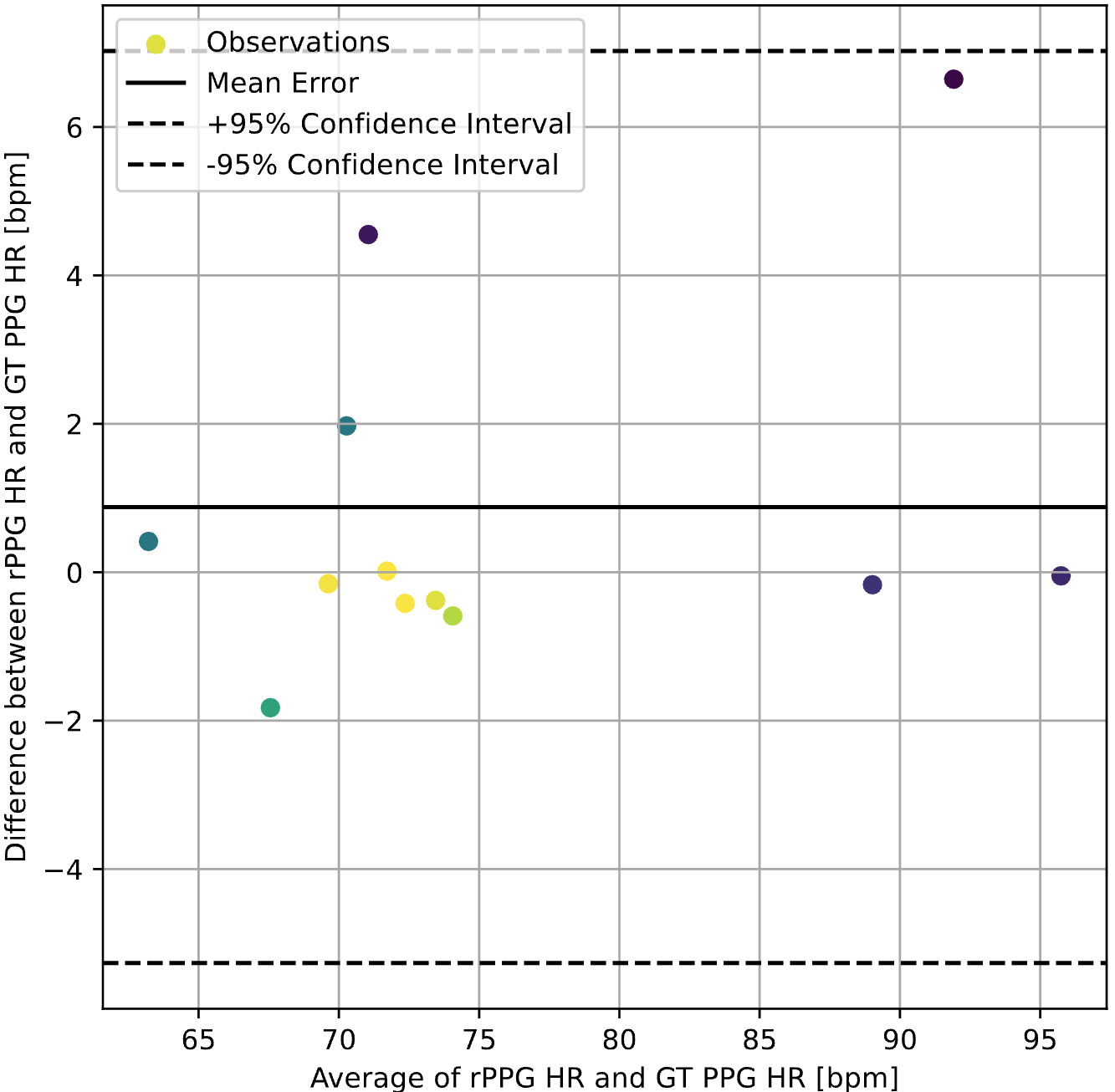} \end{minipage} &
    \begin{minipage}{.29\textwidth} \includegraphics[width=\linewidth]{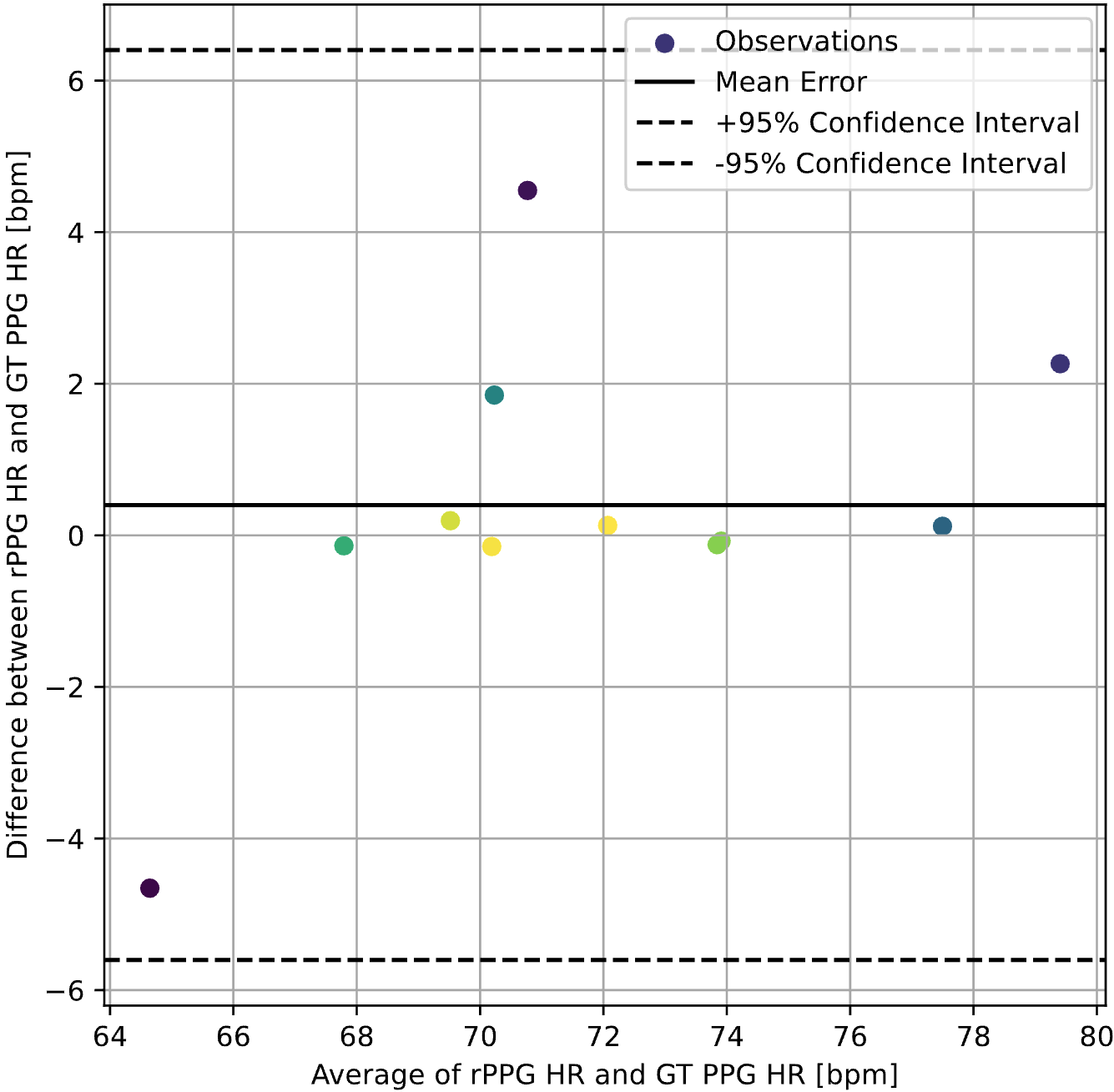} \end{minipage} &
    \begin{minipage}{.29\textwidth} \includegraphics[width=\linewidth]{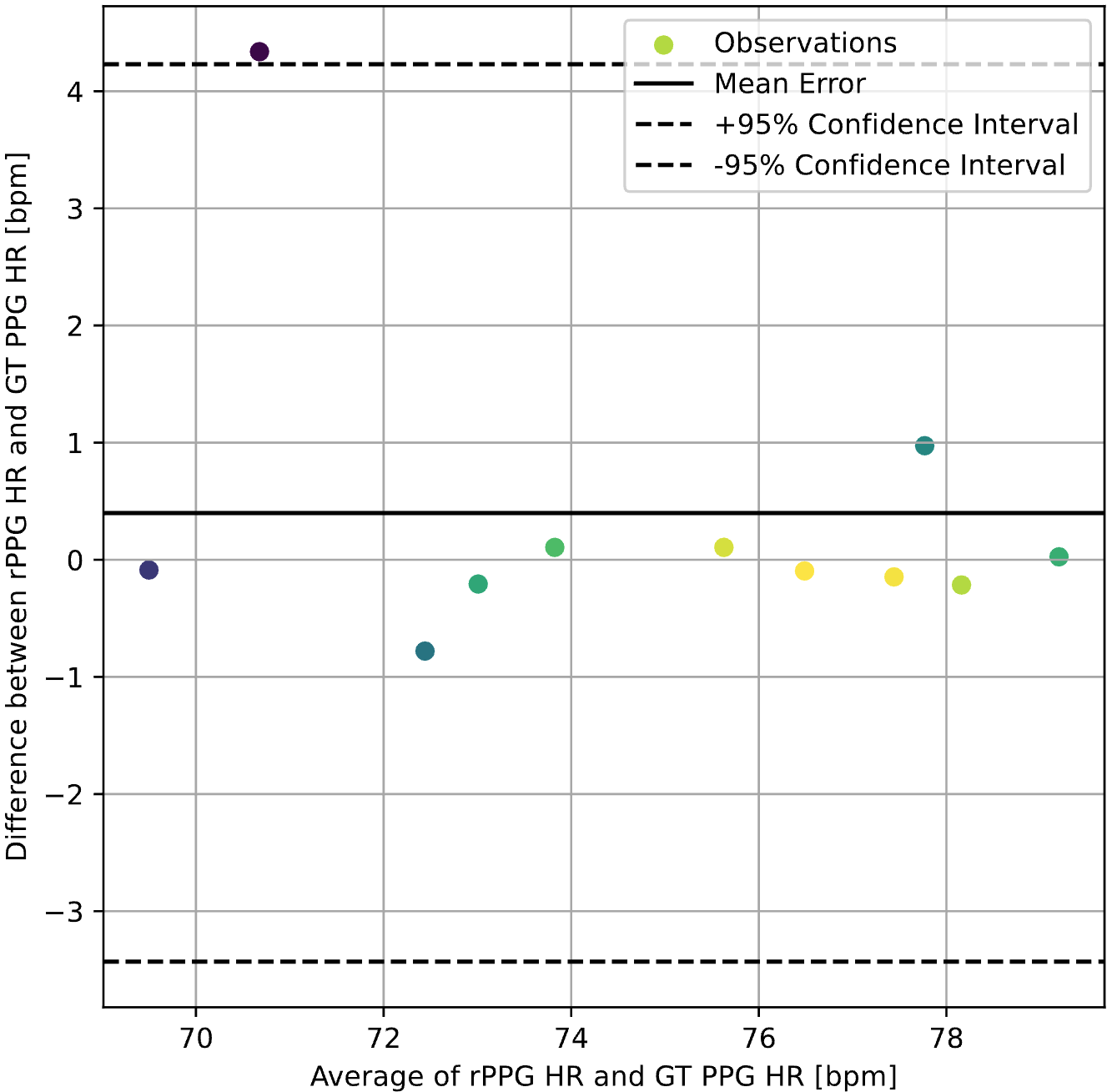} \end{minipage} \\
\begin{tabular}[c]{@{}l@{}}GVT2RPM-\\ Swin-\\ optimal\end{tabular} & 
    \begin{minipage}{.29\textwidth} \includegraphics[width=\linewidth]{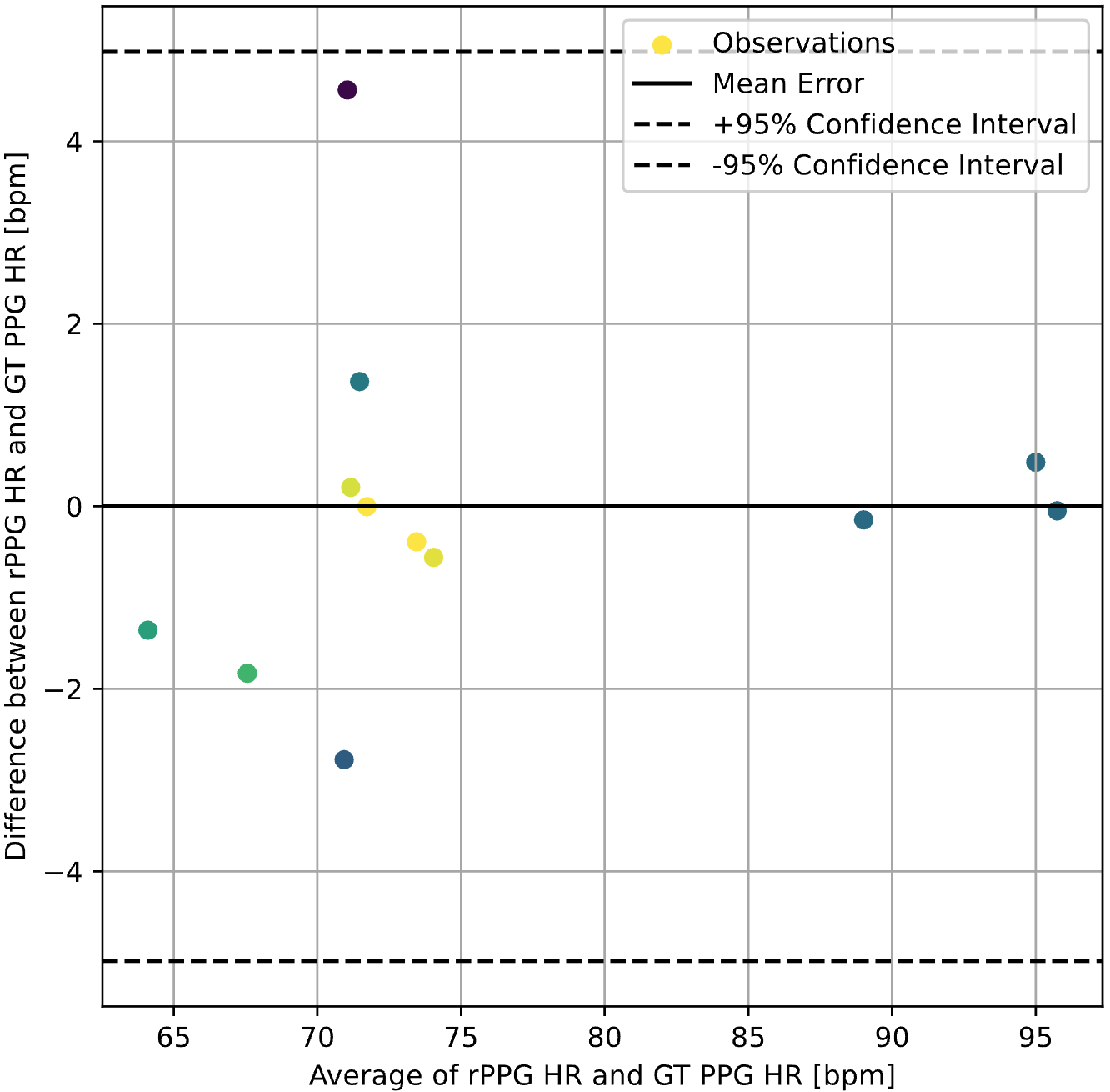} \end{minipage} &
    \begin{minipage}{.29\textwidth} \includegraphics[width=\linewidth]{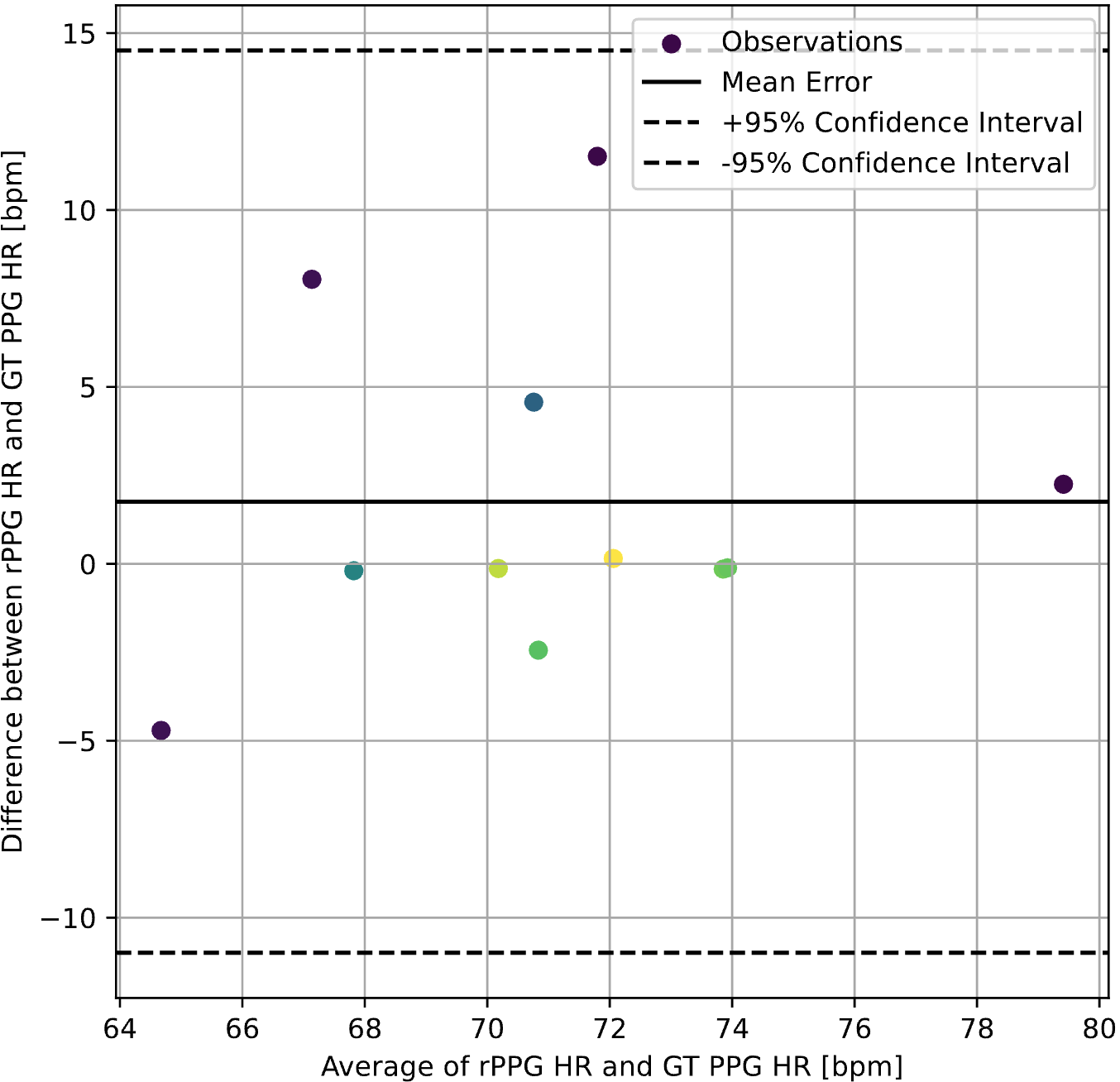} \end{minipage} &
    \begin{minipage}{.29\textwidth} \includegraphics[width=\linewidth]{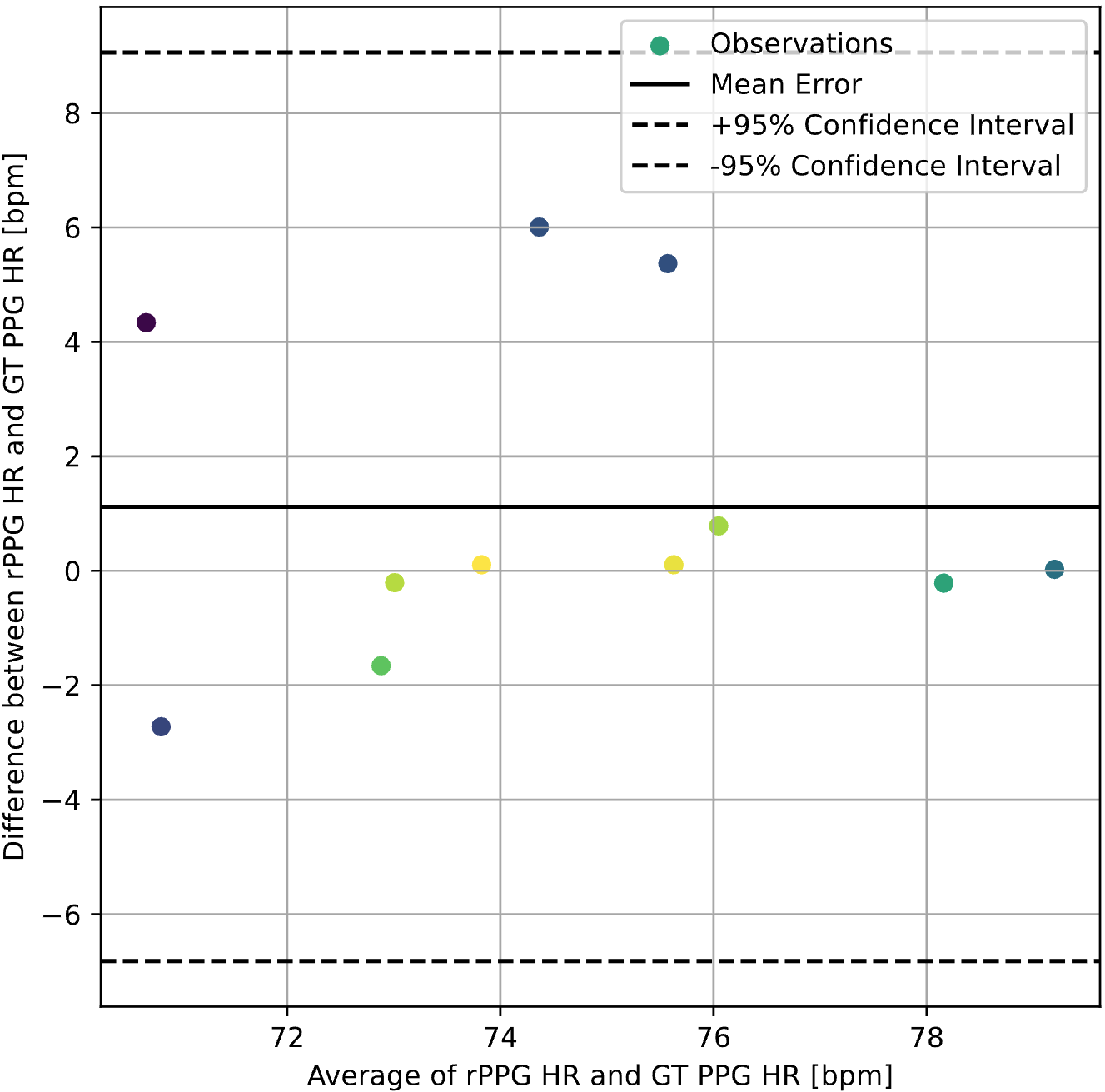} \end{minipage} \\ \bottomrule
\end{tabular}
\end{figure*}

\begin{figure*}[]
\caption{Intra-dataset experiment results on MMPD.}
\label{fig:4}
\begin{tabular}{@{}p{0.05\textwidth}lll@{}}
\toprule
\multicolumn{1}{c}{Methods} & \multicolumn{1}{c}{Fold 0} & \multicolumn{1}{c}{Fold 1} & \multicolumn{1}{c}{Fold 2} \\ \midrule
\begin{tabular}[c]{@{}l@{}}GVT2RPM-\\ MViT-\\ optimal\end{tabular} & 
    \begin{minipage}{.29\textwidth} \includegraphics[width=\linewidth]{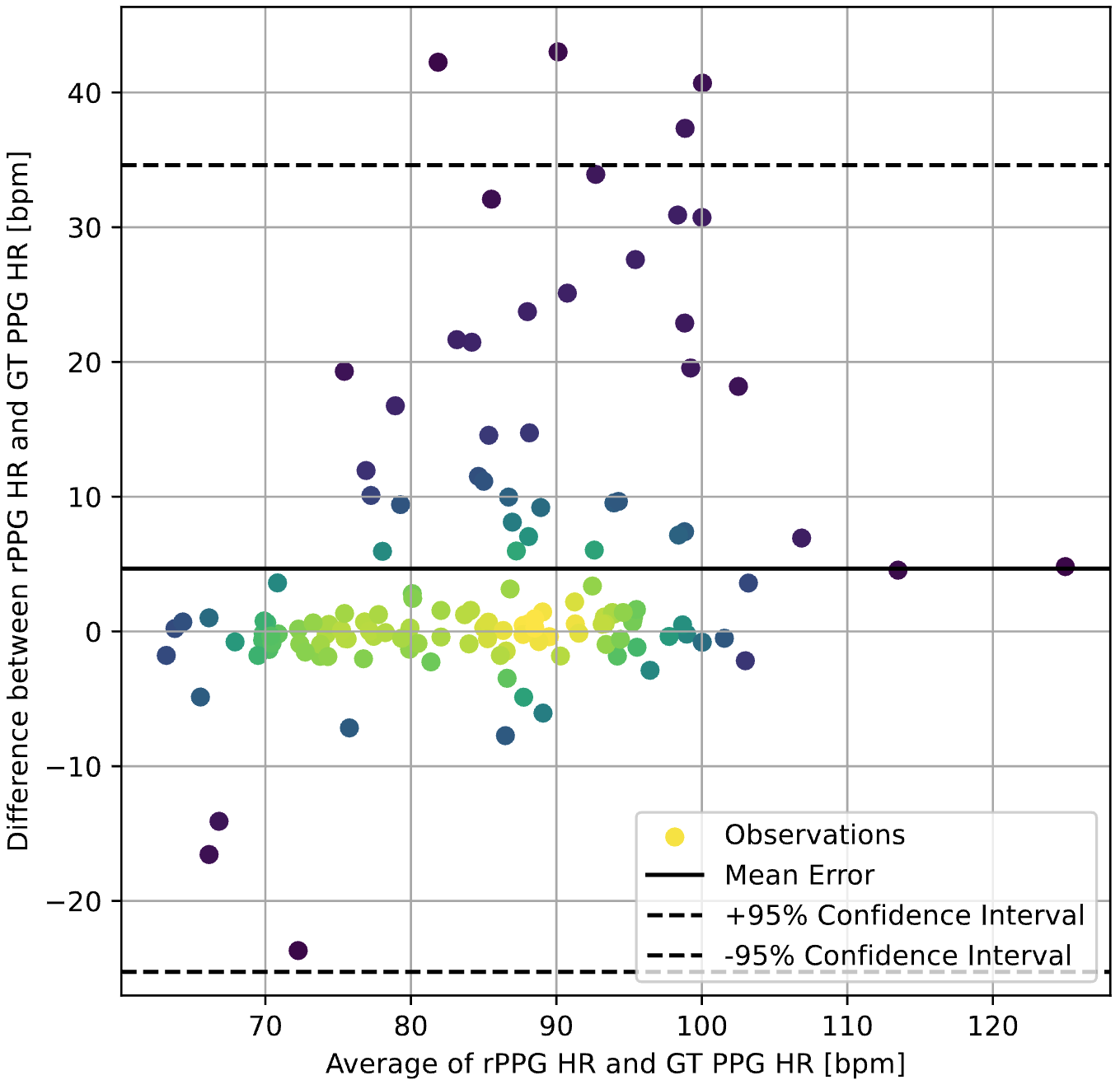} \end{minipage} &
    \begin{minipage}{.29\textwidth} \includegraphics[width=\linewidth]{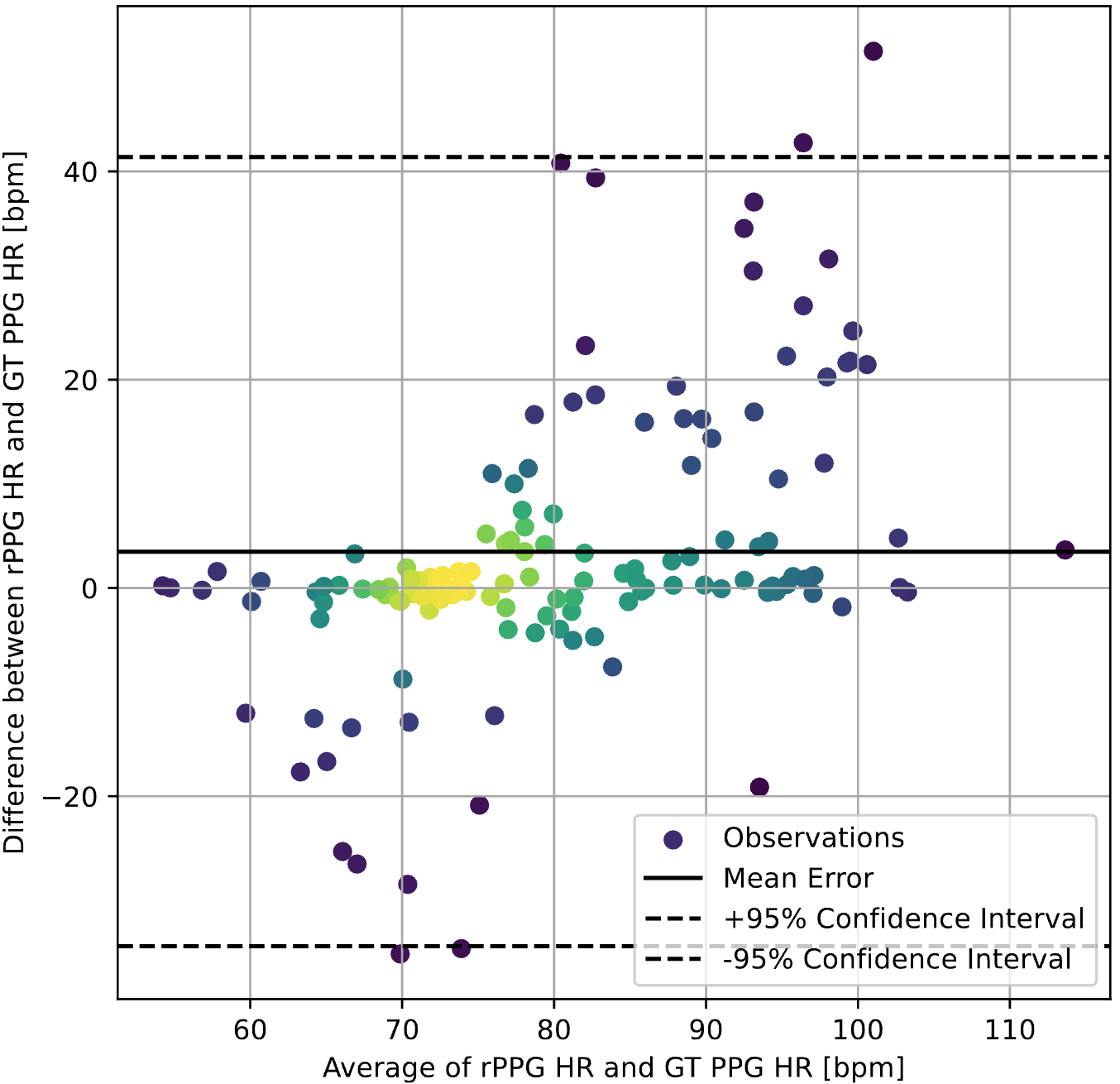} \end{minipage} &
    \begin{minipage}{.29\textwidth} \includegraphics[width=\linewidth]{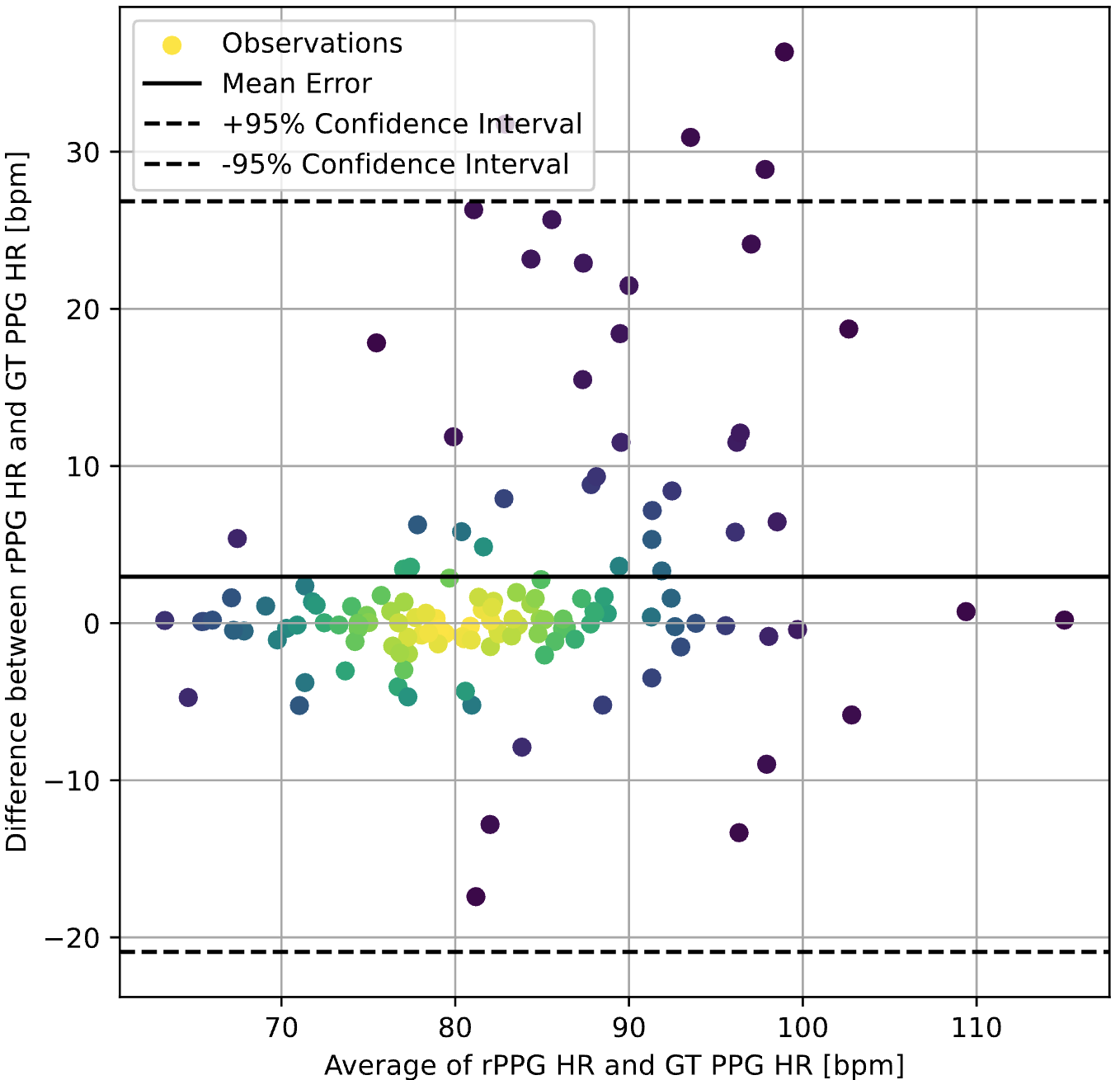} \end{minipage} \\
\begin{tabular}[c]{@{}l@{}}GVT2RPM-\\ UniFormer-\\ optimal\end{tabular} &  
    \begin{minipage}{.29\textwidth} \includegraphics[width=\linewidth]{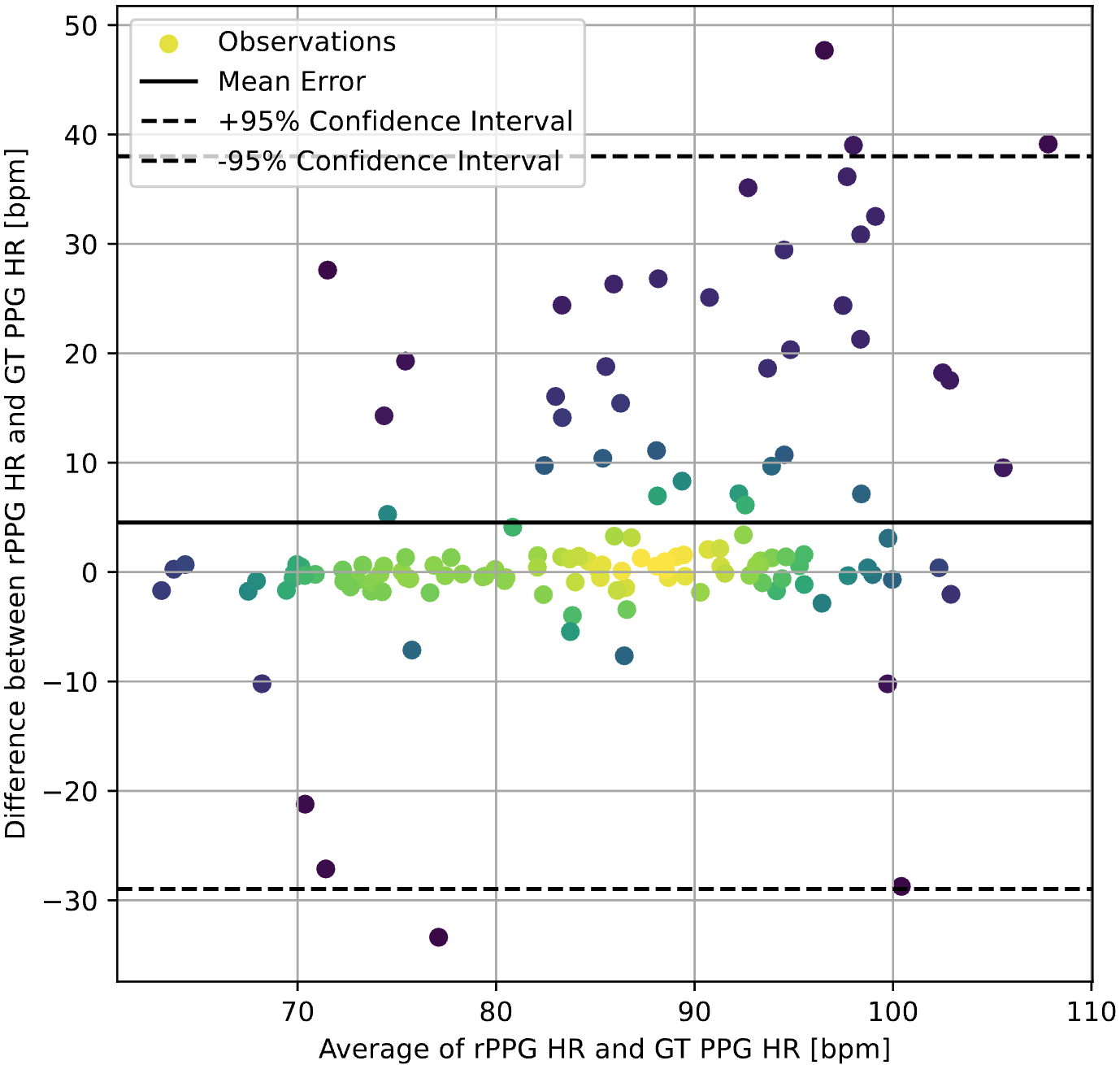} \end{minipage} &
    \begin{minipage}{.29\textwidth} \includegraphics[width=\linewidth]{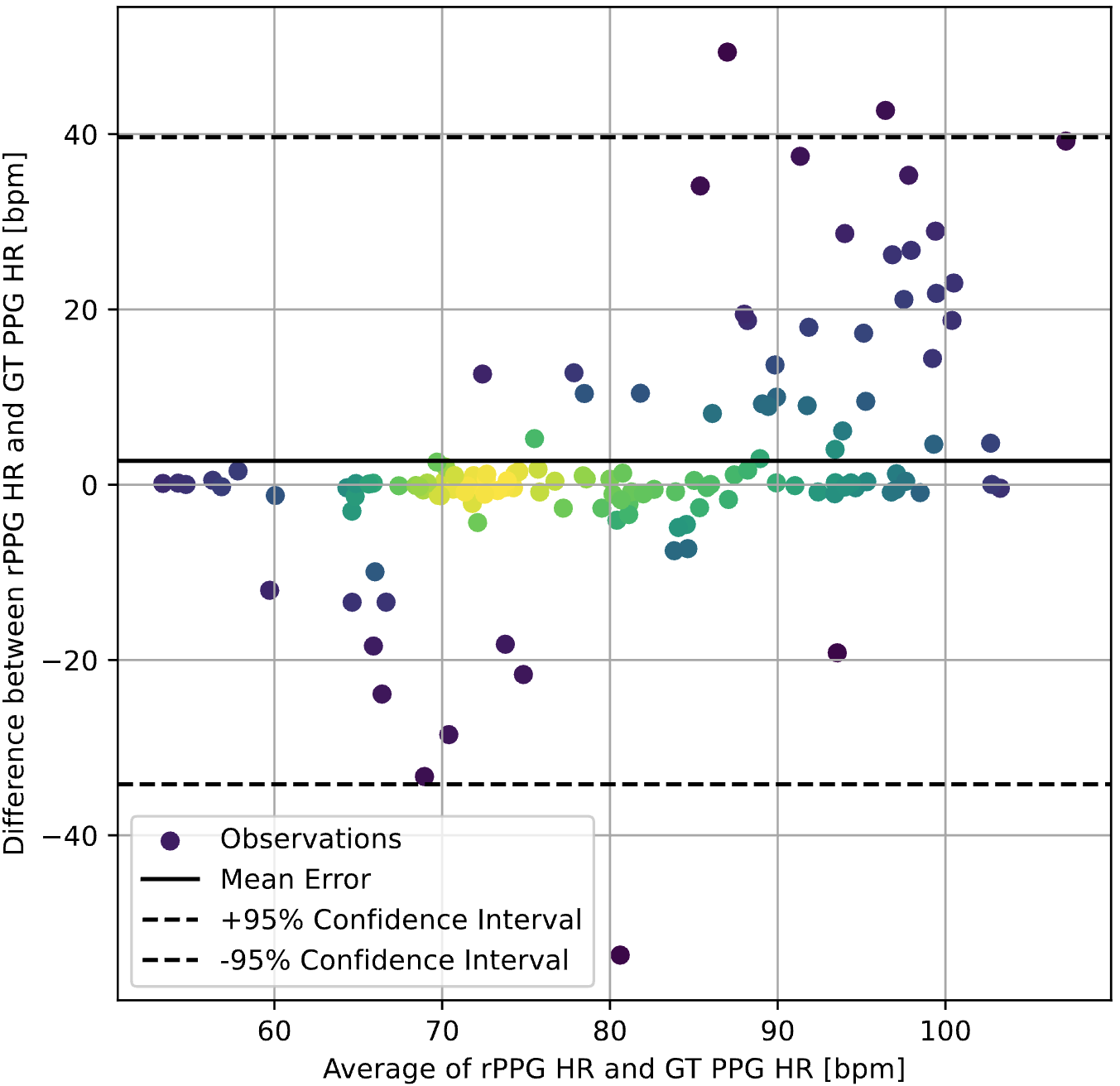} \end{minipage} &
    \begin{minipage}{.29\textwidth} \includegraphics[width=\linewidth]{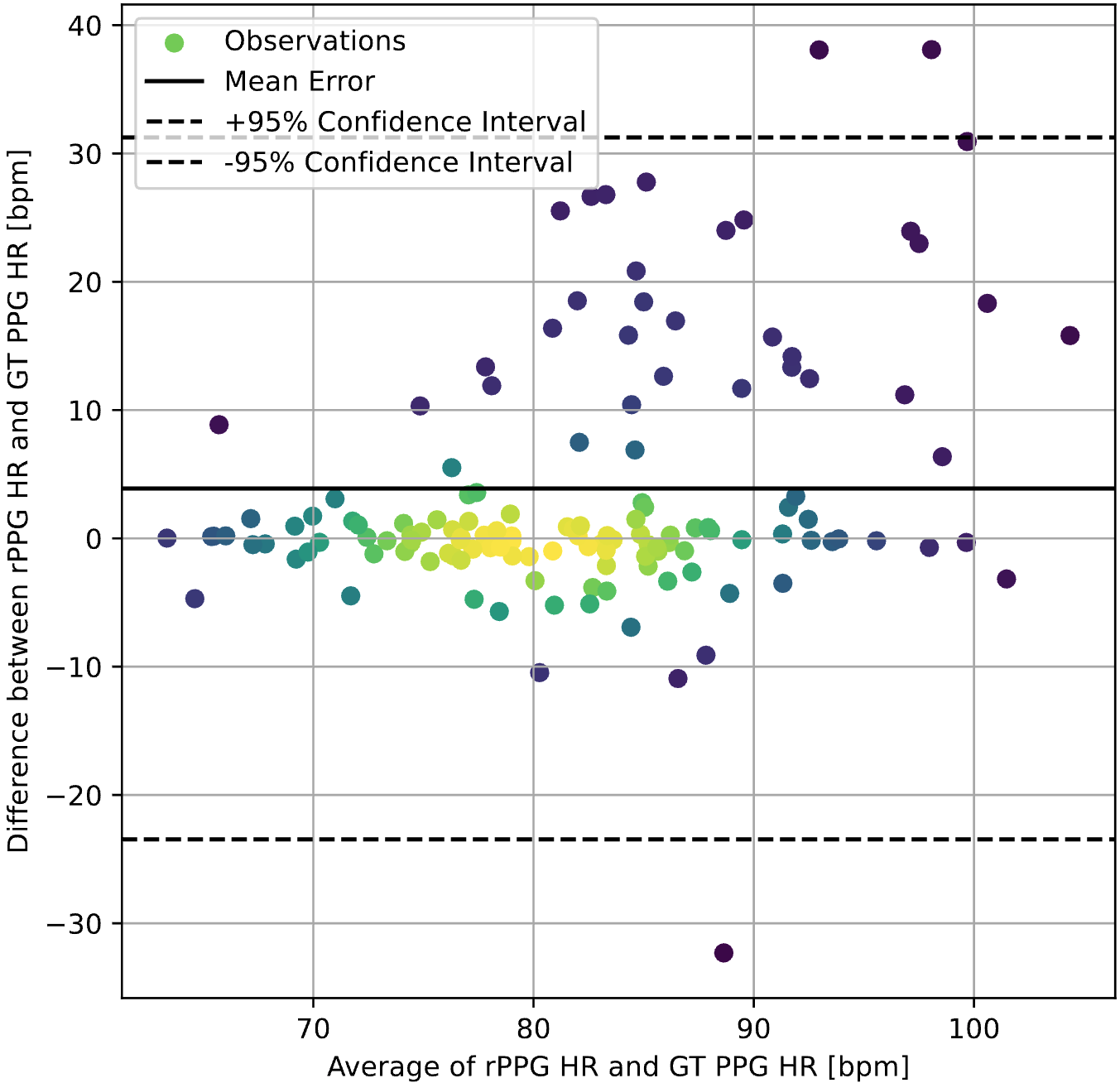} \end{minipage} \\
\begin{tabular}[c]{@{}l@{}}GVT2RPM-\\ Swin-\\ optimal\end{tabular} & 
    \begin{minipage}{.29\textwidth} \includegraphics[width=\linewidth]{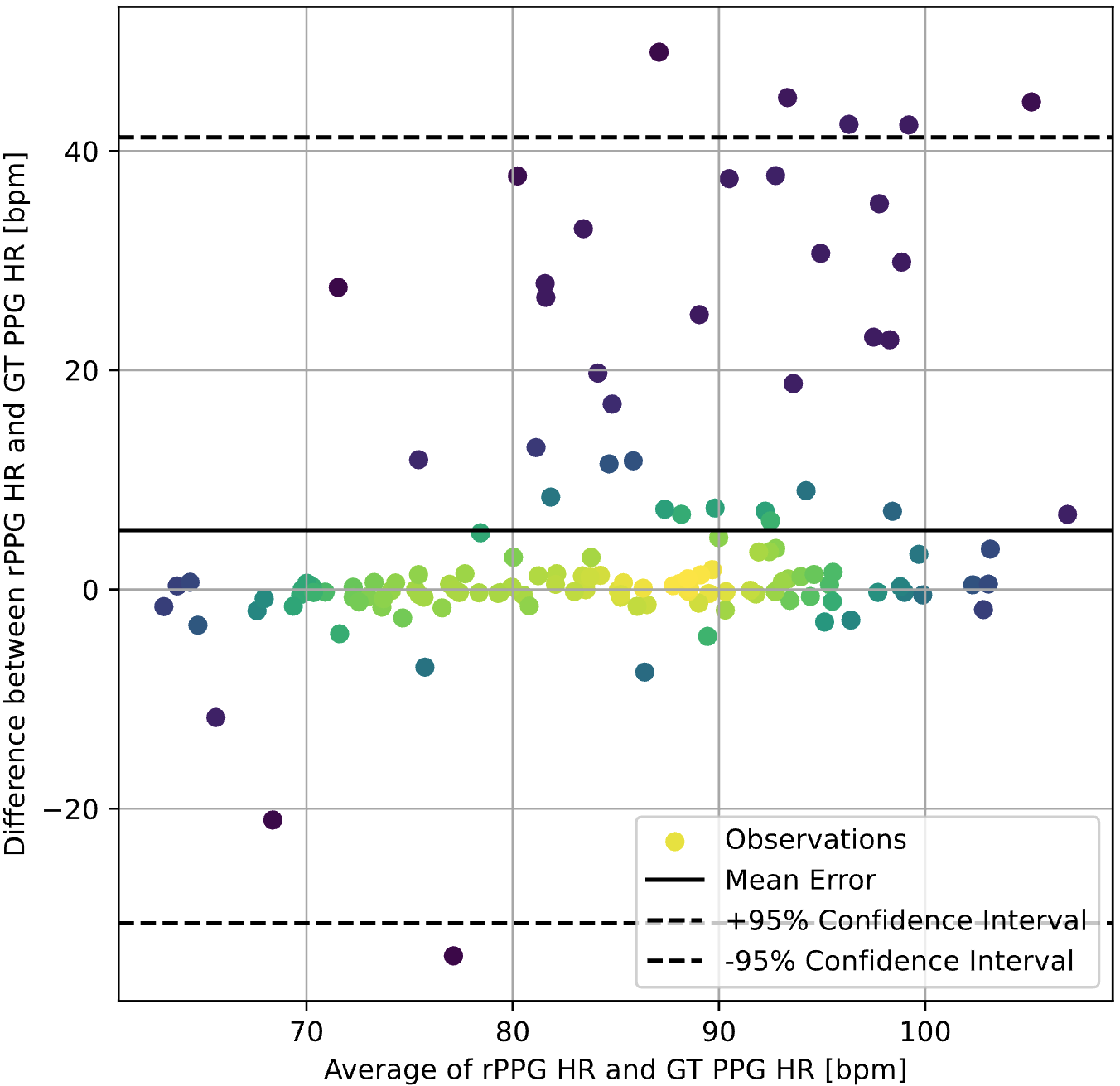} \end{minipage} &
    \begin{minipage}{.29\textwidth} \includegraphics[width=\linewidth]{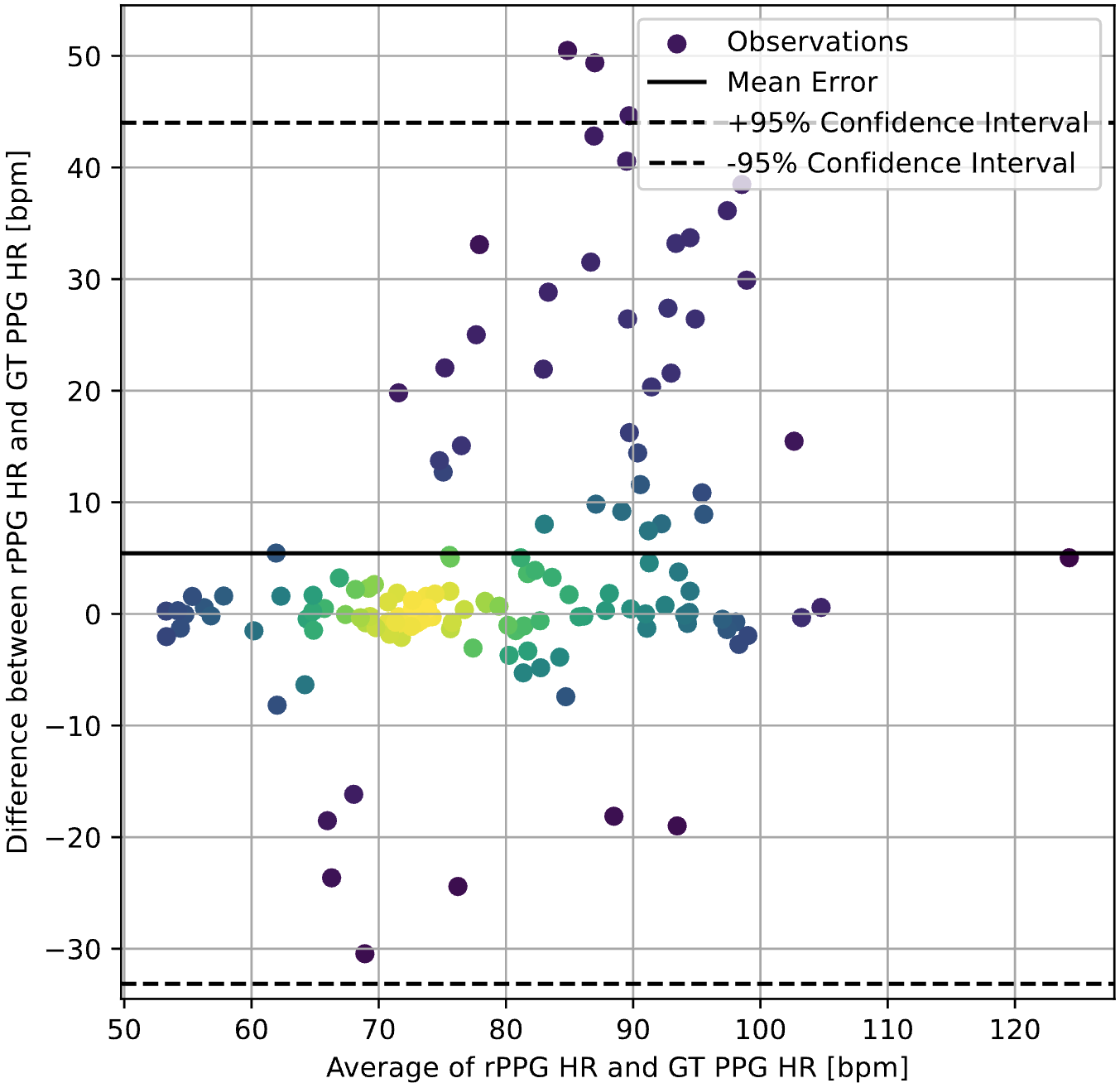} \end{minipage} &
    \begin{minipage}{.29\textwidth} \includegraphics[width=\linewidth]{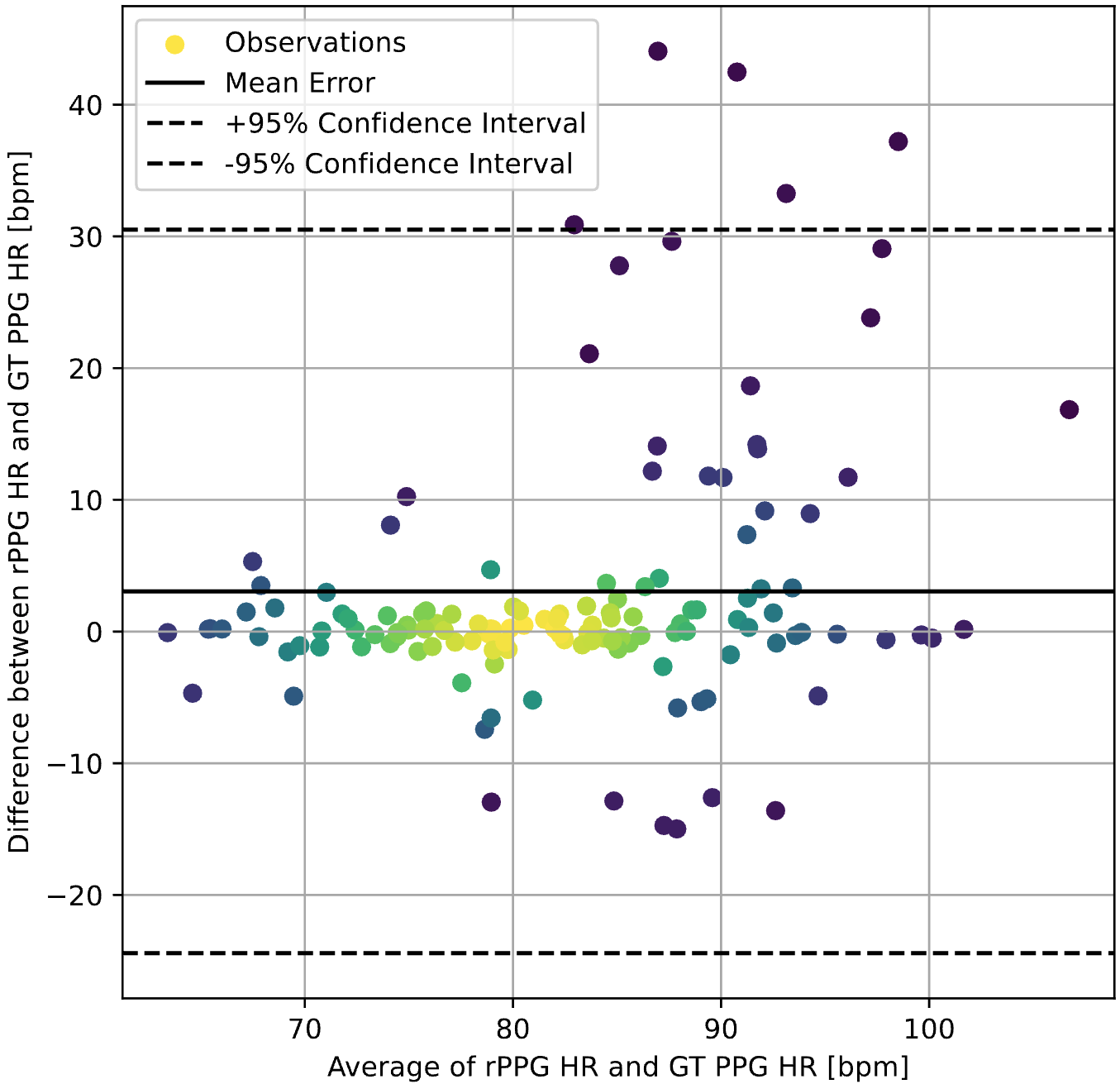} \end{minipage} \\ \bottomrule
\end{tabular}
\end{figure*}

\begin{figure*}[]
\caption{Intra-dataset experiment results on RLAP.}
\label{fig:5}
\begin{tabular}{@{}p{0.05\textwidth}lll@{}}
\toprule
\multicolumn{1}{c}{Dataset} & \multicolumn{1}{c}{GVT2RPM-MViT-optimal} & \multicolumn{1}{c}{GVT2RPM-UniFormer-optimal} & \multicolumn{1}{c}{GVT2RPM-Swin-optimal} \\ \midrule
\begin{tabular}[c]{@{}c@{}}RLAP\end{tabular} & 
    \begin{minipage}{.29\textwidth} \includegraphics[width=\linewidth]{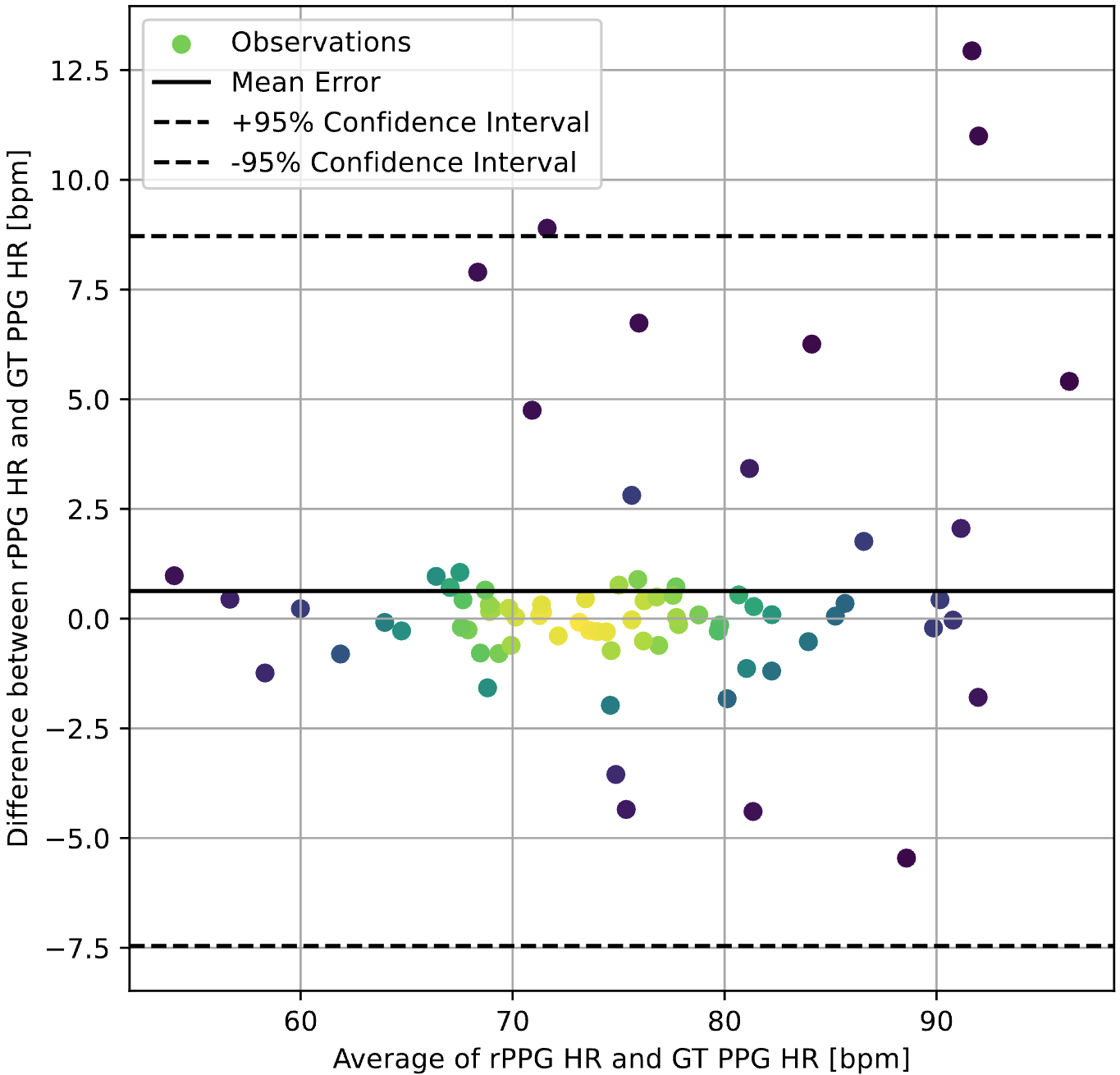} \end{minipage} &
    \begin{minipage}{.29\textwidth} \includegraphics[width=\linewidth]{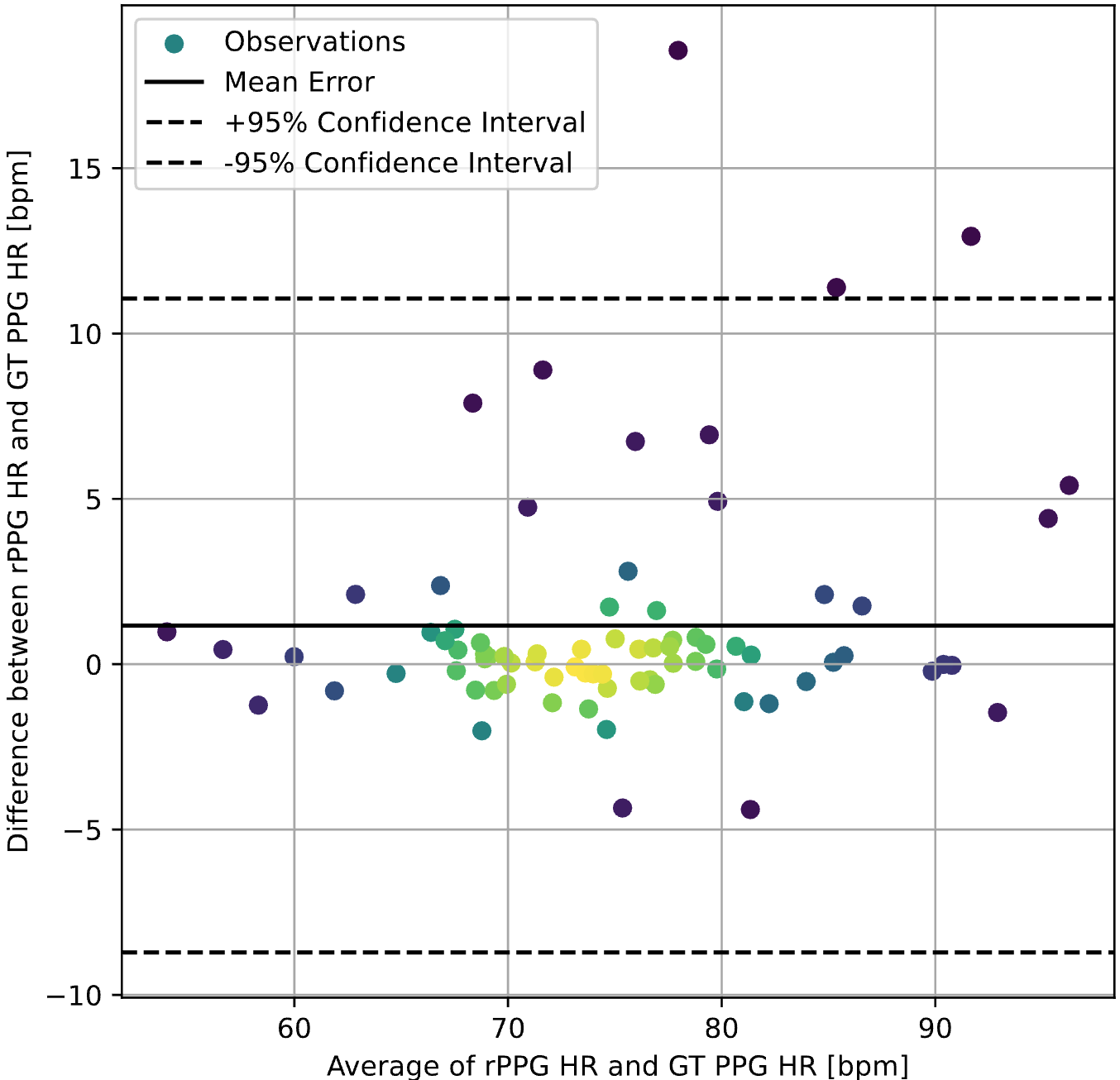} \end{minipage} &
    \begin{minipage}{.29\textwidth} \includegraphics[width=\linewidth]{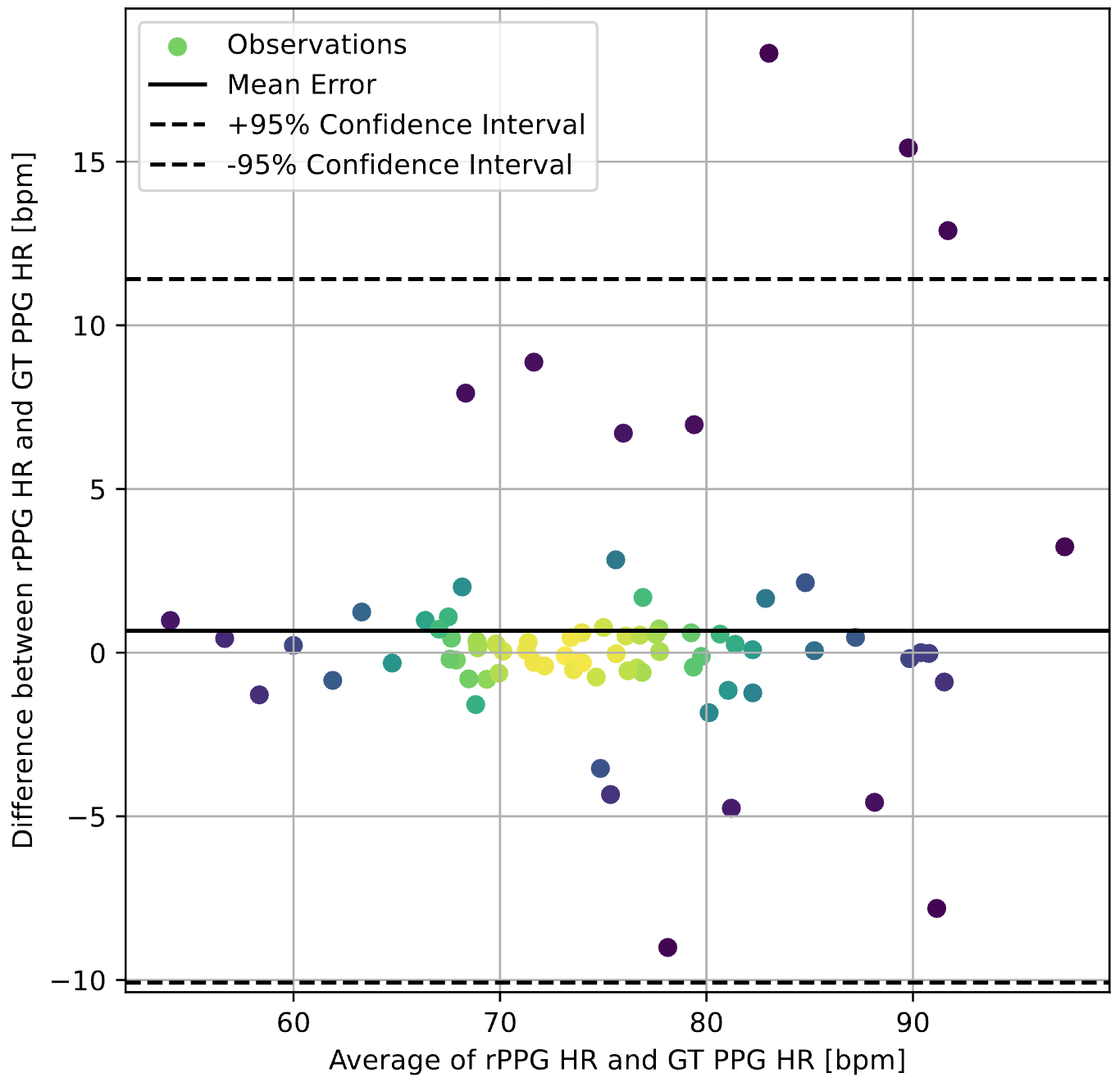} \end{minipage} \\ \bottomrule
\end{tabular}
\end{figure*}

For intra-dataset experiments, we evaluated on four datasets, including MMPD-Simple \cite{ref30}, MMPD \cite{ref30}, UBFC-rPPG \cite{ref32}, and RLAP \cite{ref31}. We show method performances of each fold in Table~\ref{tab:2},\ref{tab:3},\ref{tab:4},\ref{tab:5}. The results are shown with \textit{mean} $\pm$ \textit{std}. Additionally, for three GVT2RPM-*-optimal methods, we visualized the differences between predictions and ground truth values by Bland-Altman plots in Figure~\ref{fig:2},\ref{fig:3},\ref{fig:4},\ref{fig:5}.

For cross-dataset experiments, we used five datasets, including MMPD-Simple \cite{ref30}, MMPD \cite{ref30}, UBFC-rPPG \cite{ref32}, RLAP \cite{ref31}, and UBFC-Phys \cite{ref33}. We show detailed results in Table~\ref{tab:6},\ref{tab:7},\ref{tab:8},\ref{tab:9},\ref{tab:10}. The results are shown with \textit{mean} $\pm$ \textit{std}.

\begin{table*}[]
\caption{Intra-dataset experiment results on UBFC-rPPG.}
\label{tab:2}
\resizebox{.98\textwidth}{!}{%
\begin{tabular}{@{}lccclccclccc@{}}
\toprule
\multicolumn{1}{c}{\multirow{2}{*}{Methods}} & \multicolumn{3}{c}{Fold 0} &  & \multicolumn{3}{c}{Fold 1} &  & \multicolumn{3}{c}{Fold 2} \\ \cmidrule(lr){2-4} \cmidrule(lr){6-8} \cmidrule(l){10-12} 
\multicolumn{1}{c}{} & MAE$\downarrow$ & RMSE$\downarrow$ & $\rho \uparrow$ & \multicolumn{1}{c}{} & MAE$\downarrow$ & RMSE$\downarrow$ & $\rho \uparrow$ & \multicolumn{1}{c}{} & MAE$\downarrow$ & RMSE$\downarrow$ & $\rho \uparrow$ \\ \midrule
DeepPhys & 1.76$\pm$1.22 & 4.08$\pm$13.57 & 0.98$\pm$0.07 &  & 0$\pm$0 & 0$\pm$0 & 1$\pm$0 &  & 3.91$\pm$1.50 & 5.96$\pm$14.90 & 0.97$\pm$0.09 \\
TS-CAN & 1.07$\pm$0.91 & 2.94$\pm$8.08 & 0.99$\pm$0.05 &  & 0$\pm$0 & 0$\pm$0 & 1$\pm$0 &  & 3.13$\pm$1.50 & 5.48$\pm$15.24 & 0.97$\pm$0.08 \\
EfficientPhys & 1.07$\pm$0.91 & 2.94$\pm$8.08 & 0.99$\pm$0.05 &  & 7.52$\pm$6.30 & 20.35$\pm$384.66 & 0.60$\pm$0.30 &  & 1.86$\pm$1.16 & 3.94$\pm$9.75 & 0.99$\pm$0.03 \\
PhysNet & 1.95$\pm$1.21 & 4.12$\pm$13.53 & 0.97$\pm$0.07 &  & 1.86$\pm$0.96 & 3.42$\pm$6.75 & 0.98$\pm$0.07 &  & 2.93$\pm$1.43 & 5.19$\pm$14.43 & 0.98$\pm$0.08 \\
PhysFormer & 1.46$\pm$0.98 & 3.28$\pm$8.09 & 0.99$\pm$0.06 &  & 1.27$\pm$0.75 & 2.57$\pm$5.10 & 0.99$\pm$0.04 &  & 2.34$\pm$1.13 & 4.12$\pm$8.61 & 0.99$\pm$0.05 \\ \midrule
GVT2RPM-MViT-optimal & 1.56$\pm$1.21 & 4.12$\pm$13.52 & 0.99$\pm$0.06 &  & 1.56$\pm$0.97 & 3.31$\pm$6.85 & 0.99$\pm$0.06 &  & 0.87$\pm$0.82 & 2.63$\pm$6.55 & 0.99$\pm$0.03 \\
GVT2RPM-UniFormer-optimal & 0.48$\pm$0.46 & 1.46$\pm$2.02 & 0.99$\pm$0.02 &  & 1.85$\pm$0.96 & 3.42$\pm$6.75 & 0.98$\pm$0.06 &  & 1.85$\pm$1.15 & 3.94$\pm$9.75 & 0.98$\pm$0.07 \\
GVT2RPM-Swin-optimal & 1.36$\pm$1.09 & 3.56$\pm$11.61 & 0.99$\pm$0.06 &  & 1.07$\pm$0.75 & 2.50$\pm$5.13 & 0.99$\pm$0.03 &  & 2.14$\pm$1.36 & 4.63$\pm$14.40 & 0.99$\pm$0.04 \\ \bottomrule
\end{tabular}%
}
\end{table*}

\begin{table*}[]
\caption{Intra-dataset experiment results on MMPD-simple.}
\label{tab:3}
\resizebox{.98\textwidth}{!}{%
\begin{tabular}{@{}lccclccclccc@{}}
\toprule
\multicolumn{1}{c}{\multirow{2}{*}{Methods}} & \multicolumn{3}{c}{Fold 0} &  & \multicolumn{3}{c}{Fold 1} &  & \multicolumn{3}{c}{Fold 2} \\ \cmidrule(lr){2-4} \cmidrule(lr){6-8} \cmidrule(l){10-12} 
\multicolumn{1}{c}{} & MAE$\downarrow$ & RMSE$\downarrow$ & $\rho \uparrow$ & \multicolumn{1}{c}{} & MAE$\downarrow$ & RMSE$\downarrow$ & $\rho \uparrow$ & \multicolumn{1}{c}{} & MAE$\downarrow$ & RMSE$\downarrow$ & $\rho \uparrow$ \\ \midrule
DeepPhys & 17.50$\pm$4.18 & 22.32$\pm$159.07 & -0.002$\pm$0.33 &  & 2.72$\pm$1.14 & 4.51$\pm$12.20 & 0.60$\pm$0.28 &  & 0.62$\pm$0.39 & 1.39$\pm$1.28 & 0.87$\pm$0.17 \\
TS-CAN & 2.00$\pm$0.76 & 3.21$\pm$5.06 & 0.95$\pm$0.10 &  & 1.23$\pm$0.60 & 2.26$\pm$2.99 & 0.92$\pm$0.14 &  & 0.79$\pm$0.50 & 1.78$\pm$2.06 & 0.94$\pm$0.12 \\
EfficientPhys & 1.54$\pm$0.63 & 2.67$\pm$3.53 & 0.97$\pm$0.08 &  & 1.32$\pm$0.62 & 2.51$\pm$3.17 & 0.88$\pm$0.15 &  & 0.88$\pm$0.56 & 2.07$\pm$2.79 & 0.91$\pm$0.14 \\
PhysNet & 1.52$\pm$0.77 & 2.96$\pm$6.61 & 0.97$\pm$0.08 &  & 1.32$\pm$0.81 & 2.87$\pm$6.00 & 0.85$\pm$0.19 &  & 0.18$\pm$0.11 & 0.39$\pm$0.10 & 0.99$\pm$0.04 \\
PhysFormer & 18.78$\pm$3.03 & 21.30$\pm$136.53 & 0.20$\pm$0.33 &  & 3.60$\pm$0.81 & 4.42$\pm$6.04 & 0.51$\pm$0.30 &  & 1.67$\pm$0.74 & 2.87$\pm$4.78 & 0.73$\pm$0.24 \\ \midrule
GVT2RPM-MViT-optimal & 0.51$\pm$0.36 & 1.37$\pm$1.53 & 0.99$\pm$0.03 &  & 0.80$\pm$0.38 & 1.79$\pm$2.41 & 0.94$\pm$0.10 &  & 0.64$\pm$0.45 & 1.63$\pm$2.39 & 0.85$\pm$0.17 \\
GVT2RPM-UniFormer-optimal & 1.17$\pm$0.57 & 2.29$\pm$3.21 & 0.98$\pm$0.06 &  & 1.19$\pm$0.52 & 2.10$\pm$2.20 & 0.90$\pm$0.14 &  & 0.55$\pm$0.37 & 1.37$\pm$1.66 & 0.91$\pm$0.13 \\
GVT2RPM-Swin-optimal & 1.02$\pm$0.39 & 1.72$\pm$1.54 & 0.99$\pm$0.05 &  & 3.03$\pm$1.09 & 4.72$\pm$11.61 & 0.49$\pm$0.28 &  & 1.91$\pm$0.67 & 2.95$\pm$3.86 & 0.56$\pm$0.27 \\ \bottomrule
\end{tabular}%
}
\end{table*}

\begin{table*}[]
\caption{Intra-dataset experiment results on MMPD.}
\label{tab:4}
\resizebox{.98\textwidth}{!}{%
\begin{tabular}{@{}lccclccclccc@{}}
\toprule
\multicolumn{1}{c}{\multirow{2}{*}{Methods}} & \multicolumn{3}{c}{Fold 0} &  & \multicolumn{3}{c}{Fold 1} &  & \multicolumn{3}{c}{Fold 2} \\ \cmidrule(lr){2-4} \cmidrule(lr){6-8} \cmidrule(l){10-12} 
\multicolumn{1}{c}{} & MAE$\downarrow$ & RMSE$\downarrow$ & $\rho \uparrow$ & \multicolumn{1}{c}{} & MAE$\downarrow$ & RMSE$\downarrow$ & $\rho \uparrow$ & \multicolumn{1}{c}{} & MAE$\downarrow$ & RMSE$\downarrow$ & $\rho \uparrow$ \\ \midrule
DeepPhys & 22.37$\pm$1.56 & 28.85$\pm$88.97 & 0.04$\pm$0.09 &  & 20.30$\pm$1.44 & 26.40$\pm$78.66 & 0.05$\pm$0.09 &  & 17.98$\pm$1.23 & 23.02$\pm$57.35 & 0.15$\pm$0.09 \\
TS-CAN & 10.40$\pm$1.36 & 19.01$\pm$67.11 & 0.32$\pm$0.08 &  & 10.95$\pm$1.20 & 17.88$\pm$50.47 & 0.48$\pm$0.07 &  & 9.45$\pm$1.06 & 15.55$\pm$38.63 & 0.43$\pm$0.07 \\
EfficientPhys & 13.67$\pm$1.50 & 22.36$\pm$76.37 & 0.19$\pm$0.08 &  & 13.30$\pm$1.35 & 20.75$\pm$66.85 & 0.25$\pm$0.08 &  & 11.96$\pm$1.15 & 18.09$\pm$44.92 & 0.37$\pm$0.08 \\ \midrule
GVT2RPM-MViT-optimal & 6.04$\pm$0.86 & 11.72$\pm$29.39 & 0.63$\pm$0.06 &  & 8.31$\pm$0.96 & 14.05$\pm$35.28 & 0.58$\pm$0.07 &  & 4.69$\pm$0.66 & 9.08$\pm$18.64 & 0.67$\pm$0.06 \\
GVT2RPM-UniFormer-optimal & 6.91$\pm$0.92 & 12.85$\pm$31.05 & 0.52$\pm$0.07 &  & 7.38$\pm$0.96 & 13.54$\pm$37.46 & 0.60$\pm$0.06 &  & 5.81$\pm$0.75 & 10.57$\pm$22.04 & 0.54$\pm$0.07 \\
GVT2RPM-Swin-optimal & 6.94$\pm$1.04 & 13.97$\pm$40.20 & 0.45$\pm$0.07 &  & 8.34$\pm$1.04 & 14.89$\pm$41.41 & 0.58$\pm$0.07 &  & 5.18$\pm$0.76 & 10.33$\pm$26.39 & 0.55$\pm$0.07 \\ \bottomrule
\end{tabular}%
}
\end{table*}

\begin{table*}[]
\caption{Intra-dataset experiment results on RLAP.}
\label{tab:5}
\resizebox{.5\textwidth}{!}{%
\begin{tabular}{@{}lccc@{}}
\toprule
\multicolumn{1}{c}{\multirow{2}{*}{Methods}} & \multicolumn{3}{c}{Official Split} \\ \cmidrule(l){2-4} 
\multicolumn{1}{c}{} & MAE$\downarrow$ & RMSE$\downarrow$ & $\rho \uparrow$ \\ \midrule
DeepPhys & 3.80$\pm$0.69 & 7.90$\pm$17.35 & 0.71$\pm$0.07 \\
TS-CAN & 2.58$\pm$0.57 & 6.33$\pm$13.54 & 0.78$\pm$0.06 \\
EfficientPhys & 2.98$\pm$0.57 & 6.35$\pm$11.71 & 0.80$\pm$0.06 \\
PhysNet & 1.30$\pm$0.32 & 3.38$\pm$4.35 & 0.93$\pm$0.04 \\
PhysFormer & 1.53$\pm$0.37 & 3.98$\pm$6.29 & 0.91$\pm$0.04 \\ \midrule
GVT2RPM-MViT-optimal & 1.32$\pm$0.30 & 2.96$\pm$3.01 & 0.95$\pm$0.03 \\
GVT2RPM-UniFormer-optimal & 1.60$\pm$0.38 & 3.73$\pm$5.67 & 0.92$\pm$0.04 \\
GVT2RPM-Swin-optimal & 1.64$\pm$0.40 & 3.91$\pm$5.62 & 0.91$\pm$0.04 \\ \bottomrule
\end{tabular}%
}
\end{table*}

\begin{table}[]
\caption{Cross-dataset experiment results testing on UBFC-rPPG.}
\label{tab:6}
\resizebox{.95\linewidth}{!}{%
\begin{tabular}{@{}llccc@{}}
\toprule
\multicolumn{1}{c}{} &  & \multicolumn{3}{c}{Testing on UBFC-rPPG} \\ \cmidrule(l){3-5} 
Method & Training Set & MAE$\downarrow$ & RMSE$\downarrow$ & $\rho \uparrow$ \\ \midrule
\multirow{3}{*}{DeepPhys} & MMPD-simple & 27.25$\pm$3.84 & 36.90$\pm$279.99 & 0.09$\pm$0.16 \\
 & MMPD & 29.72$\pm$3.16 & 36.10$\pm$195.37 & 0.21$\pm$0.15 \\
 & RLAP & 1.15$\pm$0.40 & 2.87$\pm$3.77 & 0.99$\pm$0.02 \\
\multicolumn{5}{l}{} \\
\multirow{3}{*}{TS-CAN} & MMPD-simple & 15.34$\pm$3.55 & 27.63$\pm$217.61 & 0.35$\pm$0.15 \\
 & MMPD & 16.22$\pm$3.24 & 26.53$\pm$181.05 & 0.47$\pm$0.14 \\
 & RLAP & 0.96$\pm$0.37 & 2.60$\pm$3.62 & 0.99$\pm$0.02 \\
\multicolumn{5}{l}{} \\
\multirow{3}{*}{EfficientPhys} & MMPD-simple & 15.11$\pm$3.27 & 26.03$\pm$196.11 & 0.36$\pm$0.15 \\
 & MMPD & 17.47$\pm$3.41 & 28.15$\pm$201.80 & 0.40$\pm$0.14 \\
 & RLAP & 1.95$\pm$0.86 & 5.90$\pm$26.31 & 0.95$\pm$0.05 \\
\multicolumn{5}{l}{} \\
\multirow{3}{*}{PhysNet} & MMPD-simple & 13.20$\pm$2.66 & 21.69$\pm$143.00 & 0.55$\pm$0.13 \\
 & MMPD & N/A & N/A & N/A \\
 & RLAP & 8.20$\pm$2.52 & 18.30$\pm$140.44 & 0.51$\pm$0.14 \\
\multicolumn{5}{l}{} \\
\multirow{3}{*}{PhysFormer} & MMPD-simple & 19.11$\pm$3.46 & 29.45$\pm$211.58 & 0.14$\pm$0.16 \\
 & MMPD & N/A & N/A & N/A \\
 & RLAP & 6.88$\pm$2.06 & 15.03$\pm$92.90 & 0.67$\pm$0.12 \\ \midrule
\multirow{3}{*}{\begin{tabular}[c]{@{}l@{}}GVT2RPM-\\ MViT-optimal\end{tabular}} & MMPD-simple & 1.46$\pm$0.37 & 2.79$\pm$2.43 & 0.99$\pm$0.02 \\
 & MMPD & 2.05$\pm$0.61 & 4.48$\pm$10.17 & 0.97$\pm$0.04 \\
 & RLAP & 1.21$\pm$0.41 & 2.92$\pm$3.76 & 0.99$\pm$0.02 \\ \bottomrule
\end{tabular}%
}
\end{table}

\begin{table}[]
\caption{Cross-dataset experiment results testing on UBFC-Phys.}
\label{tab:7}
\resizebox{.95\linewidth}{!}{%
\begin{tabular}{@{}llccc@{}}
\toprule
\multicolumn{1}{c}{} &  & \multicolumn{3}{c}{Testing on UBFC-Phys} \\ \cmidrule(l){3-5} 
Method & Training Set & MAE$\downarrow$ & RMSE$\downarrow$ & $\rho \uparrow$ \\ \midrule
\multirow{4}{*}{DeepPhys} & UBFC-rPPG & 6.01$\pm$0.88 & 10.70$\pm$28.05 & 0.64$\pm$0.08 \\
 & MMPD-simple & 13.36$\pm$1.26 & 18.43$\pm$52.96 & 0.20$\pm$0.10 \\
 & MMPD & 14.65$\pm$1.28 & 19.49$\pm$56.77 & 0.07$\pm$0.10 \\
 & RLAP & 4.81$\pm$0.68 & 8.32$\pm$16.15 & 0.76$\pm$0.07 \\
\multicolumn{5}{l}{} \\
\multirow{4}{*}{TS-CAN} & UBFC-rPPG & 5.27$\pm$0.68 & 8.63$\pm$14.86 & 0.74$\pm$0.07 \\
 & MMPD-simple & 7.91$\pm$1.02 & 12.91$\pm$34.83 & 0.49$\pm$0.09 \\
 & MMPD & 7.19$\pm$0.95 & 11.96$\pm$30.15 & 0.53$\pm$0.08 \\
 & RLAP & 4.36$\pm$0.65 & 7.89$\pm$16.25 & 0.78$\pm$0.06 \\
\multicolumn{5}{l}{} \\
\multirow{4}{*}{EfficientPhys} & UBFC-rPPG & 6.07$\pm$0.86 & 10.57$\pm$24.47 & 0.64$\pm$0.08 \\
 & MMPD-simple & 5.14$\pm$0.79 & 9.49$\pm$27.11 & 0.70$\pm$0.07 \\
 & MMPD & 5.79$\pm$0.80 & 9.95$\pm$21.92 & 0.67$\pm$0.07 \\
 & RLAP & 4.28$\pm$0.67 & 7.95$\pm$16.55 & 0.78$\pm$0.06 \\
\multicolumn{5}{l}{} \\
\multirow{4}{*}{PhysNet} & UBFC-rPPG & 4.54$\pm$0.75 & 8.80$\pm$24.62 & 0.75$\pm$0.07 \\
 & MMPD-simple & 7.28$\pm$0.87 & 11.42$\pm$25.85 & 0.50$\pm$0.09 \\
 & MMPD & N/A & N/A & N/A \\
 & RLAP & 4.48$\pm$0.74 & 8.66$\pm$25.33 & 0.76$\pm$0.07 \\
\multicolumn{5}{l}{} \\
\multirow{4}{*}{PhysFormer} & UBFC-rPPG & 5.13 $\pm$0.73 & 8.98$\pm$18.49 & 0.73 $\pm$0.07 \\
 & MMPD-simple & 9.23 $\pm$0.95 & 13.28$\pm$30.64 & 0.37$\pm$0.09 \\
 & MMPD & N/A & N/A & N/A \\
 & RLAP & 4.48$\pm$0.70 & 8.36$\pm$22.12 & 0.76$\pm$0.07 \\ \midrule
\multirow{4}{*}{\begin{tabular}[c]{@{}l@{}}GVT2RPM-\\ MViT-optimal\end{tabular}} & UBFC-rPPG & 4.22$\pm$0.64 & 7.65$\pm$13.59 & 0.79$\pm$0.06 \\
 & MMPD-simple & 5.09$\pm$0.60 & 7.97$\pm$11.99 & 0.76$\pm$0.07 \\
 & MMPD & 4.32$\pm$0.64 & 7.76$\pm$14.83 & 0.78$\pm$0.06 \\
 & RLAP & 4.17$\pm$0.71 & 8.26$\pm$23.49 & 0.77$\pm$0.06 \\ \bottomrule
\end{tabular}%
}
\end{table}

\begin{table}[]
\caption{Cross-dataset experiment results testing on MMPD-simple.}
\label{tab:8}
\resizebox{\linewidth}{!}{%
\begin{tabular}{@{}llccc@{}}
\toprule
\multicolumn{1}{c}{} &  & \multicolumn{3}{c}{Testing on MMPD-simple} \\ \cmidrule(l){3-5} 
Method & Training Set & MAE$\downarrow$ & RMSE$\downarrow$ & $\rho \uparrow$ \\ \midrule
\multirow{2}{*}{DeepPhys} & UBFC-rPPG & 2.98$\pm$0.81 & 6.35$\pm$21.14 & 0.82$\pm$0.09 \\
 & RLAP & 1.87$\pm$0.61 & 4.60$\pm$14.38 & 0.88$\pm$0.07 \\
\multicolumn{5}{l}{} \\
\multirow{2}{*}{TS-CAN} & UBFC-rPPG & 1.61$\pm$0.40 & 3.22$\pm$4.20 & 0.94$\pm$0.05 \\
 & RLAP & 1.32$\pm$0.37 & 2.87$\pm$3.58 & 0.96$\pm$0.04 \\
\multicolumn{5}{l}{} \\
\multirow{2}{*}{EfficientPhys} & UBFC-rPPG & 0.91$\pm$0.25 & 2.01$\pm$1.60 & 0.98$\pm$0.03 \\
 & RLAP & 0.97$\pm$0.25 & 2.02$\pm$1.43 & 0.98$\pm$0.03 \\
\multicolumn{5}{l}{} \\
\multirow{2}{*}{PhysNet} & UBFC-rPPG & 2.69$\pm$0.91 & 6.95$\pm$31.79 & 0.70$\pm$0.10 \\
 & RLAP & 1.52$\pm$0.42 & 3.23$\pm$4.32 & 0.95$\pm$0.05 \\
\multicolumn{5}{l}{} \\
\multirow{2}{*}{PhysFormer} & UBFC-rPPG & 7.38$\pm$1.97 & 15.53$\pm$106.46 & 0.14$\pm$0.15 \\
 & RLAP & 2.55$\pm$0.78 & 5.96$\pm$19.24 & 0.78$\pm$0.09 \\ \midrule
\multirow{2}{*}{\begin{tabular}[c]{@{}l@{}}GVT2RPM-\\ MViT-optimal\end{tabular}} & UBFC-rPPG & 1.87$\pm$0.82 & 6.07$\pm$31.55 & 0.78$\pm$0.09 \\
 & RLAP & 0.79$\pm$0.22 & 1.70$\pm$1.15 & 0.98$\pm$0.03 \\ \bottomrule
\end{tabular}%
}
\end{table}

\begin{table}[]
\caption{Cross-dataset experiment results testing on MMPD.}
\label{tab:9}
\resizebox{\linewidth}{!}{%
\begin{tabular}{@{}llccc@{}}
\toprule
\multicolumn{1}{c}{} &  & \multicolumn{3}{c}{Testing on MMPD} \\ \cmidrule(l){3-5} 
Method & Training Set & MAE$\downarrow$ & RMSE$\downarrow$ & $\rho \uparrow$ \\ \midrule
\multirow{2}{*}{DeepPhys} & UBFC-rPPG & 17.72$\pm$0.67 & 24.63$\pm$37.43 & 0.14$\pm$0.04 \\
 & RLAP & 16.74$\pm$0.72 & 24.82$\pm$40.87 & 0.05$\pm$0.04 \\
\multicolumn{5}{l}{} \\
\multirow{2}{*}{TS-CAN} & UBFC-rPPG & 13.52$\pm$0.62 & 20.84$\pm$31.29 & 0.22$\pm$0.04 \\
 & RLAP & 13.34$\pm$0.63 & 20.97$\pm$32.46 & 0.21$\pm$0.04 \\
\multicolumn{5}{l}{} \\
\multirow{2}{*}{EfficientPhys} & UBFC-rPPG & 13.08$\pm$0.64 & 20.99$\pm$32.99 & 0.20$\pm$0.04 \\
 & RLAP & 12.69$\pm$0.62 & 20.38$\pm$31.73 & 0.21$\pm$0.04 \\
\multicolumn{5}{l}{} \\
\multirow{2}{*}{PhysNet} & UBFC-rPPG & 9.94$\pm$0.48 & 15.84$\pm$20.38 & 0.32$\pm$0.04 \\
 & RLAP & 9.15$\pm$0.50 & 15.67$\pm$21.65 & 0.35$\pm$0.04 \\
\multicolumn{5}{l}{} \\
\multirow{2}{*}{PhysFormer} & UBFC-rPPG & 12.98$\pm$0.54 & 19.01$\pm$27.16 & 0.13$\pm$0.04 \\
 & RLAP & 9.99$\pm$0.49 & 15.91$\pm$21.19 & 0.32$\pm$0.04 \\ \midrule
\multirow{2}{*}{\begin{tabular}[c]{@{}l@{}}GVT2RPM-\\ MViT-optimal\end{tabular}} & UBFC-rPPG & 10.23$\pm$0.48 & 15.94$\pm$20.38 & 0.31$\pm$0.04 \\
 & RLAP & 8.28$\pm$0.44 & 13.90$\pm$16.63 & 0.45$\pm$0.03 \\ \bottomrule
\end{tabular}%
}
\end{table}

\begin{table}[]
\caption{Cross-dataset experiment results testing on RLAP.}
\label{tab:10}
\resizebox{\linewidth}{!}{%
\begin{tabular}{@{}llccc@{}}
\toprule
\multicolumn{1}{c}{} &  & \multicolumn{3}{c}{Testing on RLAP} \\ \cmidrule(l){3-5} 
Method & Training Set & MAE$\downarrow$ & RMSE$\downarrow$ & $\rho \uparrow$ \\ \midrule
\multirow{3}{*}{DeepPhys} & UBFC-rPPG & 4.90$\pm$0.44 & 10.25$\pm$15.77 & 0.54$\pm$0.04 \\
 & MMPD-simple & 10.65$\pm$0.56 & 15.64$\pm$21.29 & 0.17$\pm$0.05 \\
 & MMPD & 10.89$\pm$0.54 & 15.50$\pm$19.66 & 0.19$\pm$0.05 \\
\multicolumn{5}{l}{} \\
\multirow{3}{*}{TS-CAN} & UBFC-rPPG & 3.20$\pm$0.31 & 7.07$\pm$9.37 & 0.76$\pm$0.03 \\
 & MMPD-simple & 5.89$\pm$0.43 & 10.55$\pm$13.08 & 0.51$\pm$0.04 \\
 & MMPD & 7.07$\pm$0.47 & 11.89$\pm$18.62 & 0.37$\pm$0.06 \\
\multicolumn{5}{l}{} \\
\multirow{3}{*}{EfficientPhys} & UBFC-rPPG & 3.77$\pm$0.38 & 8.49$\pm$12.95 & 0.66$\pm$0.04 \\
 & MMPD-simple & 3.89$\pm$0.35 & 8.06$\pm$10.65 & 0.67$\pm$0.04 \\
 & MMPD & 4.05$\pm$0.38 & 8.34$\pm$13.54 & 0.64$\pm$0.04 \\
\multicolumn{5}{l}{} \\
\multirow{3}{*}{PhysNet} & UBFC-rPPG & 3.39$\pm$0.37 & 8.43$\pm$10.89 & 0.68$\pm$0.04 \\
 & MMPD-simple & 6.11$\pm$0.42 & 10.36$\pm$12.71 & 0.49$\pm$0.04 \\
 & MMPD & N/A & N/A & N/A \\
\multicolumn{5}{l}{} \\
\multirow{3}{*}{PhysFormer} & UBFC-rPPG & 4.44$\pm$0.46 & 10.18$\pm$17.35 & 0.54$\pm$0.04 \\
 & MMPD-simple & 9.12$\pm$0.64 & 15.73$\pm$27.08 & 0.32$\pm$0.05 \\
 & MMPD & N/A & N/A & N/A \\ \midrule
\multirow{3}{*}{\begin{tabular}[c]{@{}l@{}}GVT2RPM-\\ MViT-optimal\end{tabular}} & UBFC-rPPG & 2.83$\pm$0.38 & 7.39$\pm$13.73 & 0.72$\pm$0.04 \\
 & MMPD-simple & 3.02$\pm$0.34 & 6.80$\pm$11.10 & 0.76$\pm$0.04 \\
 & MMPD & 2.77$\pm$0.35 & 6.78$\pm$10.82 & 0.79$\pm$0.03 \\ \bottomrule
\end{tabular}%
}
\end{table}


\section{Dataset details}
For cross-dataset experiments, the training set was spited into train/val set with ratio 8:2. For intra-dataset experiments, 3-fold cross-validation was used and we show the fold details in the following:

\subsection{MMPD-simple}
Due to the difficulty of the original MMPD, authors created a subset to contain videos with stationary, skin tone type 3, and artificial light conditions.

For Fold 0, the training set is \{'subject25', 'subject6', 'subject24', 'subject4', 'subject22', 'subject12', 'subject21', 'subject9', 'subject18', 'subject5', 'subject3', 'subject19', 'subject20'\}, and the testing set is \{'subject32', 'subject33', 'subject29', 'subject27'\}.

For Fold 1, the training set is \{'subject29', 'subject6', 'subject21', 'subject19', 'subject4', 'subject25', 'subject18', 'subject3', 'subject32', 'subject22', 'subject24', 'subject33', 'subject12'\}, and the testing set is \{'subject27', 'subject9', 'subject5', 'subject20'\}.

For Fold 2, the training set is \{'subject20', 'subject19', 'subject29', 'subject9', 'subject6', 'subject21', 'subject12', 'subject5', 'subject32', 'subject25', 'subject3', 'subject33', 'subject18'\}, and the testing set is \{'subject27', 'subject4', 'subject22', 'subject24'\}.

\subsection{MMPD}
For Fold 0, the training set is \{'subject8', 'subject26', 'subject21', 'subject15', 'subject25', 'subject12', 'subject19', 'subject9', 'subject23', 'subject10', 'subject11', 'subject24', 'subject16', 'subject4', 'subject5', 'subject3', 'subject18', 'subject14', 'subject13', 'subject20', 'subject1', 'subject22', 'subject6', 'subject17', 'subject7', 'subject2'\}, and the testing set is \{'subject33', 'subject28', 'subject32', 'subject30', 'subject27', 'subject31', 'subject29'\}.

For Fold 1, the training set is \{'subject20', 'subject7', 'subject11', 'subject4', 'subject25', 'subject32', 'subject6', 'subject15', 'subject33', 'subject16', 'subject22', 'subject26', 'subject23', 'subject19', 'subject10', 'subject21', 'subject8', 'subject14', 'subject18', 'subject3', 'subject13', 'subject2', 'subject28', 'subject24', 'subject29', 'subject27'\}, and the testing set is \{'subject5', 'subject31', 'subject17', 'subject30', 'subject1', 'subject12', 'subject9'\}.

For Fold 2, the training set is \{'subject33', 'subject2', 'subject3', 'subject13', 'subject26', 'subject21', 'subject7', 'subject10', 'subject22', 'subject4', 'subject20', 'subject27', 'subject6', 'subject5', 'subject12', 'subject19', 'subject30', 'subject11', 'subject1', 'subject24', 'subject32', 'subject25', 'subject8', 'subject14', 'subject16', 'subject29'\}, and the testing set is \{'subject23', 'subject18', 'subject31', 'subject9', 'subject28', 'subject17', 'subject15'\}.

\subsection{RLAP}
For Fold 0, the training set is \{'subject6', 'subject46', 'subject53', 'subject48', 'subject58', 'subject31', 'subject13', 'subject52', 'subject56', 'subject14', 'subject36', 'subject18', 'subject38', 'subject9', 'subject49', 'subject50', 'subject43', 'subject4', 'subject44', 'subject23', 'subject15', 'subject57', 'subject33', 'subject11', 'subject24', 'subject55', 'subject27', 'subject40', 'subject45', 'subject2', 'subject3', 'subject42', 'subject35', 'subject20', 'subject21', 'subject32', 'subject10'\}, the validation set is \{'subject28', 'subject26', 'subject17', 'subject5', 'subject30'\}, and the testing set is \{'subject37', 'subject51', 'subject8', 'subject34', 'subject54', 'subject25', 'subject22', 'subject47', 'subject1', 'subject19', 'subject12', 'subject41', 'subject16', 'subject39'\}.

\subsection{UBFC-rPPG}
For Fold 0, the training set is \{'subject30', 'subject22', 'subject1', 'subject5', 'subject11', 'subject45', 'subject25', 'subject15', 'subject32', 'subject35', 'subject43', 'subject42', 'subject13', 'subject24', 'subject23', 'subject31', 'subject8', 'subject39', 'subject37', 'subject17', 'subject49', 'subject38', 'subject18', 'subject14', 'subject16', 'subject27', 'subject41', 'subject46', 'subject10', 'subject36', 'subject3', 'subject47', 'subject34'\}, and the testing set is \{subject26', 'subject4', 'subject33', 'subject9', 'subject40', 'subject12', 'subject44', 'subject48', 'subject20'\}.

For Fold 1, the training set is \{'subject45', 'subject8', 'subject12', 'subject20', 'subject27', 'subject38', 'subject30', 'subject13', 'subject23', 'subject47', 'subject16', 'subject5', 'subject41', 'subject26', 'subject25', 'subject35', 'subject22', 'subject31', 'subject10', 'subject49', 'subject44', 'subject3', 'subject18', 'subject17', 'subject46', 'subject9', 'subject39', 'subject32', 'subject42', 'subject43', 'subject37', 'subject24', 'subject40'\}, and the testing set is \{'subject4', 'subject1', 'subject36', 'subject15', 'subject34', 'subject33', 'subject11', 'subject14', 'subject48'\}.

For Fold 2, the training set is \{'subject46', 'subject49', 'subject12', 'subject13', 'subject35', 'subject9', 'subject5', 'subject17', 'subject18', 'subject3', 'subject26', 'subject20', 'subject4', 'subject31', 'subject14', 'subject24', 'subject11', 'subject16', 'subject40', 'subject45', 'subject1', 'subject32', 'subject34', 'subject41', 'subject33', 'subject36', 'subject43', 'subject42', 'subject39', 'subject10', 'subject48', 'subject38', 'subject37'\}, and the testing set is \{'subject47', 'subject25', 'subject44', 'subject27', 'subject8', 'subject22', 'subject30', 'subject23', 'subject15'\}.

\bibliographystyle{ACM-Reference-Format}
\bibliography{main}